# Visual Words for Automatic Lip-Reading

By
Ahmad Basheer Hassanat

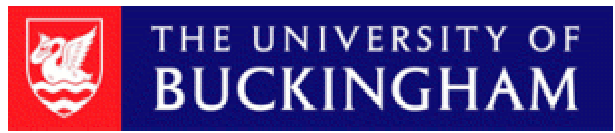

Department of Applied Computing
University of Buckingham
United Kingdom

A thesis submitted for the degree of Doctor of philosophy in computer science to the school of science in the University of Buckingham.

December 2009

# ABSTRACT

Lip reading is used to understand or interpret speech without hearing it, a technique especially mastered by people with hearing difficulties. The ability to lip read enables a person with a hearing impairment to communicate with others and to engage in social activities, which otherwise would be difficult. Recent advances in the fields of computer vision, pattern recognition, and signal processing has led to a growing interest in automating this challenging task of lip reading. Indeed, automating the human ability to lip read, a process referred to as visual speech recognition, could open the door for other novel applications. This thesis investigates various issues faced by an automated lip-reading system and proposes a novel *"visual words"* based approach to automatic lip reading. The proposed approach includes a novel automatic face localisation scheme and a lip localisation method.

The traditional approaches to automatic lip reading are based on visemes (mouth shapes (or appearances) or sequences of mouth dynamics that are required to generate a phoneme in the visual domain). However, several problems arise while using visemes in visual speech recognition systems such as the low number of visemes (between 10 and 14) compared to phonemes (between 45 and 53), Visemes cover only a small subspace of the mouth motions represented in the visual domain, and many other problems. These problems contribute to the bad performance of the traditional approaches, hence the visemic approach is something like digitising the signal of the spoken word, digitising causes a loss of information. While the proposed *"visual words"* considers the signature of the whole word rather than only parts of it. This approach can provide a good alternative to the visemic approaches to automatic lip reading.

The proposed approach consists of three major stages: detecting/localizing human faces, lips localisation and lip reading. For the first stage, we propose a face localization method, which is a hybrid of a knowledge-based approach, a template-matching approach and a feature invariant approach (skin colour). This method was tested on the PDA database (a video database, which was recorded using a personal digital assistant camera, contains thousands of video clips of 60 subjects uttering 18 different categories of speech in 4 different indoor/outdoor lighting conditions). The results were compared against a benchmark face detection scheme, and the results indicate that the proposed approach to localize faces outperforms the benchmark scheme. The proposed method is robust against varying lighting conditions and complex backgrounds.

For the second stage, we propose two colour-based lips detection methods, which are evaluated on a newly acquired video database and compared against a number of state-of-the-art approaches that include model-based and image-based methods. The results demonstrate that the proposed (*nearest-colour*) approach performed significantly better than the existing methods.

The proposed visual words approach uses a signature (2-dimensional feature matrix) that represents an entire spoken word. The signature of a spoken word is an aggregation of 8 features. These include appearance-based features, temporal information and geometric-based features extracted from the sequence of frames that correspond to the spoken word.

During the word recognition stage, the signatures of two words are compared by first calculating the similarity of each feature of the two signatures, which produces 8 similarity scores, one for each feature. A match score for the two words is calculated by taking a weighted average (i.e. score level fusion) of these scores that is then passed on to a KNN classifier. Differences in the duration of a spoken word are dealt with by using either dynamic time warping and/or linear interpolation. A weighted KNN classifier is proposed to enhance the word recognition rate.

The proposed visual words recognition system was evaluated using a large video database that consists of different people from varying backgrounds, including native and non-native English speakers, and large experiment sets of different scenarios. The evaluation has proved the "visual words" superiority over the traditional visemic approach, which researchers used to use for this kind of problem. These experiments have shown many results such as that the lip reading problem is a speaker-dependent problem, some persons produce relatively weak visual signals while speaking (termed as visual speechless persons), the performance of a lip-reading system can be enhanced by using a language model, etc.

The proposed approach for visual speech recognition was applied to the speaker recognition tasks, which could be used for *"visual passwords"*. It was also applied to a lip-reading surveillance application. Initial experiments indicate promising results, which lays a strong foundation for future work.

This work is dedicated to my lovely daughters Talah and Lujain, whose age is the same as this work, who learnt how to walk and talk far away from me, and without having the chance to read their lips.

# ACKNOWLEDGMENT

### Allah the most gracious and merciful

Who gave me the energy, health, nerves and provided me with all the people I am dedicating this hard work, which took me to spend a lot of determination and time until it came to light.

### My parents

I would like to designate the entire fruitful outcome of this work to my father who has been waiting for so long to see the result of his son's work, to my mother, who hasn't stopped praying for this work to be done, and to my father and mother in law for their support and encouragement.

### My Family

For all the patience, I cannot put my thankfulness and appreciation to my wife and lovely daughter in words, for all your support and encouragement you offered me during the life of this thesis. Also I will not forget my daughter's words which always ringing in my ears "when are you coming back daddy?" Whatever I do, it is not easy to repay you back. Many thanks go to the rest of my extended family, particularly brothers, sisters, nephews, nieces, and brothers and sisters in-law.

### My Supervisor

Prof. Sabah Jassim with my gratitude for all his patients, viable advice, discussions, convincing arguments and more during the life of this thesis. I wish him all the best for the future.

### My sponsors

I would like to thank the Jordan armed forces and Mu'tah University, Jordan (http://www.mutah.edu.jo) for sponsoring this programme of study.

### My Friends and Colleagues

A huge appreciation to all my friends everywhere, and my colleagues in the department of applied computing, in the University of Buckingham. for all their encouragements and supports, including Dr. Harin Sellahewa, Dr Naseer Al-Jawad, Dr. Johan Ehlers, Mr Hongbo Du, Professor Chris Adams,  Mr Ali Al-Sherbaz, Dr Kenneth Langlands, Mrs. Julie Leach, Ali Aboud, Hisham Al-Assam and all the PhD candidates in the department.
I would like also to thank Mr. Stefan Schmidt for helping in developing a GUI for the HMM tool.
Many thanks go also to Mrs. Marion Coban (Marmar) for editing this thesis.

# Abbreviations

| | |
|---|---|
| 3D | 3 Dimensional |
| A-B | Appearance-Based |
| AAM | Active Appearance Model |
| ANN | Artificial Neural Networks |
| ASMs | Active Shape Models |
| AVSR | Audio Visual Speech Recognition |
| BBC | British Broadcasting Corporation |
| CHMM | Continuous Hidden Markov Model |
| D | Distance |
| DBN | Dynamic Bayesian Network |
| DCT | Discrete Cosine Transformation |
| DFT | Discrete Fourier Transform |
| DHT | Discrete Hartley Transform |
| DT | Decision Tree |
| DTW | Dynamic Time Warping |
| DWT | Discrete Wavelet Transform |
| ED | Euclidean Distance |
| EM | Expectation Maximization |
| ER | Edge Ratio Feature |
| FAE | False Acceptance Error |
| FAR | False Acceptance Ratio |
| FRE | False Rejection Error |
| FRR | False Rejection Ratio |
| GUI | Graphical User Interface |
| H | Mouth Height |
| HCI | Human Computer Interaction |
| HCM | Hyper Column Model |
| HD | Hamming Distance/Score |
| HH | High-High (Used For Wavelet Sub-Band) |
| HL | High-Low (Used For Wavelet Sub-Band) |
| HMM | Hidden Markov Model |
| HSV | Hue, Saturation And Value. |
| HTK | Cambridge Hidden Markov Model Tool Kit |
| KNN | K-Nearest-Neighbour |
| IYUV | Intel Indeo Codec |
| LAL1 | Look-Alike 1 Words |
| LAL2 | Look-Alike2 Words |
| LDA | Linear Discriminative Analysis |
| LH | Low-High (Used For Wavelet Sub-Band) |
| LL | Low-Low (Used For Wavelet Sub-Band) |
| LG | Long General Words |
| M | Mutual Information |
| MESH | Normalized Triangle Mesh |
| MHI | Motion History Image |
| ML | Maximum Likelihood |
| MLP | Multi-Layer Perceptron |
| NN | Nearest-Neighbour |
| Nu | Numeric Words |
| PC | Principal Component |
| PCA | Principal Component Analysis |
| PDA | Personal Digital Assistant |
| PMF | Probability Mass Function |
| Q | Quality Feature |
| R | Vertical To Horizontal Features Ratio |
| RC | Red Colour Feature |
| RGB | Red/Green/Blue |
| ROI | Region Of Interest |
| SD | Speaker-Dependent |
| Sec | Security Words |
| SI | Speaker-Independent |
| SVM | Support Vector Machines |
| SWT | Stationary Wavelet Transform |
| T | Teeth Feature |
| T-B | Transform-Based |
| Viseme | Visual Phoneme |
| VSP | Visual-Speechless-Persons |
| VSR | Visual Speech Recognition |
| VSU | Visual Speech Units |
| VW | Visual Word |
| Vwords | Visual Words |
| VWR | Visual Word Recognition |
| W | Mouth Width |
| WER | Word Error Rates |
| WKNN | Weighted-KNN |
| YCbCr | Luminance, Chroma (Blue), Chroma (Red) |

# CONTENTS

# LIST OF FIGURES

# LIST OF TABLES

**DECLARATION**

I hereby certify that this material, which I now submit for assessment on the programme of study leading to the award of Doctor of Philosophy is entirely my own work and has not been taken from the work of others save and to the extent that such work has been cited and acknowledged within the text of my work.

Ahmad Hassanat

# Chapter 1

# Introduction

Lip reading is a technique used to understand or interpret speech especially by people with hearing difficulties. It helps, to some extent, persons with a hearing impairment to remove obstacles to engage in social activities, without which communicating with others would be difficult. Lip reading remains a very challenging task, even for the trained expert. Ability to lip read accurately requires years of practice, and a good knowledge of the language as well as the context in which a dialogue takes place. With the recent advances in technology and in the fields of computer vision, pattern recognition and signal processing, there is a growing interest in automating this challenging task of lip reading. Indeed, automating the lip-reading process could open the door for many novel applications. This thesis aims to contribute to the field of automatic lip reading, and to the wider area of computer vision and human computer interaction (HCI) by developing novel techniques for automatic lip reading.

Automatic lip reading, also referred to as visual speech recognition (VSR) or sometimes speech reading, has received a great deal of attention in the last decade for its potential use in applications such as HCI, audio-visual speech recognition (AVSR), speaker recognition, talking heads, sign language recognition and video surveillance. Its main aim is to recognise spoken word(s) by using only the visual signal that is produced during speech. Hence, VSR deals with the visual domain of speech and involves image processing, artificial intelligence, object detection, pattern recognition, statistical modelling, etc.

A typical VSR system includes image acquisition, lip localisation, feature extraction and recognition. The accuracy of a VSR system is heavily dependent on accurate lip localisation as well as the robustness of the extracted features. The lips and the mouth region of a face reveal most of the relevant visual speech information for a VSR system. Therefore, it is important for any VSR system to focus on the lips area. While some approaches aim to directly locate the lips of the subject in question (Gomez et al., 2002), others focus on the relatively easier task of locating the face and then locating the lips based on prior knowledge, e.g. the work of Zhang et al. (2003). However, accurate face and lip localisation are difficult tasks due to variations in sensor quality, background, light conditions, lips dynamic, shadowing, pose, facial expressions, scale, rotation and occlusion. In this thesis, we shall present a novel face location technique based on a hybrid of template matching, knowledge-based and feature invariant approaches.

Similarly, extracting a set of unique features from the visual signal to represent a word is difficult due to the fact that one person could utter the same word or sentence at different speeds and at different levels of loudness. This difficulty is magnified by the fact that different people talk in different ways, producing a variety of visual signals for the same word (intra-word variation). However, the most challenging task for a VSR system is to produce a unique signature (a feature representation) for a spoken word, when two or more words produce a similar visual signal (inter-word similarity), for example, the words "right" and "write".

The scope of this thesis includes three overlapping areas of research: face detection, lip localization, and VSR. The contribution of this thesis involves reviewing existing approaches and proposing novel methods in each area. This thesis investigates the use of different categories of features, namely, appearance, geometrical and temporal features to represent visual words. We shall demonstrate, through a number of experiments, the suitability of the proposed VSR system.

A brief description of the generic architecture of the VSR system is given in the next section. The rest of this chapter will present the aims and importance of this study, the claims of this thesis and the contribution of this work. Some of the challenges faced during the course of this study are highlighted in section 1.3. This chapter will conclude with a brief overview of the rest of the thesis.

## 1.1 Generic architecture of VSR

The most successful approaches to automatic lip reading depend mainly on recognising a visual speech unit called a "viseme" (the visual part of a phoneme). A viseme is a mouth shape (or appearance) or a sequence of mouth dynamics that are required to generate a phoneme in the visual domain. A phoneme is the smallest contrastive identifiable sound unit in the sound system of a language. Hence, a viseme is the shortest visually recognisable part of speech, and a phoneme is the smallest (shortest) audible component of speech. The human speech production system produces phonemes to construct a word and, by hearing these phonemes, we understand the spoken word. If we watch the lips at the same time, we can see the visemes to understand the spoken word.

Stages of a typical VSR system that use visemes include image/video acquisition, lip detection, feature extraction, visemes recognition followed by word recognition based on their visemes. This is based on the assumption that a recognised viseme or a sequence of visemes can be mapped onto a specific word. Typically, a phoneme is associated with a unique viseme or a sequence of visemes, but this is not true for all phonemes (Yu, 2008). Hence, an automated VSR system that relies on visemes is faced with the difficult scenario of having to recognise words, which have some phonemes that share the same viseme(s) or phonemes that have no associated viseme. This problem and other related issues will be discussed in detail in chapter 2, section 3.2.

The main goal of this study is to investigate and develop a *viseme-independent* approach to the automatic lip-reading problem to overcome some of the difficulties associated with the visemes-based approaches. The proposed system (see Figure 1.1) consists of three major stages: detecting/localizing human faces, lips localization and lip reading. Each stage consists of several steps and the three stages depend on each other (the accuracy of the preceding stage affects the accuracy of the proceeding stage).

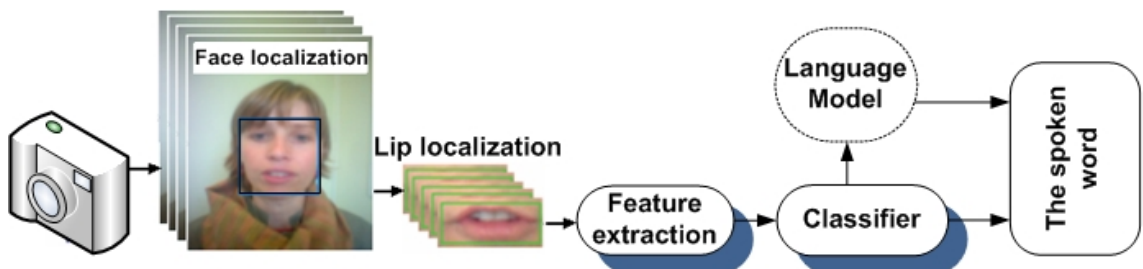

**Figure 1.1**. The proposed automatic lip-reading system (Vwords system).

In the first stage, which is a pre-process step for the second stage, the system attempts to localize the face in an image/video-frame. Localizing the face is an essential step as it decreases the search area for the lip-localization process (in the next stage) and increases the lip-localization accuracy (Hassanat and Jassim, 2008). (See also chapter 3).

Most of the VSR relevant information is located in the lip's appearance, shape and dynamics (Luettin et al., 1996a), and that is why it is called *lip reading*. Therefore, locating the lips and mouth area is vital for the word recognition accuracy of the proposed system. Hence, the second stage (lip localization) plays a central role in this system.

The last stage is the core of the system in which the visual features are extracted and the words are recognised. Unlike the visemic approach, this study proposes an holistic approach to tackle the VSR problem, where the system recognizes the whole word rather than just parts of it. In the proposed system, a word is represented by a signature that consists of several signals or feature vectors (feature matrix), e.g. height of the mouth, mutual information, etc.

Each signal is constructed by temporal measurements of its associated feature. The mouth height feature, for instance, is measured over the time period of a spoken word. This approach is referred to as the "*visual words*" (Vwords) approach.

A language model is an optional step that can be used to capture the properties of a specific language (English language in our case), and to predict the next word in a speech sequence by estimating the distribution of that natural language as accurately as possible. This can be done using some pre-defined syntax and/or semantic rules with one of the most common techniques for language models such as the N-gram model, the Structural Language Model, the Maximum Entropy Language Model, etc.

Adopting a language model could increase the VSR accuracy by reducing the size of the words domain (the number of the candidate words to be chosen by the VSR system as the predicted word) particularly when there are fairly similar signatures for different words (similar Vwords for different words) such as the words "right" and "write".

## 1.2 Aims and the importance of this study

Visual speech recognition has many real world applications such as in human-computer interaction, as a tool to help deaf and hearing-impaired people, talking heads applications, sign language recognition, remote surveillance systems, and speaker identification/ verification. The last two applications will be investigated in this thesis as potential application scenarios for the proposed VSR solution (see chapter 6). VSR is also used extensively to aid speech recognition systems to increase their accuracy, especially in noisy environments. This area of research is known as audio-visual speech recognition (AVSR).

An automatic speech or speaker recognition system uses the audio signal to recognise the spoken word or the speaker itself. However, the audio signal is susceptible to acoustic noise or in very noisy environments. Under such conditions, an AVSR system combines the visual and the audio signals of the spoken words in order to obtain higher recognition accuracy than using only the audio signal.

However, there are scenarios where the audio signal is unavailable or it is difficult to acquire, for example, when there are many people speaking at the same time in front of a camera, or talking in a noisy environment such as a stock market or a nightclub. The only people who can communicate in a noisy nightclub, for instance, are deaf people, who can use sign language, lip reading or both. In such circumstances, the visual signal becomes the only source of information to perceive speech. Therefore, studying VSR alone becomes very important as an active research area. Other applications related to VSR are discussed in chapter 2, section 2.3.1.

## 1.3 Challenges and difficulties

In spite of the expanding research in the fields of image processing, face detection, lip localization, VSR, AVSR, etc., there are many difficulties still facing VSR, which remain unsolved completely. Here we highlight some of the impediments to accurate and robust solutions to automatic visual speech recognition.

### 1.3.1 Difficulties related to image processing

The appearance of an object could vary due to sensor quality, the distance between the object and the sensor, the direction the object is being imaged and lighting conditions. Such variations affect the accuracy of automatic object detection and recognition. Therefore, there is a need for a pre-processing step to "normalise" such variations, prior to object detection, localization, recognition, etc.

**1.3.2 Difficulties related to the human face**

Automatic detection/localization of the face in a given image or a video is essential to many applications. However, due to variations in face size, pose and rotation and occlusion (e.g. glasses, caps, and scarves), face detection remains a difficult task. Another important factor to mention is the changes in face appearance due to different facial expressions such as anger, laughing, disgust, etc.

**1.3.3 Difficulties related to the human mouth and lip reading**

The human mouth is one of the most deformable parts of the human body, leading to different appearances such as mouth opened, closed, widely opened, tightly closed, appearance of teeth, tongue, etc. The variety in the appearance of the mouth is due to its major functions such as talking and contributing to facial expressions (laughing, sadness, disgust, etc.).

The major complication of the lip-reading problem for English language is that only 50% or less of speech can be seen. Also, each individual has his/her own way of speaking, particularly the visual aspect of speech, not to mention the length (in time) of a spoken word which is different from one person to another, and different for the same person depending on his/her mood and the speech time.

**1.3.4 Data collection and analysis**

Other major difficulties that emerged in this research were connected with data collection and analysis. A new database was recorded for this study, which consisted of 26 participants of different races and nationalities. Each participant recorded 2 videos (sessions 1 and 2) at different times, and uttered 30 different words five times in each recorded video. The video frame rate was 30 frames per second, and the average time for each video was about 8 minutes and 33 seconds. Each video contains about 15000 frames, and the total number of frames in the database ≈ 15,000 frame * 2 Sessions * 26 persons = 780,000 frames.

Another video database was used in this study to evaluate the proposed face detection method (see chapter 3), which is called the PDA database (Morris, et al., 2006). This database has 60 speakers of 3 age groups, 10 males and 10 females in each group. Each speaker was recorded in 3 recording sessions separated by at least one week. Each session consists of 2 outdoor and 2 indoor recordings. The light conditions were not controlled indoor and outdoor.

Videos were recorded for 3 types of prompt (5-digit, 10 digit and short phrase), with 6 examples from each prompt type. Video data consists of a total of: (2 genders) x (3 age groups) x (10 subjects per gender per 2 age group) x (3 well separated sessions) x (2 recording locations) x (2 combinations of voice and face recording conditions) x (3 prompt types) x (6 examples per prompt type) = 12,960 recordings (see Table 1.1).

Table 1.1. The three prompt types in the PDA database.

| Index | 5-digit strings |
|---|---|
| 1 | 5 3 8 2 4 |
| 2 | 6 2 1 9 7 |
| 3 | 4 2 7 1 3 |
| 4 | 2 8 3 7 6 |
| 5 | 1 9 8 5 4 |
| 6 | 4 5 2 3 9 |
| **Index** | **10-digit strings** |
| 7 | 4 3 1 3 8 7 4 6 1 5 |
| 8 | 2 9 2 8 7 3 7 9 3 8 |
| 9 | 5 7 9 2 4 7 9 1 2 6 |
| 10 | 3 9 6 4 6 3 7 6 3 1 |
| 11 | 6 4 2 1 4 7 1 5 3 4 |
| 12 | 1 2 6 1 6 9 2 9 8 1 |
| **Index** | **Phrases** |
| 1 | Stop each car if its little |
| 2 | Play in the street up ahead |
| 3 | A fifth wheel caught speeding |
| 4 | Charlie, did you think to measure the tree? |
| 5 | Tina got cued to make a quicker escape |
| 6 | Here I was in Miami and Illinois |

Evaluating large databases (780,000 frames/images from the in-house database and 521299 frames/images from the PDA database) makes following up face and lip localization results for all frames (to make sure that the algorithms work well) a very long and hard task. Also the large number of experiments needed to be conducted for this study on this large database yielded a vast amount of data and results to analyze.

It was also difficult to convince participants to sit in front of a camera for 8 minutes, and to come back at another time for the second session. For privacy reasons, it was also difficult to find people who were willing to be recorded and to have their face images published.

The aforementioned difficulties generate additional difficult tasks for computer systems to deal with, which drains computer resources and consumes time, instead of utilizing all the computer resources to deal with the actual problem, and saving time, which is precious for on-line applications.

This study is not intending to solve all these problems and difficulties, but rather aims to investigate the VSR problem, trying to obtain a better understanding of the problem, and also achieve better performance through alleviating some of these challenges and problems.

## 1.4 Thesis claims

The central goal of this research is to validate the following claims:

1. *The VSR problem is a speaker-dependent problem.*

2. *Using the proposed visual words (Vwords) as a substitute for a Visemes approach, it is possible to increase the performance and accuracy of current automatic lip- reading systems depending on visual information only.*

3. *Vwords can be applied efficiently for speaker identification/verification, and security, through the person's utterance, depending on his/her different (unique to some extent) way of speech.*

4. *The (relatively) little information provided by the visual side of speech can be compensated by using a language model, which limits the choices for the predicted word and thus reduces errors.*

This study supports these claims by conducting a large number of experiments, which have shown encouraging results compared to recent studies in *VSR*. It is believed that there is also room for considerable improvements in future work.

## 1.5 Contributions

The central contribution of this study to the research community (particularly the VSR community) is the development of a robust VSR system using the proposed Vwords approach (see Figure 1.1). There are also some other important contributions of this research, which are related to other areas such as computer graphics, biometrics, AVSR, the deaf community and security.

**Computer graphics**

Most of the work done on automatic lip reading, face detection, lip localization and visual feature extraction contributes (fully or partially) to the computer graphic research community. This project is adding (to this area) new techniques by proposing new methods for face and lip localization, as well as new visual features for the proposed lip-reading approach (Vwords). The proposed localization methods can be altered to be used for general object detection/localization and pattern recognition problems. Two new filters were also proposed for edge/features detection.

Furthermore, the new-recorded database for lip-reading purposes can be used in other biometrics and image processing research studies. At least two PhD candidates in the University of Buckingham who work on image quality and biometrics cancellation started to have benefits from using this database.

Moreover, a new library with a graphical user interface (GUI) has been designed using Microsoft® visual c++.net 2003. The library consists of several image processing algorithms such as filtering, segmentation, edge detection, features extraction, motion detection, colour conversion, image transformation, clustering and recognizing. The library was used for the purposes of this study and some other researchers in image processing have started to use some of its functions.

**Biometrics**

The proposed visual words system can be applied in many biometric areas. This study focuses on speaker identification and verification problems as possible areas where Vwords can be applied effectively. Vwords can also be used in conjunction with face recognition to improve a system's performance. Moreover, the proposed face localization method can be used as a pre-process for face or iris recognition.

**Speech recognition**

VSR approaches are used extensively to aid speech recognition systems and improve their performance, especially in noisy environments, which has created a new area of research known as AVSR. Therefore, Vwords can be used for the same purpose and contribute to the speech recognition community.

**Deaf community**

This community gains the most benefit from such a project because it touches their lives directly by providing them with a robust lip-reading system, which can help them to perceive human speech, either by using the proposed system alone or in conjunction with other gesture language recognition systems.

**Security**

To increase system security, Vwords is proposed to be used as "visual passwords" to strengthen the normal text passwords. Vwords are also useful for security purposes such as in the case of surveillance when it is difficult to capture the voice of a person whose face appears on a camera from a distance.

## 1.6 Thesis overview

This thesis consists of 8 chapters. The first chapter presents a brief introduction to the study, provides some definitions, briefly describes the proposed system, highlights some challenges and states the contribution of this work.

Chapter 2 presents human lip reading (speech production and human ability to read lips), briefly reviews automatic lip-reading literature and describes the (state-of-the-art) Visemic approach for VSR.

Face detection/localization literature is reviewed in chapter 3, as well as a detailed description of the proposed face localization method, and a discussion of the related experimental results.

Lip localization literature is reviewed in chapter 4, in addition to describing the new database and the proposed lip-localization methods, and discussing the related experimental results.

Chapter 5 describes the proposed visual speech recognition method in detail by illustrating the visual features and the different methods used to extract them. In addition, the distance functions and the recognition methods that were used in the proposed system are described and explained.

Chapter 6 evaluates the proposed system, by discussing the experimental results and findings, as well as comparing the proposed work with some state-of-the-art methods, particularly visemes and Hidden Markov Models (HMM).

Chapter 7 describes some possible applications for the proposed Vwords approach, which include speaker identification/verification and lip-reading security surveillance system.

The last chapter (8) summarizes the overall results, draws final conclusions and indicates possible future directions.

# Chapter 2

# Lip-Reading Literature

## 2.1 Introduction

Since the first word was spoken by mankind, speech became the most widely spread means of communication between individuals. This communication depends very much on the audio signal that is produced by the human speech system. The audio signal alone was more than enough for human communication in normal conditions, perhaps because humans were yet to evolve to have needs for a more sophisticated communication system. It was sufficient for man's survival to use sound to warn friends if there was an enemy or a predator around, or to call on someone who had got lost in a forest. The same applied to the women back in camps – if one needed to call her neighbour all she had to do was to use her voice. And none had to see the others, or the others had to see the speaker to understand the speech if the speech was produced using the human voice (the audio signal). Consequently, the art of speech has thrived throughout human history and has yielded a large number of different languages, dialects and accents.

The concept of recognizing/interpreting speech using visual signals only (i.e. using the changing image of the mouth area due to movement of lips during speech) is called lip reading, i.e. lip reading is akin to "seeing" speech. In contrast to hearing speech, lip readers need to observe the speaker's lips, which is difficult to attain at all times.

Lip reading is not as essential as hearing for most humans. However, for deaf people or when the environment is very noisy, it becomes important and possibly the only way for speech to be perceived/comprehended. It is also useful for security purposes such as in the case of surveillance when it is difficult to capture the voice of a person whose face is captured on a camera from a distance. Lip reading has been known for a long time, but because the visual signal does not give as much information as the audio one, it has not received as much attention as it deserves in order to develop a better communication system. This thesis is concerned with automatic recognition of words and phrases from the visual signals captured by a camera while the speaker utters these words/phrases.

The rest of this chapter will be devoted to discussing the human ability to produce and read speech, in addition to reviewing the VSR literature, paying extra attention to the most used state-of-the-art approach (the visemes) and ending the chapter with the motivation for this study.

## 2.2 Human lip reading

Lip reading is not a contemporary invention; it was practised as early as 1500 AD, and probably before that time (De Land, 1931). The first successful lip reading teacher was the Spanish Benedictine monk, Pietro Ponce, who died in 1588. Lip reading teaching subsequently spread to other countries (Bruhn, 1920).

The German Samuel Heinecke opened the first lip reading school in Leipzig in 1787 (Bruhn, 1920). The first speech reading conference was held at Chautauqua, USA in 1894 (De Land, 1931).

Several different methods have been described in the literature for human lip reading such as the Muller-Walle, Kinzie, and the Jena methods. The Muller-Walle method focuses on the lip movement to produce a syllable as part of words, and the Kinzie method divides lip reading teaching into 3 teaching levels, depending on the difficulty (beginners, intermediate and advance) (De Land, 1931).

Although only 50% or less of speech can be seen, the reader must guesstimate those words that he/she has missed. This was the core of the Jena method: training the eye and exercising the mind (De Land, 1931). However, regardless of the variety of known lip reading methods, all methods still depend on the lip movement that can be seen by the lip reader.

### 2.2.1 Speech production and physiology

In order to understand the link between the audio signal and the corresponding visual signal that can be detected on the face of the speaker, we need to have some understanding of how speech is produced. "Physiology is the science that deals with the functions of biological systems" (Lieberman and Blumstein, 1988:3). The different biological systems responsible for the production of speech in humans can be categorized into three subsystems (Lieberman and Blumstein, 1988):

1. The subglottal system, which consists of the lungs (see Figure 2.1).

2. The larynx, consisting of the vocal cords.

3. The supralaryngeal vocal tract, which consists of airways for nose, mouth and pharynx.

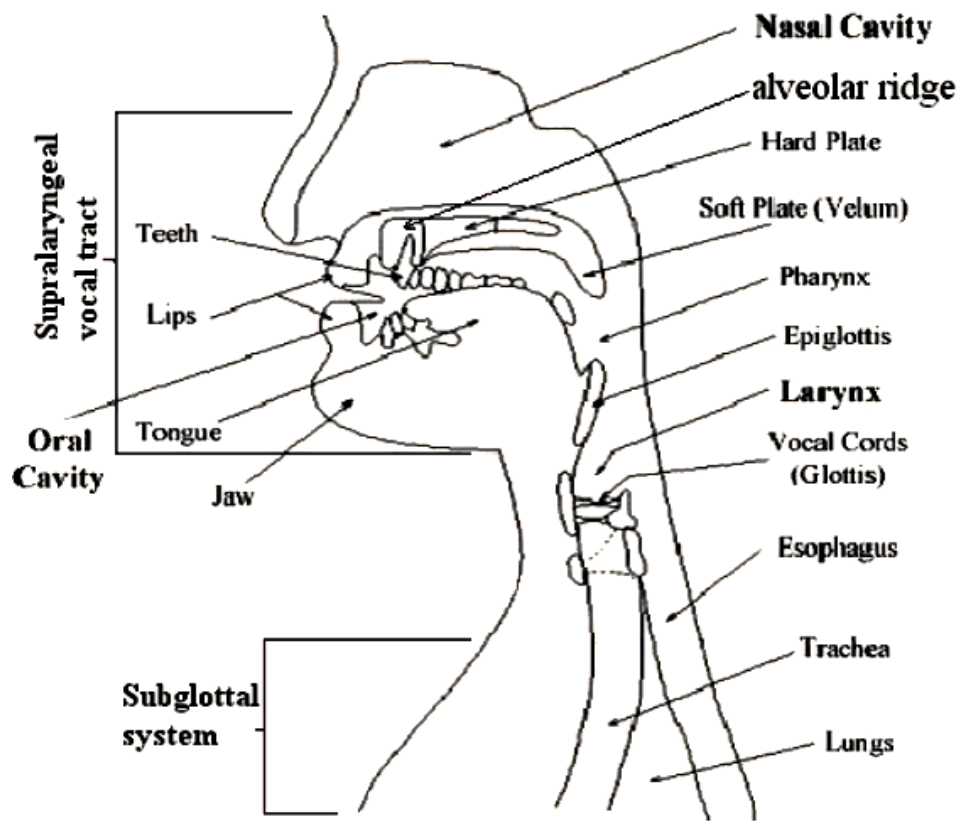

**Figure 2.1**. Human Physiology of Speech Production[*].

Human sound is produced by blowing air from the lungs through the larynx which starts to open and close rapidly by moving the vocal cords together or pulling them apart, to produce puffs of air, which contain acoustic energy at an audible frequency. These puffs of air are then passed to the upper system (supralaryngeal vocal tract) which works as a variable acoustic filter for the source of acoustic energy, the variation of the filter depending on changing the shape of the vocal tract by the speaker while speaking.

---

[*] As in Tan (1996) but with some modifications.

Therefore, the airflow is changed in a range of different ways by different aspects of the human speech system (see Figure 2.1). When the lung blows the air out, each subsystem and its component contribute (fully or partially) in producing the uttered phoneme.

Humans can hear all phonemes but cannot see all of them; the visual parts of the human speech system are the lips, jaw, teeth and the tongue; consequently one of the major problems for (human or machine) lip-reading is the full production of the phoneme by the invisible parts of the human speech system, such as the glottal (h), and the velars (k and g). Table 2.1 shows the places and manners of some phonemes.

**Table 2.1**. Classifying phonemes according to their manner and production place, the paired signs are (voiceless, voiced) consonants, SAMPA signs are used.

| Place / Manner | Labial | Labio-Dental | Dental | Alveolar | Palato-Alveolar | Palatal | Velar | Glottal |
|---|---|---|---|---|---|---|---|---|
| **Plosive** | (p, b) | | | (t, d) | | (c, J) | (k, g) | |
| **Fricative** | (p\\, B) | (f, v) | (T, D) | (s, z) | (S, Z) | (C, j\\) | | h |
| **Affricate** | | | | | (tS, dZ) | | | |
| **Nasals** | m | F | | n | | | N | |
| **liquid** | | | | l, r | | | | |
| **semivowel** | w, H | | | | | j | | |

Phonemes are named after the places where they are produced, such as the *labial* sounds, which are the sounds that are produced by pressing the lips together. The *labio-dental* sounds are produced using the lips and teeth together, placing the tip of the tongue behind the upper teeth produces the *dental* sounds, tapping the tongue against the area a bit behind the teeth creates the *alveolar* sounds. The *palate-alveolar* sounds initiate between the alveolar ridge and the hard palate. The *palatal* are produced from the hard palate (middle of the roof of the mouth). The *velar* sounds are produced in the soft palate (velum). And the *glottal* sounds are initiated in the glottis.

The manner of articulation is generally used to explain how sounds are produced, such as the *plosive* sound, which occurs suddenly after blocking and releasing the air by the oral and nasal cavities. The *fricative* sound is created by blowing air through a thin channel made by placing two organs closely together (e.g. lower teeth and upper lip in /f/). The *nasal* sound occurs when air flows out through the nose, the position of the tongue determines the sound, then the *liquid* sound is articulated using the side of the tongue and the semivowel sound is articulated like a vowel but with the tongue closer to the roof of the mouth (Cawley, 1996).

A vowel is a sound in spoken language that has its own sounding voice (vocal sound), it is pronounced by comparatively open configuration of the vocal tract. So that there is no build-up of air pressure at any point above the glottis, in contrast to a consonant, where there is a constriction or closure at some point along the vocal tract. A single vowel sound outlines the basis of a syllable; two adjacent vowel sounds can be blended together into a single syllable. Semivowels are non-syllabic vowels.

Different vowel sounds are distinguished by their articulatory features, such as height (vertical dimension), backness (horizontal dimension) and roundedness (lip position). Vowel height describes the vertical location of the tongue relative to either the aperture of the jaw or the roof of the mouth. The International Phonetic Alphabet (IPA) identifies seven different vowel heights: close (high), near-close, close-mid, mid, open-mid, near-open and open (low) vowel. Vowel backness describe the location of the tongue relative to the back of the mouth, IPA identifies five levels of vowel backness: front vowel, near-front, central, near-back and back vowel. *Roundedness* refers to the shape of the lips whether it is rounded or not. Usually mid to high back vowels are rounded vowels. Some vowels articulatory features are indicated in Table 2.2. Appendix A shows a simplified list of both constant and vowels phonemes using both IPA and the Speech assessment methods phonetic alphabet (SAMPA)

**Table 2.2.** Classifying vowels according to their height and backness, the paired signs are (unrounded, rounded) vowels, SAMPA signs are used.

| Backness / Height | Front | Central | Back |
|---|---|---|---|
| **Closed or high** | (i, y), (I, Y) | (1, }) | (M, u), U |
| **Half closed** | (e, 2) | (@\, 8), @ | (7, o) |
| **Half open** | (E, 9) | (3, 3\), 6 | (V, O) |
| **Open or low** | { | (a, &) | (A, Q) |

According to De Land (1931), Mole and Peacock (2005) and Bauman (2000), only 50% or less of English sounds can be seen, i.e. visually identified. In order for a sound to be easily read, it must be articulated on the lips and/or in the visual part of the mouth. Many English sounds are articulated in the middle of the mouth. Others come from the back of the mouth or the throat. Therefore, the easiest sounds for lip reading are the labials, labio-dental sounds, and the dentals. While the hardest are the glottal sounds and the velars (see Table 2.1).

## 2.2.2 Human lip-reading skills

According to a recent study conducted at the University of Manchester, people with hearing problems could understand about 21% of speech, but if they used either a hearing aid or lip reading, they could understand 64%. If they used both hearing aids and lip reading, their speech comprehension rises to 90% (Bauman, 2000).

Potamianos et al. (2001) described a human speech perception experiment. A small number of human listeners were presented with the audio once and the audio and video of 50 database sequences from an IBM ViaVoice database single speaker, with different bubble noises added each time. The participants were asked to transcribe what they heard and viewed.

Potamianos et al.'s (2001) experiment is not a pure lip reading experiment, as its aim was to measure the effect of the visual cues on the human speech perception, rather than the perception of the speech without the audio. The experiment showed that human speech perception increases by seeing the video and watching the visual cues. The word error rate was reduced by 20% when participants viewed the video, showing that the human audio-visual speech perception is about 62% word accuracy. According to the previous study, about 30% of the participants were non-native speakers, and this is one of the reasons why the recognition rate was very low, despite both the audio and video signals being revealed.

A lip-reading experiment was conducted in this study to roughly measure the human ability for lip reading, and the amount of information that can be seen from speech.

Four video sequences from the PDA Database were used in this experiment, 2 males and 2 females; each video spoke 10 digits; the digits and their sequences are different from one video to another, and the audio signals were removed from the four videos. Fifty five participants were asked to transcript what each video spoke; each participant can play each video up to 3 times, so participants would have enough time to decide what the spoken digits were, and they would not be fooled by the speed of the video. These videos were uttering only digits, {1,2,3,...,9}, the participants were informed about this domain (the speech subject), hence it is much easier for humans to read lips if they know the subject of the talk, and also it mimics automatic lip-reading experiments since the recognizer algorithm knows in advance and is trained on all the classes (the words) to be classified; the average word recognition rate for all the participants was 53%. See Table 2.3.

**Table 2.3**. Human lip reading results.

| Subject | Result |
|---|---|
| 1 (Female) | 61% |
| 2 (Male) | 50% |
| 3 (Male) | 37% |
| 4 (Female) | 63% |
| Average | 53% |

As can be seen from Table 2.3, some videos were easier to read than others (61% and 63% for videos 1 and 4 respectively), where some other videos have less information for lip readers, or those people by nature either speak faster than normal, or do not produce enough information for the lip readers. We can notice that the females give more information for the readers; it is, of course, difficult to substantiate such a claim because this is a small experiment using a small number of videos, so it is too early to draw such conclusions with such evidence. The most important thing that this experiment can reveal so far is the overall human lip reading ability, which is 53%. Another interesting thing to mention is that different people also have different abilities to perceive speech from visual cues only. In this experiment the best lip reader result was 73%, while the worst was 23%. These experiments illustrate the variation in individual lip reading skills, and the variation in individual ability to produce a clear readable visual signal, which would add to the challenge of designing an automatic lip reading system.

## 2.3 Automatic lip reading

The human ability for lip reading varies from one person to another, and depends mainly on guessing to overcome the lack of visual information. Needless to say, lip readers need to have a good command of the spoken language, and in some cases the lip reader improvises and uses his/her knowledge of the language and context to pick the nearest word that he/she feels fits into the speech. Moreover, human lip readers benefit from visual cues detected outside the mouth area (e.g. gestures and facial expressions). The complexity and difficulties of modelling these processes present serious challenges to the task of automatic visual speech recognition.

## 2.3.1 Literature review

Automatic lip reading, also known as speech reading or visual speech recognition (VSR), is a technique of recognizing speech by machines, using visual cues obtained from the images of the mouth, lips, chin and any other related part of the face.

Automatic lip reading or VSR problems have received a great deal of attention from researchers in the last two decades, because of the large areas and applications that are related to this problem. A VSR system can:

- Aid speech recognition – speech recognition systems are highly affected by acoustic noise, while visual speech recognition does not because it depends on the visual signal rather than the audio signal, thanks to the researchers of the speech recognition field, as most of the visual speech recognition work has been accomplished through audio-visual speech recognition researches.

- Work as a complete system, with sign language recognition, to help deaf people.

- Be used in human computer interaction.

- Help in facial expressions recognition.

- Be used in talking heads systems for synthesizing visual speech.

- Be used in surveillance systems for security reasons, and be used for speaker identification/verification – these applications will be touched on and evaluated in this study.

An objective of this study is to investigate if automatic lip reading could help in speaker identification/verification, by exploiting individual differences in pronunciation, articulation, and talking in general as behavioural biometric characteristics.

A typical lip-reading solution consists of two major steps (Sagheer, et al., 2006):

1- Feature extraction.

2- Visual speech feature recognition.

Existing approaches for feature extraction can be categorised as:

1. Geometric-feature based - obtains geometric information from the mouth region such as the mouth shape, height, width and area.

2. Appearance-based methods - these methods consider the pixel values of the mouth region, and they apply to both grey and coloured images. Normally some sort of dimensionality reduction of the region of interest (ROI) (the mouth area) is used such as the principle component analysis (PCA), which was used for the Eigenlips approach, where the first *n* coefficients of all likely lip configurations represented each Eigenlip.

3. Image-transformed-based methods - these methods extract the visual features by transforming the mouth image to a space of features, using some transform technique, such as the discrete Fourier, discrete wavelet, and discrete cosine transforms. This transform is important for dimensionality reduction, and to eliminate redundant data. Although some researchers consider it as image or appearance-based approach, this approach is different in the sense that not all the pixels of ROI will be taken into consideration, but rather it considers the most distinguished pixels (the real features) such as edges, contours, corners, etc. These details are refined using some kind of image transformation.

4. A hybrid approach, which exploits features from more than one approach.

**Geometric features-based approaches**

A geometric features-based approach includes the first work on VSR done by Petajan in 1984, who designed a lip-reading system to aid his speech recognition system. His method was based on using geometric features such as the mouth's height, width, area and perimeter. These features were obtained from the mouth binary images, which were extracted from the face images using a simple thresholding technique. Petajan used the dynamic time warping (DTW) method for recognition, which improved the performance of his speech recognition system (Petajan, 1984).

Another recent work in this category is the work done by Werda et al. (2007), where they proposed an Automatic Lip Feature Extraction prototype (ALiFE), including lip localization, lip tracking, visual feature extraction and speech unit recognition.

Lip localization was done by employing a snake technique (see chapter 4). Their shape-based visual feature extraction method extracts the mouth height, width, and inner area features. The extracted visual information is classified to recognize visemes using artificial neural networks (ANN); their experiments yielded 72.73% accuracy of French vowels uttered by multiple speakers (female and male) under natural conditions (Werda, et al., 2007).

Since the first automatic lip-reading system proposed by Petajan (1984) to enhance speech recognition, tens of articles have been published to investigate the same subject for its importance, using different approaches for extracting features such as appearance-based, image-transformed-based and hybrid methods.

**Appearance-based approaches**

Such an approach was inspired by the methods of Turk and Pentland (1991) and first proposed by Bregler and Konig (1994). Eigenlips are: "a finite set of orthogonal images which constitute, up to a certain accuracy, a subspace for the representation of all likely lip configurations" (Belongie and Weber, 1995: 4).

In other words, Eigenlips are the compact representation of mouth Region of Interest using PCA. Bregler and Konig (1994) determined the Eigenlips after approximating the location of the lips using a snake technique. Then they considered the first 10 eigenvectors to represent lip shapes in the grey level. After constructing the set of Eigenlips, images of subsequent lip configurations can be represented in terms of their projections onto the set of Eigenlips.

A hybrid recognition method was designed by employing a Multi-Layer Perceptron (MLP) and an HMM. Their method was evaluated using a German database consisting of 2 females and 4 males. The reported word error rates (WER) were in the range of 10% to 51% depending on the provided noise. The performance of the proposed system is demonstrated by the case of word recognition over a small database of spoken digits.

Another Eigenlips-based system was investigated by Arsic and Thiran (2006), who aimed to exploit the complementarity of audio and visual sources, focusing on the principal components projections of the mouth region images. According to this study, the retained eigenvectors capture the major variations across the training set, but these variations are related to lighting directions and carry no information about the relevance of selected features with respect to speech classes.

Therefore they proposed using an information theoretic approach and the basic principle of mutual information for selecting the relevant eigenfeatures. This is done by calculating mutual information for each principal component and word classes. Eigenvectors were sorted in descending order of the mutual information, and the most informative eigenvectors were then used to find projections of each mouth image from the database on the Eigenspace. The obtained projection coefficients were used as inputs of the visual feature vector for the recognition task (Arsic and Thiran, 2006).

Their study utilized a relatively small audio-visual database called Tulips1, where 12 persons are pronouncing the first four digits in English twice. The best reported visual-only rate was 89.6% using mutual information with PCA, and 81.25% using normal PCA, while the result of using both audio and visual signals was about 95%.

Belongie and Weber (1995) introduced a lip-reading method using optical flow and a novel gradient-based filtering technique for the features extraction process of the vertical lip motion and the mouth elongation respectively. The extracted features are then encoded as 1D waveforms, which are sent to principle components analysis (PCA) to be classified.

The method was evaluated using a small database of spoken digits, with only two speakers speaking 7 digits. Belongie and Weber (1995) claimed a recognition rate of more than 98% for the speaker-dependent experiment, and about 47% for the speaker independent. Noting that the small number of participants and words (classes) in the database, both numbers of subjects (2) and classes (7) are insignificant, we cannot use insignificant numbers to draw statistically significant conclusions. A 98% word recognition rate (which is highly questionable even in such circumstances) outperforms state-of-the-art speech recognition systems, which use the rich audio signal!

In a more recent study, Hazen et al. (2004), developed a speaker-independent audio-visual speech recognition (AVSR) system using a segment-based modelling strategy. This AVSR system includes information collected from visual measurements of the speaker's lip region using a novel audio-visual integration mechanism which they call a segment-constrained Hidden Markov Model (HMM).

Visual features were extracted from the raw images of the mouth region using the visual front end of the AVSR Toolkit. Each image was normalized for lighting variation using histogram equalization. Then, a PCA transform was applied, and the top 32 coefficients retained as feature vectors.

Hazen et al.'s (2004) *Segment-Constrained-HMMs* was implemented in three primary steps:

1. Fixed-length video frames are mapped to variable-length audio segments defined by the audio recognition process.

2. Each context-dependent phonetic segment is mapped to a context-dependent segment-constrained viseme HMM.

3. The segment-constrained viseme HMM uses a frame-based Viterbi search over visual frames in the segment to generate a segment-based score for the visemic model.

By using the aforementioned techniques, the *Hazen et al system* phonetic error rate reduced by 2.5% during their experiments.

Gurban and Thiran (2005) developed a hybrid SVM-HMM system for audio-visual speech recognition, the lips being manually detected. The pixels of down-sampled images of size 20 x 15 are coupled to get the pixel-to-pixel difference between consecutive frames. They used the Tulips1 database, and they evaluated their method using the cross-validation using leave-one-out method. The word accuracy was around 80% for both VSR and AVSR, but increased to 91% when both approaches were integrated together.

Saenko et al. (2005) proposed a feature-based model for pronunciation variation to visual speech recognition; the model uses dynamic Bayesian network DBN to represent the feature stream. Support vector machine SVM was used for classification.

Sagheer et al. (2006) introduced an appearance-based lip-reading system, employing a novel approach for extracting and classifying visual features termed as "Hyper Column Model" (HCM); HMM was used for recognition.

Yau et al. (2006) described a voiceless speech recognition system that employs dynamic visual features to represent the facial movements. The system segments the facial movement from the image sequences using motion history image MHI (a spatio-temporal template). The system uses discrete stationary wavelet transform (SWT) and Zernike moments to extract rotation invariant features from MHI (Zernike moments are image moments normally used in image patterns recognition), an artificial neural network ANN is used for viseme classification.

**Image-transformed-based approaches**

Lucey and Sridharan's (2008) work was designed to be pose invariant. Their audio-visual automatic speech recognition was designed to recognize speech regardless of the pose of the head, the method starting with face detection and head pose estimation. They used the pose estimation method described by Viola and Jones (2001). The pose estimation process determines the visual feature extraction to be applied either on the front face, the left or the right face profile.

The visual feature extraction was based on the DCT, which was reduced by the LDA, and the feature vectors were classified using HMM. The best performance for this method was 47% word error rate for the normal pose, with about 72% word error rate for lip reading using either the left or the right profile. The results were obtained by applying the method on the CUAVE database (Patterson et al., 2002). The part of the database used was 33 speakers speaking isolated digits sequences.

A very recent study which also fits into this category was done by Jun and Hua (2009), where they used discrete cosine transformation (DCT) for feature extraction from the mouth region, in order to extract the most discriminative feature vectors from the DCT coefficients. The dimensionality was reduced by using linear discriminative analysis (LDA), and Hidden Markov model (HMM) is employed to recognize the words.

Jun and Hua (2009) evaluated their method using two databases, one containing only one subject. They used it for the speaker-dependent experiment, and another database, containing 40 subjects, was used for the speaker-independent experiment. They compared their method (DCT reduced by LDA) with DCT reduced by PCA (principle component analysis), and the Zig-Zag method. Their results (maximum 65.9% word recognition rate for the user-dependent experiments and about 23% for speaker-independent) showed that the LDA approach is better than both the PCA and the Zig-Zag method. The classifier used for this work was the continuous HMM (CHMM).

The speaker-dependent result is much higher than that of the speaker-independent. They justify this difference by the use of DCT, which is an appearance-based feature extraction method, and the mouth appearance is very different from one to one, even when both persons speak the same word. Another reason is that the DCT and appearance-based feature extraction methods in general lack the robustness to illumination and lighting conditions (Jun and Hua, 2009).

**Hybrid approaches**

Neti et al. (2000) proposed an audio-visual speech recognition system, where visual features obtained from DCT and active appearance model AAM were projected onto a 41 dimensional feature space using the LDA. Linear interpolation was used to align visual features to audio features.

Simple audio-visual feature concatenation and hierarchical discriminant features fusion were both used for audio-visual information fusion. Asynchronous decision fusion by means of the multi-stream HMM was also employed (Neti et al., 2000).

Their system was evaluated using a large vocabulary database (the IBM ViaVoice audio-visual database), which consists of 290 subjects, continuously reading speech with a 10400-word vocabulary. The system does not have a specific part for visual-only recognition, because of the audio-visual feature fusion, but they reported in general "introducing the visual modality reduced [speech recognition] word error rate by 7% relative in clean speech, and by 27% relative at an 8.5 dB SNR audio condition" (Neti et al., 2000).

A comparative viseme recognition study by Leszczynski and Skarbek (2005) compared 3 classification algorithms for visual mouth appearance (visemes): 1) DFT + LDA, 2) MESH + LDA, 3) MESH + PCA. They used two feature extraction procedures: one was based on normalized triangle mesh (MESH), and the other was based on the Discrete Fourier Transform (DFT), the classifiers designed by principle component analysis (PCA), and Linear Discriminant Analysis (LDA).

The goal of this study was to compare the performance and accuracy for those viseme recognition algorithms, and the recognition rate was 97.6%, 94.4%, 90.2% respectively, MESH + LDA was the fastest, while MESH + PCA was the slowest. The previous study used the Cambridge Hidden Markov Model tool kit (HTK) for phoneme classification, but tested only the viseme recognition, and did not test or mention anything about the word recognition, or the word error rate.

Each of the previous approaches has its own strengths and weaknesses. Sometimes the data reduction methods cause the loss of a considerable amount of related data, while using all the available information takes a much longer processing time, and not necessarily to obtain better results due to video or image-dependent information. Much effort has been invested to propose any combination of the different approaches, to trade the disadvantages of each individual approach (Sagheer et al., 2006).

The previous approaches can be further classified depending on their recognition and /or classification method. Researchers usually use dynamic time warping (DTW), e.g. the work done by Petajan (1984), artificial neural networks (ANN), e.g. the work done by Yau et al. (2006) and Werda et al. (2007), Dynamic Bayesian Network (DBN), e.g. the work done by Belongie and Weber (1995), and support vector machines (SVM), e.g. the work done by Gurban and Thiran (2005) and Saenko et al. (2005).

The most widely used classifier in the VSR literature is the hidden Markov models (HMM), which are statistical model transitions between the visual speech classes (HMMs are discussed in detail in chapter 6). The methods that use HMM include Bregler and Konig (1994), Neti, et al. (2000), Potamianos et al. (2003), Hazen et al (2004), Leszczynski and Skarbek, (2005), Arsic and Thiran (2006), Sagheer, et al. (2006), Lucey and Sridharan (2008), Yu (2008) and Jun and Hua (2009).

All these classifiers and more are also used to detect faces. Such classifiers are discussed in detail in chapter 3.

## 2.3.2 The Visemes Approach

From the above review we can conclude that most of the work done was based on visemes (see Table 2.3), while each word is viewed and recognized by its visemes combination. (This conclusion is backed up by Foo and Lian (2004) and Yu (2008).) A viseme is a mouth shape (or appearance) or a sequence of mouth dynamics that are required to generate a phoneme in the visual domain. In other words, a viseme is the visual part of a phoneme, hence phonemes are the basic parts in speech, visemes are the basic parts in visual speech, and the smallest unit that can be identified using visual information (Yu, 2008).

Several problems arise while using visemes in visual speech recognition systems (for English language in this study) such as:

1. There is no absolute (standard) number for the visemes and phonemes classes. Some researchers distinguished 10, 13, 14 or 16 visemes, and 44 or 53 for the phonemes. See Figure 2.2.

2. There is no standard set for the visemes. Every researcher has his own visemic set, and even if the number of the visemes was the same for two studies, it is not necessary to have the same visemic set in both studies. See Figure 2.2.

3. Some phonemes share the same viseme, i.e. several phonemes are mapped to one viseme. See Figure 2.2.

4. Some visemes share the same phoneme, i.e. one phoneme is mapped to more than one viseme, depending on the next phoneme, for example: notice the phoneme *n* in the words "ba<u>n</u>ana" and "<u>N</u>ottingham", try to articulate both words and watch your mouth using a mirror, you will notice that the mouth opened wider in the "ba<u>n</u>ana's" *n* phoneme, because it is affected by the next phoneme "{", while the mouth shape was more circular when articulating *n* in "Nottingham", because it is affected by the next phoneme "o".

5. Some phonemes have weak visual representation, particularly the ones articulated from the inside of the mouth, such as the glottal (h). For example, try to say "Robinhood" and "Robinood" and you will not find much difference.

6. Similar visemes can be associated with different visual signatures due to various pronunciation styles (Yu, 2008).

7. The major difficulty of VSR systems is the identification of visemes (Yu, 2008).

8. Visemes cover only a small subspace of the mouth motions represented in the visual domain (Yu, 2008).

| Hazen et al. (2004) ||
|---|---|
| **Viseme** | **Phonemes** |
| Sil | - |
| OV | [ax], [ih], [iy], [dx] |
| BV | [ah], [aa] |
| FV | [ae], [eh], [ay], [ey], [hh] |
| RV | [aw], [uh], [uw], [ow], [ao], [w], [oy] |
| L | [el], [l] |
| R | [er], [axr], [r] |
| Y | [y] |
| LB | [b], [p] |
| LCl | [bel], [pel], [m], [em] |
| AlCl | [s], [z], [epi], [tcl],[del], [n], [en] |
| Pal | [ch], [jh], [sh], [zh] |
| SB | [t], [d], [th], [dh],[g], [k] |
| LFr | [f], [v] |
| VlCl | [gel], [kel], [ng] |

| Potamianos et al. (2003) ||
|---|---|
| **Viseme** | **Phonemes** |
| Silence | [sil], [sp] |
| Lip-rounding based vowels | [ao], [ah], [aa], [er],[oy], [aw], [hh], [uw],[uh], [ow], [ae], [eh],[ey], [ay], [ih], [iy], [ax] |
| Alveolar-semivowels | [l], [el], [r], [y] |
| Alveolar-fricatives | [s], [z] |
| Alveolar | [t], [d], [n], [en] |
| Palato-alveolar | [sh], [zh], [ch], [jh] |
| Bilabial | [p], [b], [m] |
| Dental | [th], [dh] |
| Labio-dental | [f], [v] |
| Velar | [ng], [k], [g], [w] |

**Figure 2.2**. (left) The 44 English phoneme mapping to 13 viseme (Potamianos et al. 2003), (right) 14 visemes mapped to 53 English phonemes used in Hazen et al. (2004)[*].

---

[*] Different researchers have different phonemes symbols and sets, here we leave them as they are stated in the previous work to show this problem.

9. A viseme model cannot represent transitions between visemes in continuous speech (Yu, 2008).

In order to alleviate the visemes problems, Yu (2008) made visual speech recognition the process of recognizing individual words based on a manifold representation instead of visemes representation. This is done by introducing a generic framework (called Visual Speech Units) to recognise words without resorting to viseme classification.

Yu's (2008) lip-reading system consists of 3 main steps: firstly, lip localization using pseudo hue from RGB colour system, then lips are segmented using histogram thresholding. Secondly, for the features extraction he used the expectation maximization PCA, EM-PCA as described by Roweis (1998), which is a combination of the standard PCA and EM algorithm; EM-PCA was used on the appearance of the lips in order to estimate the maximum likelihood values for missing information. The maximum 3 values were taken for each frame, to form the feature points for a spoken word, which are joined by a plotline based on the frame order to form a surface in the feature space, which is called manifold, each word manifold representing its visual speech units.

Thirdly, Yu (2008) used both K- nearest-neighbour KNN, and HMM to classify the visual speech units sequences. To solve the problem of different lengths of words, Yu used cubic spline interpolation.

Yu's system was evaluated using a small database consisting of 2 speakers, the first one speaking 50 words and the second one 20 words; all the experiments were speaker-dependent, and the study compared the results of the new proposed visual speech units VSU approach with the traditional visemic approach. According to this study, the VSU recognition rate was 80-90%, significantly higher than the recognition rate of the visemes approach (62%-72%).

Yu's interesting study provides promising solutions for the VSR problem in general, and unravels some of the problems of the visemes by proposing the visual speech units (VSU) (the core contribution of his study).

Table 2.4. Summary of automatic lip-reading studies, shows the used feature extraction methods (which include (A-B) appearance-based and (T-B) transform-based), classification methods, and database.

| Study | Features | Classification | Approach | Database |
|---|---|---|---|---|
| Petajan, (1984) | Geometric-based | DTW | Optical flow | small db |
| Bregler and Konig (1994) | A-B + Eigenlips | MLP/HMM | Viseme | Small Germany DB, 6 speakers |
| Belongie and Weber, (1995). | A-B, optical flow | Bayesian | No viseme | small database |
| Neti, et al. (2000). | Hybrid: AAM, DCT/LDA | HMM | N/A | Large Database IBM ViaVoice |
| Potamianos et al. (2003) | T-B: LDA | HMM | Viseme | s-LVCSR |
| Hazen et al (2004). | A-B | HMM | Viseme | AV-TIMIT |
| Gurban and Thiran (2005) | A-B | SVM/HMM | Viseme | Tulips1 |
| Leszczynski and Skarbek, (2005). | Hybrid: DFT, MESH using LDA and PCA | HMM | Viseme | Polish database |
| Saenko, et al. (2005) | A-B | SVM | Viseme | Single speaker from AVTIMIT |
| Arsic and Thiran (2006). | A-B + Eigenlips | HMM | Viseme | Tulips1 |
| Sagheer, et al. (2006) | A-B, HCM | HMM | Viseme | 2 small db (Japanese & Arabic) |
| Yau, et al. (2006) | A-B (MHI + SWT) | ANN | Viseme | Small database |
| Werda, et al. (2007) | Geometric-based | ANN | Viseme | Small database |
| Lucey and Sridharan (2008) | T-B: DCT/LDA | HMM | viseme | CUAVE |
| Yu (2008) | A-B, EM-PCA | HMM | VSU | small database |
| Jun and Hua (2009) | T-B: DCT/LDA | CHMM | No viseme | H II Bi-CAVD |

An important contribution of this study is the feature extraction scheme using EM-PCA, which reduces the dimensionality, and represents the lips motions in the visual domain (Yu, 2008). Yet Yu's study still does not provide a complete solution for the VSR problem, nor a complete solution for the visemes problems for the following reasons:

1. "The Visual Speech Unit concept as proposed represents an extension of the standard viseme model that is currently applied for VSR" (Yu, 2008:IV). So visual speech units proposed by Yu (2008) can be interpreted as visemes, and visemes are also visual speech units, so some of the disadvantages (mentioned earlier) of using visemes in visual speech recognition may still apply for Yu's approach (VSU).

2. "The image data are generated by two speakers and the database is defined only by a limited number of VSUs" (Yu, 2008:110). The small database used by Yu cannot evaluate such a new approach, as all the experiments conducted in Yu's study were speaker-dependent, using a single speaker each time.

3. Describing the factors that contribute to errors in his study, Yu said: "The errors in classification are mostly generated by the errors in registration between the VSU models and continuous manifold" (Yu, 2008:110). This comment reminds us of the major difficulty of VSR systems using visemes, which is the identification of visemes, as the VSU models registration is problematic in Yu's study.

## 2.4 Motivations and Summary

Most of the previous studies on VSR contain promising solutions, especially when combining an audio signal with a video signal. Although most of these systems rely on a clean visual signal (Saenko et al, 2004), still, for visual alone speech reading systems or subsystems, they have a high word error rate WER, sometimes WER being more than 90% for large vocabulary systems (Hazen, 2006), (Potamianos et al., 2003), and a range of 55% to 90% for small vocabulary systems (Yau et al, 2006).

The main reason behind this high word error rate is that automatic lip-reading problems represent a very difficult task by nature, i.e. the visual aspect of speech is information poor, because humans depend mainly on the audio aspect of speech, which is not produced completely from the visual parts of the human speech system such as the labial sounds. On the contrary, speech is produced from different places, starting from deeper places such as the glottal sounds.

Other reasons that increase the word error rate in VSR systems:

1. Large variations in the way that people speak (Yau et al., 2006), even in the same country or community, especially in speaker-independent systems, and it is almost impossible to read some people's lips, because of the way they talk (Bauman, 2000). This result is proved by this study. See chapter 6.

2. Errors in pre-process steps of VSR systems such as face detection and lips localization.

3. Errors in scaling, rotation and pose estimation.

4. Visual appearance differences between individuals, especially in speaker-independent systems.

5. Other general problems like light conditions and video quality.

6. The visemes problems mentioned earlier if the VSR system uses a visemic approach.

These problems, particularly the visemes problem, motivate this study to propose a new VSR system addressing the major reasons for the bad performance of the current VSR systems, trying to obtain a better understanding of the VSR problem, and also achieve better performance.

The new system includes proposing a new method for face and lips localization, and a new hybrid feature extracting method, taking all possible information from the geometry domain (mouth width and height), image appearance domain, and image transform domain.

Eight features were proposed to form eight signals, these signals being weighted depending on their representation strength to represent each word. A score level fusion approach was used to make use of all the eight signals to recognise the related word in the recognition step. The whole approach was termed the "visual words" (VW).

The system was designed and evaluated using a relatively large database and large experiment sets, with up to 83% of word recognition rate in the female speaker-dependent experiment. The evaluation has proved the superiority of visual words over the traditional visemic approach.

# Chapter 3

# Face Localization

Face detection is an essential pre-processing step in many face-related applications (e.g. face recognition, lip reading, age, gender, and race recognition). The performance of these applications depends on the reliability of the face detection step. Also face detection is an important research problem in its role as a challenging case of a more general problem, i.e. object detection.

The face detection problem encompasses several related automated computational tasks including determining if an image does contain a human face? Where is it? How many faces are in the image? The most common and straightforward example of this problem is the detection of a single face at a known scale and orientation. This example is referred to as face localization and assumes that it is guaranteed to find the location of a face in an image.

This chapter is devoted to investigating and developing a face localisation scheme suitable for use as the first pre-processing step for **visual-word** recognition. We shall first review existing approaches to face localisation and develop our own scheme, which consists of a number of steps. We shall then present results of experimental work to test the performance of our scheme and in comparisons with existing algorithms.

## 3.1 Classification of Face Localization Schemes

Face localization is a special case of face detection and assumes that it is guaranteed to find a face(s) in an image, while the face detection algorithm needs to determine whether there is actually a face(s) or not, and locates it/them, if any. This is not a minor problem (King, 2003), and no method has yet been found to solve it with 100% accuracy. Factors influencing the accuracy of face detection include variations in recording conditions/parameters such as pose, orientation, and lighting. Also faces are non-rigid objects that have a high degree of variability with respect to facial expression, occlusion, and aging, although there are several algorithms and methods that can deal with this problem, attaining various accuracy rates under varied conditions. Face localization is no less complicated than face detection; hence most face detection algorithms do not locate the face precisely, in contrast to face localization algorithms (Yang et al., 2002).

Most existing face detection/localisation schemes are based on somewhat restrictive assumptions such as there is only one face, upright, looking at the camera, etc. Some of the most successful methods use a 20×20 (or so) pixel observation window (as a matching template) to scan the image for all possible locations, scales and orientations. These methods include the use of support vector machines (Osuna et al., 1997), neural networks (Rowley et al., 1998), or the maximum likelihood approach based on histograms of feature outputs (Schneiderman and Kanade, 2000). In an extensive review of face detection schemes, Yang et al. (2002) identifies four major approaches to face detection in still images:

1) Knowledge-based methods.
2) Feature invariant approaches.
3) Template matching methods.
4) Appearance-based methods.

## 3.1.1 Knowledge-based methods

These methods use human knowledge about the face, such as: what does it look like? What are the components? What are the relationships between facial features? For example, there are two eyes in the upper part of the face, one mouth in the centre and lower part of the face, and one nose located in the centre between the eye line and the mouth, etc. Such information can help to eliminate a large number of spurious hits in the early stages and then rigorous testing need only be applied to a relatively small number of possible locations of a face in an image.

Yang and Huang (1994) used a hierarchical knowledge-based method, where different resolution mosaic images were created using sub-sampling and averaging. Their method consists of three levels of rules to detect faces. All possible face candidate locations are found by scanning a window over the input image and applying a set of rules at each location and at each level. The rules at a higher level are general descriptions of what a face looks like (two eyes, mouth etc.) while the rules at lower levels rely on details of facial features. For example, the rules at level 1 can be "the centre part of the face has four cells with a basically uniform intensity," "the upper round part of a face has a basically uniform intensity," and "the difference between the average grey values of the centre part and the upper round part is significant." (Yang and Huang, 1994).

The face candidates are then verified by the existence of prominent facial features using local minima at higher resolutions. At Level 2, local histogram equalization is performed on the face candidates, followed by edge detection. Surviving candidate regions are then examined at Level 3 with another set of rules. This system was evaluated using a set of 60 images where faces were located in only 50 images (Yang and Huang, 1994).

The problem with this approach is that it is difficult to define all the possible rules using only human knowledge, i.e. if the rules are so strict, some faces will not be detected, and if they are too general, something else may be identified as a face.

### 3.1.2 Feature invariant approaches

Feature invariant schemes assume that for all face patterns there are certain spatial and frequency characteristics of the image, which are common and possibly unique to all faces, even under different imaging conditions (Sung and Poggio, 1998).

The aim of these approaches is to locate faces using features that exist and are unchanged for all faces regardless of the variations in the face pose, viewpoint, or lighting conditions. These methods are based on the observation that humans are capable of locating structural features in images even under recording conditions/factors that have an adverse impact on the performance of the face detection process, such as variations in pose and lighting conditions. There doesn't seem to be a universal agreement on the nature of the targeted features, but they mostly target facial features (such as eyes, nose and mouth), skin colour, texture, or even a combination of these. The main idea behind these approaches is to first detect the common and possibly unique features, and if they do actually exist, the method will then reveal the location of the face.

The work of Graf et al. (1995) is an example of a feature invariant method to locate facial features and then faces. This method starts with image pre-processing and morphological operations to enhance shapes' regions, adaptive threshold values are selected (depending on a histogram of the processed image) to generate two binary images, and related components are identified in both images to recognize the areas of candidate facial features. A classifier then evaluates the combination of these areas.

The use of skin colour to detect faces includes the work of Garcia and Tziritas (1999) and Zhang et al. (2009). The latter proposed a method based on RGB colour centroids segmentation, which is used for colour image thresholding. The coloured image is segmented to face and non-face regions twice, and the first segmented binary image results from an automatic threshold, which is gained from histogram analysis of the face image. The second segmented image results from a non-linear logarithmic thresholding formula applied to the grey scale of the original image, and the two segmented binary images are then ANDed to obtain the best candidate face regions. The face is then tracked and located in these areas using a specific aspect ratio and a specific size range (Zhang et al., 2009).

This method was evaluated using only 9 images with 27 faces, single and multi faces. The segmentation results of this work were compared with some other segmentation methods showing better performance. This method was not evaluated using a large database under different lighting conditions.

Human faces can be separated from other objects using their distinctive texture; this property was used to detect faces in images by some researchers including Augusteijn and Skujca (1993) and Dai and Nakano (1996). The latter reported an excellent detection rate of 60 faces in 30 images.

Sobottka and Pitas (1996) used a multiple features approach (face shape and colour information) to detect faces. By applying colour segmentation to locate candidate face regions, the best fit ellipse was computed for each region, and those regions that were well approximated by the ellipse were chosen as face candidates. These candidates were then verified by searching for facial features such as eyes and mouth by applying morphological operations and minima localization to intensity images. Their experiments had a detection rate of 85% based on testing 100 images.

Although skin colour and texture are common and possibly unique to all faces, they are also vulnerable to extreme variations in lighting conditions, shadow lines and camera characteristics, but by using an adaptive method, colour detection can be more robust against such conditions (Hjelmas and Low, 2001).

### 3.1.3 Template matching methods

Such methods entail using one or more patterns (templates) that reflect a typical face, then scanning this pattern over the targeted image to find the best correlation between the pattern and a window in the image. These patterns can be pre-defined templates or deformable templates.

The work done by Nallaperumal et al. (2006) fits into this category. Their method segments the image depending on the colour of the skin, using an adaptive threshold to separate regions of candidate faces from background. The segmented regions represent possible locations for the face. These regions are pre-processed to be ready for the next step; this includes region counting, centres computing, size and orientation. The method then selects the most likely regions in which the face(s) might be located. The second step incorporates the use of a predefined template to match with all the candidate regions, after resizing and rotating the template according to the segmented areas. The template-matching process verifies whether there is a face in that area. Their method was evaluated using 30 images, faces were detected in 25 images, and they reported 83% detection rate when using template matching.

Other work that uses predefined template matching for face detection/localization include that of Sakai et al. (1969) and Miao et al. (1999). Of those who use deformable template matching, the work of Yuille et al. (1992), Kwon and Da Vitoria Lobo (1994) and Lam and Yan (1994) is significant.

The problem with this approach is that it is difficult to detect faces at different scales, poses and shapes. Deformable templates are used to solve such problems (Yang et al., 2002). In cases where the faces are upright and approximately the same size, this method is preferable for its simplicity and speed.

### 3.1.4 Appearance-based methods

This approach is similar to the previous approach in that it is based on using matching templates, but the templates are not specified in advance. Rather they are learned through a training process that uses a given representative set of images, which are then used for detection. Researchers have used two approaches for appearance-based methods, the first of which is the probabilistic framework, which utilizes the class-conditional density functions p(x|face) and p(x|non-face), where x is the feature vector.

Bayesian classification or maximum likelihood is normally used to classify a candidate location as face or non-face. This approach includes several methods such as the Hidden Markov Model (HMM), Naïve Bayes classifier and Information-theoretical approach.

The second approach is to find a discriminant function such as threshold, separating hyper-plane or decision surface, to determine a candidate location as face or non-face. This approach includes several methods such as Distribution-based methods, Eigenfaces, Support vector machines (SVM), and Neural networks.

### 3.1.4.1 Distribution-based methods

This approach tries to answer the question: What is the best distribution of face patterns? Sung and Poggio (1998) introduced a distribution-based face detection method, in which they used 1000 face images to create a distribution-based generic face model with all its allowable pattern variations in a high dimensional vector space. Then they trained a decision procedure on face and non-face examples, to obtain a set of operating parameters and thresholds that can classify input patterns to face and non-face patterns.

Their method starts with a window of 19 x 19 and scaled gradually to 100 x 100 pixels, to scan the image horizontally and vertically, searching for face patterns. This method was evaluated using 2 data sets, the first consisting of 301 high quality face images of 71 different subjects. 96.3% of all the faces were found using this method. The second database consisted of 23 images (with wide variation in image quality) with a total of 149 faces, and achieved 79.9% detection rate.

One problem of this approach is that when face patterns are affected by changes in view and lighting conditions, their distribution is highly non-linear and complex in any space linear to the original image space (Li et al., 2001). To solve this problem, Li et al. (2001) proposed a non-linear mapping face detection method by which multi-view face patterns in the input space are mapped into invariant (to both illumination and view) points in a low dimensional feature space.

### 3.1.4.2 Eigenfaces methods

Eigenfaces is a well-known Principle Component Analysis (PCA); face images can be reasonably represented by their projection onto a small number of basis images or feature space (low dimensional subspace), called "face space", defined by the Eigenvectors that are derived by finding the most significant Eigenvectors of the covariance matrix for a group of training face images. Significant Eigenvectors are the best account for the distribution of face images within the entire face space. This method captures the variation between the group of faces without emphasis on any one facial region like the mouth or eyes (Mathew, 2004). Eigenfaces are shown in Figure 3.1.

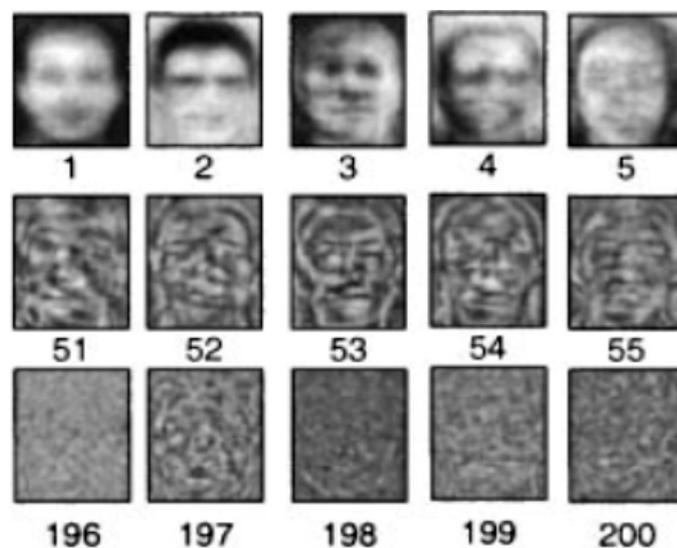

**Figure 3.1**. Some examples of Eigenfaces; the number below each image indicates the principal component number, ordered according to Eigenvalues (Hjelmas and Low, 2001).

Turk and Pentland (1991) detected faces using knowledge of the face space; images of faces do not change radically when projected into the face space, while the projection of non-face images appears fairly dissimilar. Using this fact, at every location in the image, the distance between the sub image and the face space is calculated. The distances at each location form a face map, where the dark areas (small distances) indicate the presence of face. The performance of this method was reported to be 94%, with 6% false positive in a database of 7562 frontal face images on a plain background. More recent Eigenfaces work includes that done by Monwar et al. (2006). Eigenlips, which is similar to Eigenfaces, is discussed in more detail in chapter 4.

**3.1.4.3 Support vector machines methods**

A support vector machine (SVM) is a technique for separating data points into classes by constructing a hyperplane in a high-dimensional space; a good separation is achieved by the hyperplane that has the largest distance to the nearest training data points of any class. This optimal hyperplane is defined by the support vectors, which are a small subset of the training vectors near the decision boundary. The optimal hyperplane can be estimated as a linearly constrained quadratic programming problem solution. A separating hyperplane with larger margin is more capable for better generalization. See Figure 3.2.

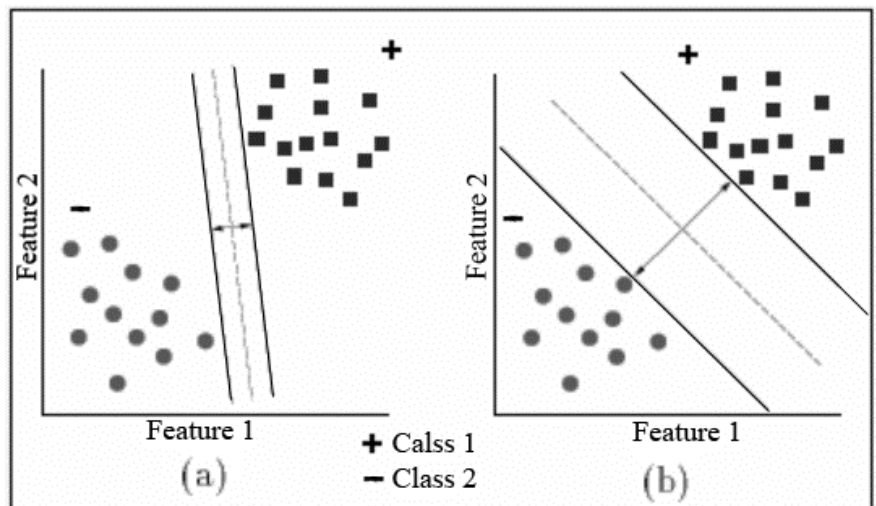

**Figure 3.2,** (a) Separating hyperplane with small margin. (b) Separating hyperplane with large margin (Osuna et al., 1997).

SVMs were first applied to face detection by Osuna et al. (1997). They trained a SVM with a 2nd-degree polynomial as a kernel function with a decomposition algorithm using a bootstrap-learning algorithm. Their method starts traditionally with scanning input images with a 19 x 19 window, pre-processing the window using masking, light correction and histogram equalization, and then classifying sub-images to faces or non-faces using the trained SVM.

This method was evaluated using 2 image sets. Set A contained 313 high quality images, with one face per image, and set B contained 23 different quality images with 155 faces. The reported results were 97.1% and 74.2% detection rate for sets A and B respectively. More recent study of face detection using SVM includes the work of Romdhani et al. (2004) and Waring and Xiuwen (2005).

**3.1.4.4 Hidden Markov Model methods**

Hidden Markov Models (HMM) are statistical models consisting of a number of nodes that represent hidden states, these nodes being connected by links recounting the conditional probabilities of the transitions between these states. HMM are discussed in detail in chapter 6, section 7.

Samaria (1994) proposed a face detection and recognition method using HMM, by dividing the face into horizontal sub-regions (strips of pixels); forehead, eyes, nose, mouth and chin (see Figure 3.3). Each region is a state in the trained HMM. The face is detected in a specific location if all these regions are detected in appropriate order (from top to bottom and left to right) and if the face likelihood obtained for each strip pattern in the image is above a threshold.

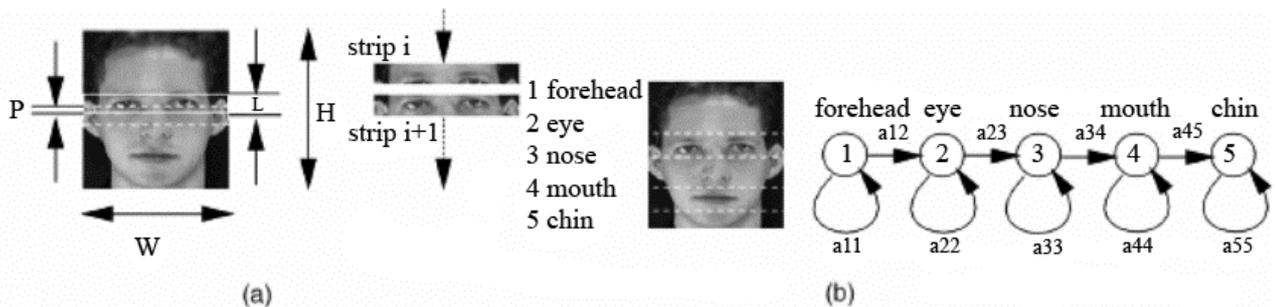

**Figure 3.3**, HMM for face detection. (a) Observation vectors; each vector is constructed from a window of W x L pixels with P pixels overlap. (b) Hidden five states trained with sequences of observation vectors (Samaria, 1994).

Samaria's (1994) face detection method was evaluated using 400 frontal face images, and the reported detection rate was 85%. Nefian (1999) proposed a face detection and recognition method using embedded HMM (or pseudo 2 dimensional HMM), where each state is one dimensional HMM on its own, called "super state". Each super state models a part of the face, starting with the forehead and ending with the chin. 2d-HMM is useful to model data from images, because images are 2 dimensional structures in nature. Nefian's (1999) method was evaluated using 432 frontal face images, and his reported detection rates were in the range of 91-96%.

**3.1.4.5 Naïve Bayes classifier methods**

This classifier is a simple probabilistic classifier based on applying Bayes's theorem with strong independent assumptions, i.e. it assumes that the presence or absence of a particular feature is unrelated to the presence or absence of any other feature. The posterior probability function gives the probability that the object is present given an input image.

Schneiderman and Kanade (1998) used this approach for their face object/face detection method; they chose a functional form of the posterior probability that captures the joint probability of multi-scale local appearance and position of face sub-images. At each scale, a face image is divided into four sub-areas, which are projected onto a lower dimensional space using PCA to form a finite set of face patterns. The probabilities of each sub-area are estimated from the projected samples. The method decides that a face is present when the likelihood ratio is larger than the ratio of prior probabilities.

Naïve Bayes classifier works well for this method because the face regions are not dependent on each other. The method can also detect faces with some rotated angle (up to 22.5°) and profile faces. Schneiderman and Kanade (1998) reported a 93% detection rate of 483 upright faces on 125 images. A more recent approach to face detection using Naïve Bayes classifier is the work of Park et al. (2005).

### 3.1.4.6 Information-theoretical approach methods

These methods depend on information theory, which involves the quantification of information, including entropy, joint entropy, conditional entropy, mutual information, Kullback relative information, etc. Information theory methods provide algorithms with the amount of information that can be seen in an image to be used for face detection, for instance.

Hotta et al. (2000) used information-theoretical attention points, which are selected using Gabor filters applied to contrast filtered image. The information from the Gabor features is then used to construct a saliency map, which indicates the attention candidates. These candidate regions are matched using a similarity function that uses correlation between the Gabor features of a model face image and the input image (saliency map). This method was evaluated using 218 images of the **same person**, in different scale and lighting conditions, and their reported face localization rate was 99.5% using one image of the same person as a model.

### 3.1.4.7 Inductive learning methods

Inductive learning methods can be defined as those methods that systematically produce general domain knowledge from the specific knowledge provided by domain examples (the training set).

Huang et al. (1996) used inductive learning for face detection. Their method starts with finding a candidate region for the face to reduce searching time, using histogram equalization, edge detection and profile analysis. A decision tree (DT) was used to detect whether there is a face in the candidate region or not. The DT was trained (learned) on 30 features of faces and non-faces in an 8 x 8 pixel window. These features include entropy, mean and standard deviation.

As inductive learning requires symbolic data, each one of the positive examples corresponding to the face region is tagged "CORRECT", while non-face regions are tagged "INCORRECT". DT is induced by taking a set of positive examples and a set of negative examples and building a classifier, as a DT structure consists of *leaves*, indicating class identity (CORRECT or INCORRECT), and *decision nodes* that are devoted to some test to be carried out (decision to be made) on a single feature, with one branch for each possible outcome (Huang et al., 1996).

Starting at the root of the tree and moving through it until reaching a leaf, at each non-leaf node a decision is made using an optimal entropy criterion. This system was evaluated using 2,340 frontal face images, and its reported accuracy rate was 96%.

**3.1.4.8 Neural networks methods**

An Artificial Neural Network (ANN) is a computer structure that provides a statistical analysis procedure in which processors are connected in the same manner as the nervous system for learning in animals. ANN can learn by trial and error and classify large and complex data sets by grouping cases together in a way similar to the human brain. ANN methods have been used extensively for pattern and object recognition, the one proposed by Rowley et al. (1998) for face detection being a good example of ANN methods.

Rowley's et al. (1998) method for face detection consists of two stages (see Figure 3.4), consisting of first applying a neural network-based filter that receives a 20 x 20 pixel region of the input image, and output values ranging from -1 to 1, which means non-face or face respectively. Assuming that faces in an image are upright and looking at the camera, the filter is applied to all locations in the image, to obtain all the possible locations of the face. To solve the scale problem, the input image is repeatedly sub-sampled by a factor of 1.2. The input image is pre-processed before inputting the proposed system. Light correction and a histogram equalizer were used to equalize the intensity values in each window.

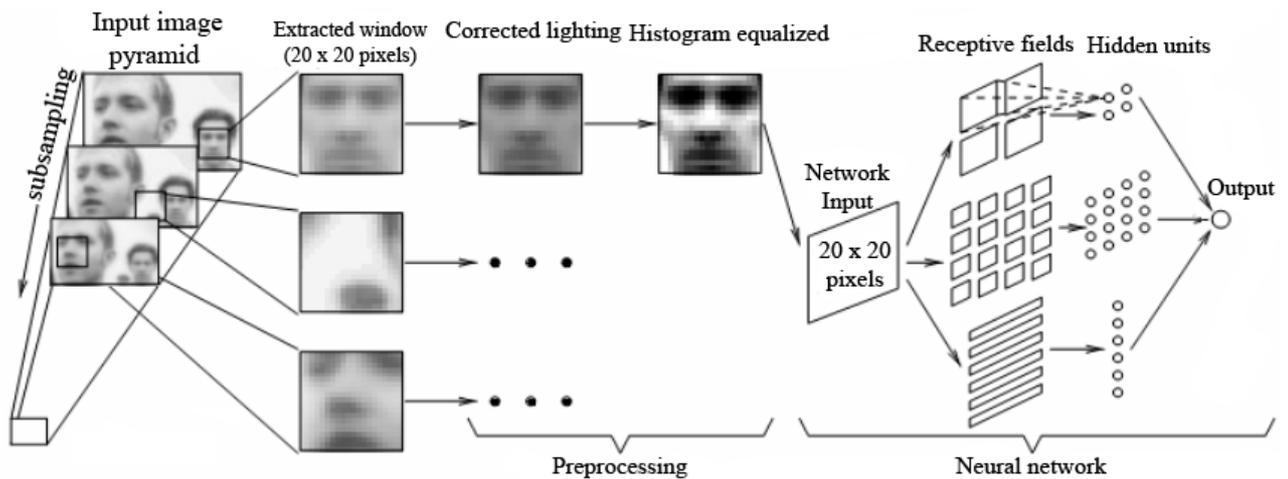

**Figure 3.4**. Rowley's et al. (1998) basic algorithm for face detection

The second stage of Rowley's method is merging overlapping detections and the arbitration process. The same face is detected many times with adjacent locations, the centre of these locations being considered as the centre of the detected face, and if two face locations are overlapped, the one with the highest score is considered the face location. Multiple networks were used to improve detection accuracy by ANDing (or ORing) the output of two networks over different scales and positions.

Rowley's system was evaluated using 130 images containing 507 faces, the images having been collected from newspaper pictures, photographs and the World Wide Web. To train the system on false examples, 1000 images with random pixel intensities were generated. The detection rate of this system ranged from 78.9% - 90.5% depending on the arbitration used (ANDing, ORing).

Hjelmas and Low (2001) classified face detection methods into two approaches: feature-based and image-based, both approaches being covered by Yang's et al. (2002) classification.

As a result of the aforementioned researches in face detection/localization methods, some online applications for face detection problems started to appear*. Despite the large number of face detection methods in the literature, few have been implemented as open source software, to be used in the proposed visual speech system. However, the Rowley's et al. (1998) system is among the exceptions and is available for public use. Moreover, as a general-purpose face detection scheme, Rowley's scheme is by far one of the best performing. However, when we applied this scheme on the PDA database (that consists of a very large number of video clips of persons uttering speech (Morris et al., 2006)) to detect the faces, its performance dropped by a noticeable percentage.

---

* http://demo.pittpatt.com and http://www.idiap.ch/~marcel/demos.php

This drop in performance, in comparison to its reported performance on other databases, can be attributed to the large variation in lighting conditions in the recorded PDA video clips. Thus, the motivation for designing a new face detection/localization method arises not from a need to replace existing methods, but to detect/localize faces, and facial features such as lips, under variations in lighting conditions, as is the case of visual speech recognition on mobile phones. Indeed, we used the Rowley scheme for face detection on a database of videos recorded with HD cameras under well-controlled conditions, and it performed very well.

## 3.2 The proposed face localization method

Existing visual speech recognition schemes rely on measuring movements of the lips and the appearance of the mouth. Because most of the information related to VSR comes from the lips and the mouth area in general, it is important to detect lips first, to be able to extract information about them. As the lips are the most deformable part in the face (especially while talking), and the fact that the face has more stable features (eyes, eyebrows, cheeks, chin and nose) than the lips (they are stable in the sense that their appearance does not change much while talking), it would be much easier and more accurate to detect the face first, and then to use some known facts about the relative position of the mouth region within the face region to detect the lips more accurately.

The aim of this chapter is to design and evaluate a new face detection/localization method. The proposed method will be a hybrid of the knowledge-based approach, template matching approach and feature invariant approach (skin colour). The system will be evaluated using the PDA database, which contains thousands of video clips of 60 subjects uttering 18 different categories of speech in 4 different indoor/outdoor lighting conditions. The size of the database images was 240 x 320 pixels.

The new proposed method assumes the following:

1. It is guaranteed to have only one face in each image. Because there is only one face in each frame in the PDA database, there is no need for this method to try to find any other faces. This method is called face localization, and not face detection, hence the face is there and the method needs to localize it.

2. The faces sizes in the PDA database are around 100 x 100 pixels. A fixed size template is adequate, so there is no need for template scaling.

3. The face is reasonably upright, allowing little variation in rotation around the X, Y or Z-axis, and that is the case in the PDA database.

The proposed method is divided into several stages, as shown in Figure 3.5: Capturing an image, pre-processing it, finding the most distinguished features using edge detection, refining the output image, applying Haar wavelet to level 3, scanning low-low $LL_3$ sub-band using a pre-selected face template on the areas in the image that best fit the template. The highest ranked potential faces regions are processed using colour template information, and finally the scores from the $LL_3$ template and the colour template are fused using fuzzy decision criteria to determine the best face location in the original image. In the case of videos, the previous face location can be used to track the face in the next frame, instead of scanning the whole image, thus reducing computation time.

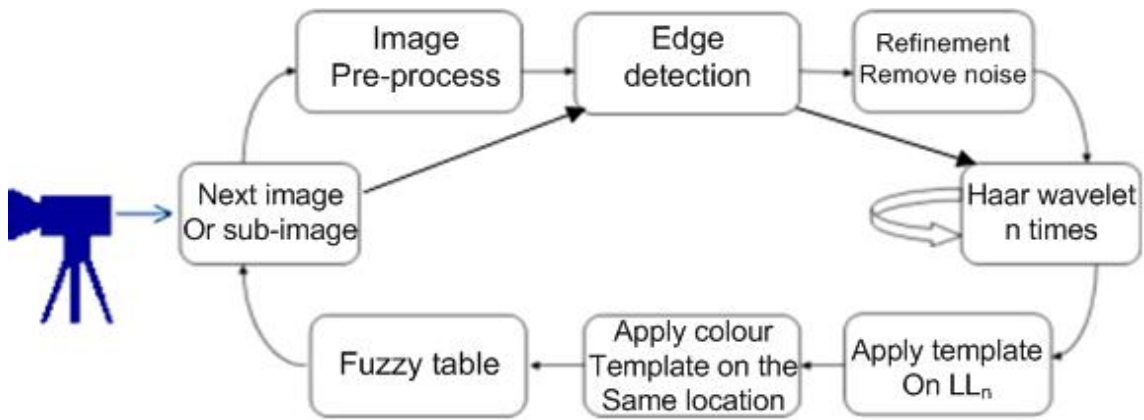

**Figure 3.5**. The proposed multistage method for face localization.

Most existing face detection methods use the first stage, "*image pre-process*", to enhance the image for sharper edges and to reduce noise and/or unimportant information. This would increase the reliability of face detection. In our method we skip this stage because it is time consuming. Moreover, we designed a special edge detection procedure to take into consideration that we only need edges that are related to facial features.

## 3.2.1 Edge detection

The Sobel edge detection method (Sobel, 1968, 1990) is the most commonly used edge detection procedure, due to its simplicity. This method depends mostly on two 3 x 3 filters – one to detect the horizontal edges and the other to detect the vertical ones. The disadvantage of this method and many other edge detection methods is that the finest edges of the undistinguishing parts of the image are included, such as face tissue, which do not need to be tracked, because they are greatly affected by illumination, and they are different from one image to another. For example, old people's face images differ from younger people's face images mainly because of the appearance of wrinkles. Moreover, face tissue and hence small edges are greatly affected by shadow and light conditions (see Figure 3.6). For face and lips region detection, we are only interested in edges defining these features. The key idea is to get rid of texture, and detect the relevant edges in the face region. The Shannon entropy function defined by equation (1) (Shannon, 1948) can be used to distinguish the homogeneous data from the heterogeneous data (see Figure 3.6 and equation (1)):

$$H(X) = -\sum_{i=1}^{n} p(x_i) \log_2 p(x_i) \qquad (1)$$

where $H(X)$ is the entropy value of a random variable $X=\{x_1,\ldots,x_n\}$ and $p(x_i)$ is the probability of the $i_{th}$ element in X.

To find the edges (face details) in an image, the above equation (1) is applied for each pixel, by taking X (in equation (1)) to be the 5 x 5 window, which is centred at the position of that pixel. In this case each pixel is classified (as being an edge or non-edge pixel) using equation (2) with an empirically determined threshold $\theta$.

$$\text{Binary image output} = \begin{cases} 0 & (edge) \quad H(X) \geq \theta \\ 1 & (tissue) \quad Otherwise \end{cases} \qquad (2)$$

However, this approach is time consuming, and it is difficult to choose a suitable threshold. Instead, an adaptive approach by using two new 5 x 5 filters is proposed. Both filters were designed based on both Sobel and the entropy function equation (1), and the values of the filters were chosen empirically using several face images from the PDA database.

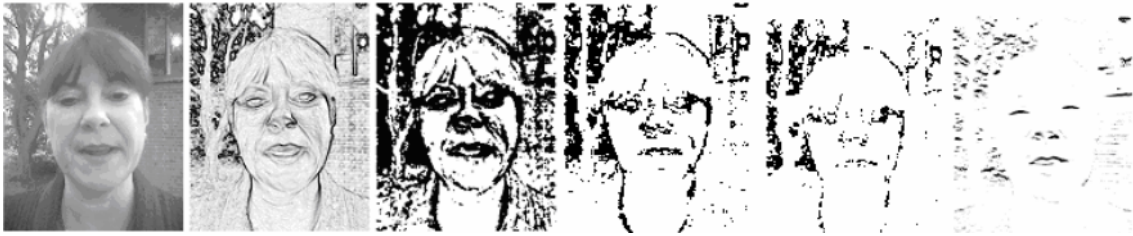

**Figure 3.6**. From left to right, the original image, negative image using Sobel edge detection, the third, fourth and fifth images are the output of edge detection using entropy information on 5 x 5 neighbours with *θ*=3,3.5 and 3.75 respectively, and the last image is the output of edge detection using the proposed filter.

| -1 | -1 | 0 | -1 | -1 |
|---|---|---|---|---|
| -2 | -2 | 0 | -2 | -2 |
| 0 | 0 | 0 | 0 | 0 |
| 2 | 2 | 0 | 2 | 2 |
| 2 | 2 | 0 | 2 | 2 |

| -1 | -1 | 0 | -1 | -1 |
|---|---|---|---|---|
| -2 | -2 | 0 | -2 | -2 |
| 0 | 0 | 0 | 0 | 0 |
| 2 | 2 | 0 | 2 | 2 |
| 1.5 | 1.7 | 0 | 1.7 | 1.5 |

**Figure 3.7**. The coarse filter (on the left), and the fine filter (on the right).

As can be seen from Figure 3.7, the filter on the left is coarse enough to avoid noise, tissue and textures. The second filter on the right is fine enough to detect important facial features, like eyes and nose, even for the highly illuminated regions. Therefore the first filter is used in darker regions of the image, and the second in the lighter regions. For obvious reasons, the choice of the filter is based on comparing between the local average in the designated 5 x 5 window and the global pixel value average. In other words, if the local average is less than or equal to the global one, then the first filter is used; otherwise the second filter is used. In Figure 3.8 below, the first filter would have been chosen on the left side, and the second would have been applied towards the right side.

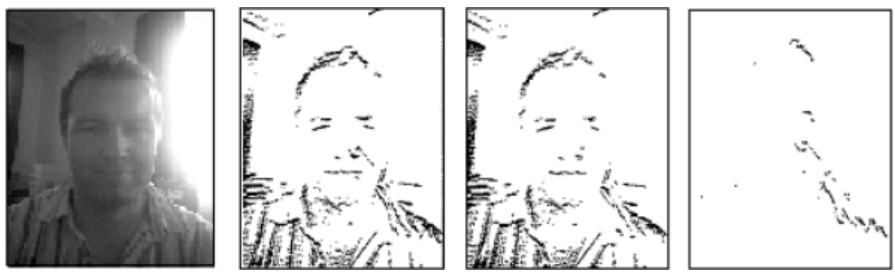

**Figure 3.8**. Edge detection from left to right, the original image, applying both filters using local information compared to global information, applying both filters using global information only, and the difference between the two approaches.

The proposed filters remove many details from the image, mostly unwanted (non-edge pixels), and they keep only the most distinguishing details, even for images with extremely light variation. Note that the numbers on the last row in the coarse filter: {2, 2, 0, 2, 2}, and the fine filter: {1.5, 1.7, 0, 1.7, 1.5}, are bigger than the upper row: {-1,-1,0,-1,-1}. When either of these filters is applied, if the pixels in the window were homogeneous (i.e. no big differences between their intensity values), we get a high value (near 255, i.e. white value). Otherwise small values are neighbouring large values, and the output of the filtering will either be: 1) a high value normally greater than 255 if the 2s were multiplied with the larger pixel values, or 2) a small value normally less than zero if the 2s were multiplied by the smaller ones.

This means that the given pixel will be replaced with a value that is significantly different from the ones above it, i.e. it is more likely to belong to a horizontal feature. Thus, this filtering favours horizontal features. This is desirable since most facial features are horizontal features.

One of the major problems of using such filters is that the output image after applying both filters is not a binary image (details and non-details); the apparent grey pixels are neither details nor tissue, so these pixels are not yet decided and need to be addressed in subsequent steps.

## 3.2.2 Refinement

This step is used in most general purpose edge detection algorithms to clean the image from the side effects such as tiny edges, lines and/or corners. This step can be thought of as a noise removal step. So this step may include salt and pepper noise removal, Gaussian noise removal, and both noises can be reduced/removed using linear or non-linear filters. This step may also include some morphology-based operations such as erosion and dilation. Erosion is the operation that causes objects to shrink, while dilation causes objects to grow in size. The next figure illustrates both operations.

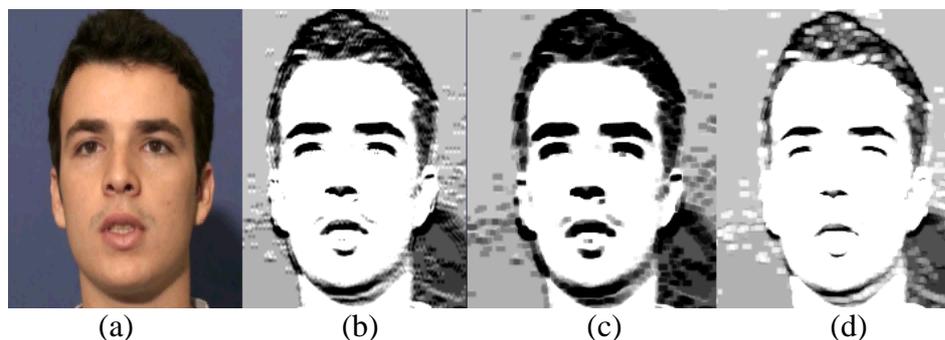

(a)  (b)  (c)  (d)

**Figure 3.9**. Illustrates erosion and dilation operations, a) is the normal face image, b) the output edge detection- noise is already removed, c) 3x3 facial features dilation, and d) 3x3 erosion.

The use of the two filters in the above-mentioned edge detection method removes the need for this step (the refinement). See Figure 3.9b.

### 3.2.3 The wavelet transform step

Wavelet transforms (WT) are widely accepted as an essential tool for image processing and analysis, including image/video compression, watermarking, content-base image retrieval, face recognition, and texture analysis. They are also an alternative for short-time analysis of quasi-stationary signals, such as speech and image signals, in contrast to the traditional short-time Fourier transform. The Discrete Wavelet Transform (DWT) is a special case of WT, providing a compact representation of a signal in the time and frequency domain.

In particular, wavelet transforms are capable of representing smooth patterns as well as anomalies (e.g. edges and sharp corners) in images. There are a number of different wavelet image decomposition schemes, the most commonly used being the pyramid scheme, which is adopted in this study.

At a resolution depth of k, the pyramidal scheme decomposes an image I into 3k+1 frequency sub-bands ($LL_k$, $HL_k$, $LH_k$, $HH_k$,..., $LL_1$, $HL_1$, $LH_1$, $HH_1$). The lowest-pass sub-band $LL_k$ represents the k–level resolution approximation of image I, while the other high frequency sub-bands highlight the significant image features on different scales. The Haar wavelet transform is used in this study mainly to reduce the dimensionality of the image as much as possible, while maintaining information as much as possible, basically to enhance performance, and to increase the speed of detection, so the method can then be used on mobile devices and/or in online detection in videos, etc. Accordingly, we will use the $LL_3$ sub-band of the output image from the edge detection process.

The size of each image in the PDA database is 240 x 320. If a window or template of 20 x 20 is used to scan such an image, looking for a face, the computational complexity would be quadratic (i.e. 240*320*20*20= 30720000 for each video frame). But with the wavelet decomposition to the third level, the size of the image is then reduced by a factor of 64 to 30 x 40 (the quarter of the quarter of the quarter of the image). Although the $LL_3$ may not maintain all the information, it is still adequate for face detection.

The values of the feature pixels in the output image after the edge detection method are zero or approaching zero, while the non-feature pixel values are 255 or approaching 255 (see Figures 3.6, 3.8 and 3.10). The LL sub-band of such an image would consist of black pixel coefficients, white pixel coefficients, and some unclassified grey values around the black coefficients. $LL_3$ can be quantized as a binary image without losing significant information, by comparing each coefficient with the local average in a 5 x 5 window using the following equation:

$$F(x, y) = \begin{cases} 0 & f(x, y) \leq \text{Local average} \\ 255 & \text{otherwise} \end{cases} \quad (3)$$

where $f(x,y)$ is the wavelet coefficient, and $F(x,y)$ is the new binary image value. Figure 3.10 illustrates the process of getting a binary $LL_3$ from an original image.

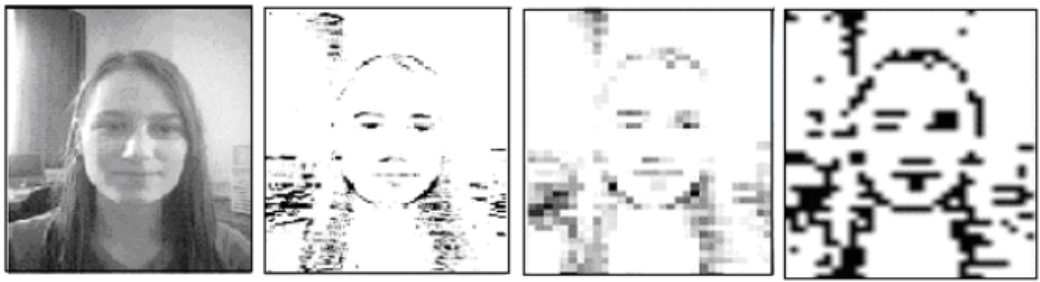

**Figure 3.10**. From left to right: the original image, output of edge detection, $LL_3$, and the binary $LL_3$ (the last two images were stretched for illustrative purposes).

Resizing the image is faster than applying wavelet, but resizing does not preserve much data, especially after the proposed edge detection method (see Figure 3.11d). Other non-LL sub-bands such as $LH_3$ do not provide a structural facial shape, which makes it difficult to apply any kind of template to detect the face (see Figure 3.11e), unlike the LL, which provides adequate information (see Figure 3.11c). Thus, LL was chosen for our face detection method.

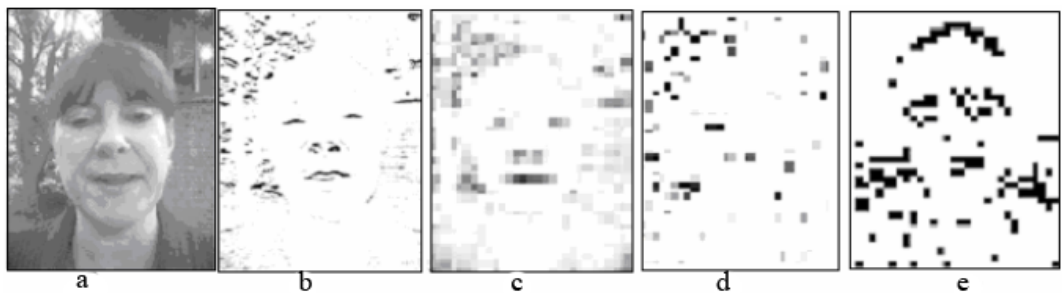

**Figure 3.11**. Applying wavelet and resizing: a) original image, b) edge detection, c) $LL_3$, d) resized to the size of $LL_3$, e) binary image of $LH_3$

### 3.2.4 Applying face template on $LL_3$

Having obtained the $LL_3$ binary image, it can be scanned and a template applied to localize the face box. The face box is the rectangular area of a human face that starts from the left eyebrow to the right eyebrow in width, and down to the horizontal line that touches the bottom of the mouth in height.

For most of the PDA database, faces sizes are nearly 100 x 100 pixels. Accordingly, a fixed size template of 13 x 13 pixels is chosen to be applied to the $LL_3$ binary image, so that when it is scaled up to become 13 x 8 in width and 13 x 8 in height, it almost fits all faces in the PDA (see Figure 3.12).

The binary template shown in Figure 3.12 is obtained by taking the pixel-by-pixel mean values of 100 random binary $LL_3$ face image samples. The mean was computed then rounded to 0 (black) or 1 (white), and the zeros used to detect the details: the left eye, right eye, nose, and the mouth. When the image is transformed to the third level, the eyebrows will be attached to eyes, forming one darkish area.

The 1s represent the white area of the face, where there are no details, and the 2s on the bottom corners are for the neutral pixels; the shape of the face is elliptic, and sometimes the corners of the chin are detected and sometimes they are not, therefore the 2s were considered as neutral pixels so that they would not affect the results.

| 1 | 1 | 1 | 1 | 1 | 1 | 1 | 1 | 1 | 1 | 1 | 1 | 1 |
|---|---|---|---|---|---|---|---|---|---|---|---|---|
| 1 | 0 | 0 | 0 | 0 | 1 | 1 | 1 | 0 | 0 | 0 | 0 | 1 |
| 1 | 0 | 0 | 0 | 0 | 1 | 1 | 1 | 0 | 0 | 0 | 0 | 1 |
| 1 | 0 | 0 | 0 | 0 | 1 | 1 | 1 | 0 | 0 | 0 | 0 | 1 |
| 1 | 1 | 1 | 1 | 1 | 1 | 1 | 1 | 1 | 1 | 1 | 1 | 1 |
| 1 | 1 | 1 | 1 | 1 | 1 | 1 | 1 | 1 | 1 | 1 | 1 | 1 |
| 1 | 1 | 1 | 1 | 1 | 1 | 1 | 1 | 1 | 1 | 1 | 1 | 1 |
| 1 | 1 | 1 | 1 | 1 | 0 | 0 | 0 | 1 | 1 | 1 | 1 | 1 |
| 1 | 1 | 1 | 1 | 1 | 0 | 0 | 0 | 1 | 1 | 1 | 1 | 1 |
| 1 | 1 | 1 | 1 | 1 | 1 | 1 | 1 | 1 | 1 | 1 | 1 | 1 |
| 1 | 1 | 1 | 0 | 0 | 0 | 0 | 0 | 0 | 0 | 1 | 1 | 1 |
| 1 | 1 | 1 | 0 | 0 | 0 | 0 | 0 | 0 | 0 | 1 | 1 | 1 |
| 1 | 1 | 1 | 1 | 1 | 1 | 1 | 1 | 1 | 1 | 1 | 1 | 1 |

**Figure 3.12**. The template used on the binary image of the $LL_3$ to detect a face.

The proposed method uses a window of 13 x 13 to scan the $LL_3$ binary image from the top left corner, to the bottom right corner, searching for a face (one face). The Hamming distance/score (HD) is used as a measure of difference between the window and the template, and is calculated as in the following algorithm:

1- Initialize, x = 0, y = 0, max = 0

2- i = 0, j = 0, *HD* = 0

3- Add the $LL_3$ window content to the template content pixel-by-pixel.

4. If ($window(x+i,y+j) + Template(i,j) = 0$ or $256$) increment *HD*.

5. Otherwise increment the missed counter, i.e. no matching.

6. i=i+,j=j+1

7. Repeat steps 3 - 6 for all contents of the window.

8. IF max < *HD* THEN max = *HD*, *XX* = *x*, *YY* = *y*

9. x=x+1, y=y+1

10. Repeat steps 2 - 9 until x = width and y = height of $LL_3$ binary image.

11. Return *(XX,YY) as the location of the face box.* Scale it to fit the face on the real image.

Alternatively, the best location of the face using the Hamming distance function can be calculated as follows:

$$(X,Y) = \underset{\substack{x=0,\, y=0}}{\arg\max}^{width,\,height} \left( \sum_{i=0}^{12} \sum_{j=0}^{12} \begin{cases} 1 & w(x+i, y+j) + T(i,j) = 0 \text{ or } 256 \\ 0 & \text{Otherwise} \end{cases} \right) \quad (4)$$

where *X,Y* is the point of the upper left corner of the face box (face location), *w(x,y)* is the window's content (pixel values from $LL_3$ binary image) starting from (x,y) location, *T(i,j)* is the template contents (see Figure 3.12), *i* and *j* are indices to cover 13 x 13 window, width and height are the width and height of the $LL_3$ binary image.

The Hamming distance approach considers all pixels as they have the same weight, i.e. if an $LL_3$ binary image pixel matches a template pixel, then the increment is one for any pixel. To take into account the variation in light conditions, different weights should be used (weighted Hamming distance). Therefore, if the $LL_3$ binary image has more black pixels (this case is common in dark images), then a higher weight should be given to white pixels to decrease the black dominance, and if the $LL_3$ binary image has more white pixels (this case is common in bright images), then a higher weight should be given to the black pixels to decrease the white dominance.

The weights for each black and white pixel can be calculated using the following formulae:

$$Ww = \frac{B}{B+W} \quad (5)$$

$$Bw = \frac{W}{B+W} \quad (6)$$

where Ww and Bw are the white and black pixel weights respectively, and W and B are the numbers of the white and black pixels in the window respectively (W+B is the window size, which is equal to 13 x 13 in this study).

Consequently, HD will be calculated as in step 4 in the previous algorithm, but instead of increasing it by 1, it will be increased by the pixel weight; if there is a black pixel match, increase HD by Bw (equation (6)), and if there is a white pixel match, increase HD by Ww (equation (5)). Figure 3.13 depicts the proposed face localization method.

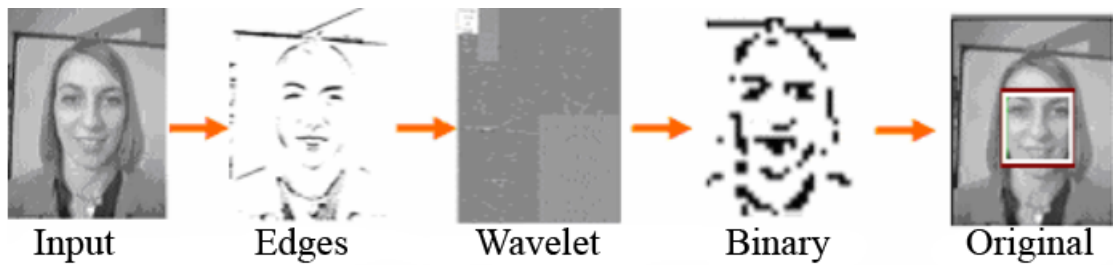

**Figure 3.13**. Applying the method on an image from the PDA database, from left to right, the input image, the proposed edge detection, 3$^{rd}$ level Haar wavelet, LL$_3$ binary image (resized for illustrative purposes), and the last image shows the result of the face localization.

### 3.2.5 The colour template score

Sometimes the highest Hamming score does not indicate the precise location of the face, or even shows it far away from the real location, due to complex backgrounds that may result in a better face pattern. A solution for this problem can be conveniently obtained by taking a few candidate potential face locations, rather than 1 location, which is associated with the highest HD (for example considering the locations associated with the highest *n* HD).

To choose the best solution from *n* candidate face locations, the colour information (face skin colour) in each location can be used, and then employing fuzzy logic to decide the best face location. This is a trade-off between speed and accuracy.

To calculate the face colour, a simplified version of the colour formula proposed by Kovac et al. (2003) for skin colour is used. The formula stipulates that in an RGB colour space a pixel belongs to a face if it satisfies the following condition:

R > 95 AND G > 40 AND B > 20 AND (R - min{G,B} > 15) AND (R - G > 15) AND R > B

This formula is applied to all pixels in the original coloured image, after resizing it to the size of $LL_3$. The colour score counter is then determined as the number of face pixels (that satisfy the previous condition and match the white (tissue) pixels in the template) in each window. Finally the wavelet counter and the colour counter are fused using equation (7) and a fuzzy decision table.

$$FZ(i) = \begin{cases} High & Counter(i) \geq \frac{Max(n) + Min(n)}{2} \\ Low & Otherwise \end{cases} \quad (7)$$

where for each counter (wavelet and colour), Max and Min represent the highest and lowest scores respectively over n candidate windows, and FZ is the input value for the fuzzy decision rule. This equation results in the first two columns in Table 3.1, and the output decision is the logical OR operation. Arbitrarily, the first candidate face window that outputs high is declared to be the location of the face.

**Table 3.1**. Fuzzy table for both the wavelet and the colour counters.

| Wavelet | Colour | Output |
|---------|--------|--------|
| High    | High   | High   |
| High    | Low    | High   |
| Low     | High   | High   |
| Low     | Low    | Low    |

It is found in this study that using the colour information increases the accuracy of the algorithm, but of course adds to the computation time. However, unlike most of the colour-based face detection schemes, colour information is used only on a small number (n) of candidate windows. The larger n is, the more accurate the algorithm will be. In this study n was chosen to be 5, to reduce computation time as much as possible.

## 3.3 Experimental results

The proposed scheme was applied using the following hardware, software and database:

Hardware: personal computer with Intel Pentium 4 CPU 2.4 GHz, and 512 MB RAM.

Software: Microsoft Visual C++ .NET version 7.1, and framework version 1.1.

Database: we used session 2 and session 3 of the PDA database[*], each session includes 4320 videos, with 521299 frames for both sessions of 60 different people, 30 males and 30 females, under different light conditions and backgrounds. The recorded subjects are of different ages, and some may have different accessories. The results were compared with Rowley's et al. (1998) face detection system, after applying both systems on the PDA database and under the same conditions.

The reason why the PDA database was chosen to evaluate face localization method is its diversity in lighting conditions and complex backgrounds. Moreover, each frame was taken as a standalone image without benefiting from the sequence and neighbouring frames. Therefore, a database of 521299 face images with different lighting conditions and complex backgrounds is appropriate and adequate for evaluating a face detection/localization scheme.

Table 3.2 presents the accuracy rates of the proposed method and the benchmark Rowley face detection scheme. The PDA database was designed to meet several circumstances: *light* versus *dark* videos, *inside* versus *outside* videos, and *males* versus *females* videos.

The proposed method and Rowley's method were evaluated using all the videos of the PDA database. The results shown in Table 2 are given for different classes of recordings and show the performance of both methods for each recording category using the first maximum 5 potential face locations.

The overall accuracy rate (93.40%) for the proposed method is significantly higher than that for Rowley's method (85.93%). The results also demonstrate that, unlike Rowley's method, the proposed method is not affected much by the characteristics of the various categories.

In fact the performance of the proposed method is almost invariant to light conditions, place of recording and gender, in contrast with the performance of Rowley's method, which fluctuates between 79.06% and 92.79%.

---

[*] Some samples of the PDA database can be found on http://johanhendrikehlers.com/index.html.

**Table 3.2**. Face localization/detection results, after applying both methods on the PDA database.

| Video | Method | Illumination | | Indoor Outdoor | | Gender | | All |
|---|---|---|---|---|---|---|---|---|
| | | Light | Dark | Inside | Outside | Male | Female | |
| Session2 | Proposed | **92.60%** | **93.55%** | **95.21%** | **91.16%** | **93.43%** | **92.65%** | **93.10%** |
| | Rowley | 92.09% | 82.48% | 85.06% | 89.52% | 83.55% | 91.02% | 87.29% |
| Session3 | Proposed | **93.77%** | **93.51%** | **96.06%** | **91.66%** | **91.97%** | **95.23%** | **93.70%** |
| | Rowley | 93.50% | 75.63% | 82.84% | 86.29% | 82.15% | 86.98% | 84.56% |
| Both | Proposed | **93.19%** | **93.53%** | **95.64%** | **91.41%** | **92.70%** | **93.94%** | **93.40%** |
| | Rowley | 92.79% | 79.06% | 83.95% | 87.90% | 82.85% | 89.00% | 85.93% |

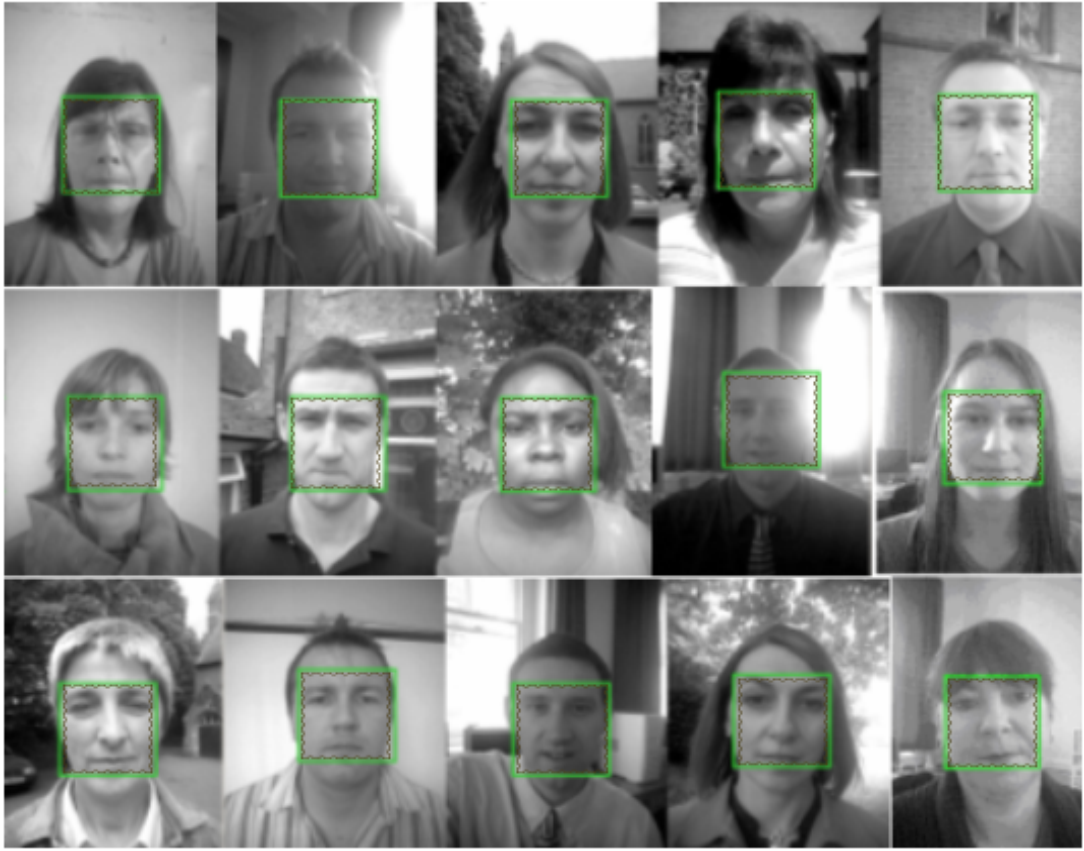

**Figure 3.14**. The output of the proposed face localization method after applying it on some videos from the PDA database.

It was also found that the proposed method works well on grey scale images and videos (i.e. without using the colour information), but the overall accuracy is reduced to 89.50%. Therefore, using the colour information has an improving impact on the proposed method, increasing the accuracy to 93.40%.

This proves that any kind of information can be helpful for such a research problem, and also shows that colour (which was neglected by some researchers because it is unstable in regard to different light conditions and different camera sensors) is a rich source of information. This can be helpful for object detection in general, and face detection in particular. Figure 3.14 illustrates the success of the proposed face localization method.

## 3.4 Time and space complexity

As stated earlier, the proposed method consists of several steps (processes), each step requiring its share of time and space from the computer resources to accomplish its task. Going back to Figure 3.5, the major processes that consume time and space are: edge detection, the wavelet transform, the template matching on $LL_3$ and the template matching on the resized coloured image.

For the edge detection, there are 2 filters, each of size 5 x 5, and only one of them is chosen each time (depending on the local information), so one filter at a time performs on the whole image, assuming the size of the image is n x n, so the edge detection complexity can be calculated as follows:

$$25n^2 (1^{st}\ filter) \quad or \quad 25n^2 (2^{nd}\ filter)$$
$$O(25n^2) = O(kn^2) \approx O(n^2) \qquad (8)$$

For the wavelet transform, the coefficients were calculated (pyramidally) for three levels, the first level of size n x n (the image size), the second n/2 x n/2 and so on, therefore, the wavelet transform complexity can be calculated as follows:

$$n^2 (1^{st}\ level) + \frac{n^2}{4}(2^{nd}\ level) + \frac{n^2}{16}(3^{rd}\ level)$$
$$O(n^2 + \frac{n^2}{4} + \frac{n^2}{16}) \approx O(n^2) \qquad (9)$$

For template matching on the $LL_3$ binary image, the template size is 13 x 13, which is matched with all possible locations in the $LL_3$ binary image of size $n^2/16$, so the template matching complexity in this case can be calculated as follows:

$$13x13 \, (template \, size) \, x \, \frac{n^2}{16} \, (LL_3 \, size)$$

$$O(169x\frac{n^2}{16}) \approx O(k\frac{n^2}{k}) = O(n^2) \qquad (10)$$

For the template matching on the resized coloured image, the coloured image is resized to the size of the $LL_3$, which is $n^2/16$, and the template size is 13 x 13, which is matched with only some candidate locations (k locations with the highest HD scores) in the resized coloured image, so the template matching complexity in this case can be calculated as follows:

$$13x13 \, (template \, size) \, x \, k \, (locations \, with \, the \, highest \, HD \, scores)$$
$$O(169k) \approx O(k) \qquad (11)$$

The overall complexity of the proposed face localization method involves the complexity of all its processes, which can be calculated using equations 8, 9, 10 and 11 as follows:

$$O(n^2) + O(n^2) + O(n^2) + O(k) \approx O(n^2) \qquad (12)$$

Both time and space complexity are equal for the proposed method which is $O(n^2)$. However, applying the proposed method to the videos decreases the computation time by tracking the face, i.e. the method can localize the face in a sub-area instead of searching the whole space ($n^2$), but this will not decrease the complexity theoretically, neither for the time nor the space.

## 3.5 Summary

In this chapter the development of an efficient scheme for face localization in video images captured on mobile constrained devices was discussed. In a wide range of lighting conditions and different backgrounds, the proposed method was evaluated for use in the proposed visual word/speech recognition system in a mobile environment.

Due to the specific nature of facial features, the multi-stage scheme is initiated with a special edge detection procedure that highlights the main facial features, and fuses wavelet information extracted from the transformed edge-highlighted image with colour information extracted from a small number of candidate windows in the original RGB images. A large number of experiments were conducted on the PDA database by testing the proposed method as well as the widely used high performing general purposes Rowley's et al. (1998) face detection scheme.

It has been demonstrated in this chapter that the proposed scheme has a high accuracy rate, and significantly outperforms the Rowley method in almost every category of recordings. Unlike Rowley's scheme, the performance of the proposed scheme is stable across different categories, indicating strong robustness against variations in lighting conditions, which is a major problem in many existing schemes.

Using wavelet decreases the time needed for various scanning steps compared to scanning the spatial domain, and indirectly reduces the amount of work done in relation to colour information by limiting the work to a few candidate windows.

Although the proposed method was designed to work only under certain assumptions, it is not difficult to modify it to work under any other different set of assumptions. Moreover, the method is very efficient with modest time-complexity ($O(n^2)$), and easy to implement even on constrained devices such as mobile phones and PDAs.

As mentioned previously, this method is not proposed as a substitute for existing face detection methods, but aims to facilitate mouth and lips detection for visual word/speech recognition for mobile devices where variation of lighting conditions are to be expected and catered for.

# Chapter 4

# Lip Localization

The lips and mouth region are the visual parts of the human speech production system; these parts hold the most visual speech information, therefore, it is imperative for any VSR system to detect/localize such regions to capture the related visual information, i.e. we cannot read lips without seeing them first.

Lip localization is an essential process in the proposed visual words (VW) system. Actually, the first step of this system – the face detection – was proposed only to facilitate lip localization, hence the face is more easily detected than the lips. Building up our knowledge of the human face, lips are located in the lower centre part of the face, which makes localization much easier for the algorithm.

The major aim of this chapter is to investigate lip detection/localization problems and propose a robust method to address the needs of the proposed VW system. The rest of this chapter is devoted to reviewing existing trends in lip localization, presenting the proposed lip localization methods and discussing, which is the best method to be used in the VW system. In addition, the evaluation experiments and results are discussed, after applying the proposed method to an in-house video database, which is also described in this chapter.

## 4.1 Existing trends in lip localization

Over the last few decades, the number of applications that are concerned with the automatic processing/analysis of human faces has grown remarkably, such as face detection, face recognition, visual speech recognition, and facial expression recognition. A recent application was the use of a face recognition system at Heathrow airport for security reasons (BBC News, 2008). Many of these applications have a particular interest in the lips and mouth area. For such applications a robust and real-time lips detection/localization method is a major factor contributing to their reliability and success. Since lips are the most deformable part of the face, detecting them is a nontrivial problem, adding to the long list of factors that adversely affect the performance of image processing/analysis schemes, such as variations in lighting conditions, pose, head rotation, facial expressions and scaling.

Lip detection methods that have been proposed in the literature so far have their own strengths and weaknesses. The most accurate methods often take more processing time, but some methods proposed for use in online applications trade accuracy for speed.

Many techniques for lips detection/localization in digital images have been reported in the literature, and can be categorized into two main types of solutions:

1. *Model-based* lips detection methods. Such models include the use of spline-based deformations called "Snakes", Active Shape Models (ASM), Active Appearance Models (AAM), and deformable templates.

2. *Image-based* lips detection methods. These include the use of spatial information, pixel colour and intensity, lines, corners, edges, and motion.

We shall now review some of the most common lip detection schemes in these different categories.

### 4.1.1 Model-based lip detection methods

This approach depends on building lip model(s), with or without using training face images and subsequently using the defined model to search for the lips in any freshly input image. The best fit to the model, with respect to some prescribed criteria, is declared to be the location of the detected lips. Each model has its own criteria for measuring a fitness score. These methods include the active contours (Snakes), Active Shape Models, Active Appearance Models and deformable templates.

#### 4.1.1.1 Snakes

Snakes or Active Contour Models were first introduced by Kass, Witkin, and Terzopoulos (see Kass et al., 1987). Snakes have been proposed as a general method for shape detection. Their optimisation techniques are based on splines, and the idea is to iteratively minimize the energies of the spline to fit local minima. The optimal fit between a snake and the detected shape is found by minimizing the energies in the following equation:

$$E_{snake} = \int_0^1 (E_{int}s(i) + E_{img}s(i) + E_{con}s(i))di \qquad \cdots(4.1)$$

where $s(i)$ is the snake parameter at point $(x_i, y_i)$; $E_{int}$ is the internal energy caused by the snake stretching and bending; $E_{img}$ is the image features energy which attract the snake such as lines and edges; and $E_{con}$ is the external energy (constraints) forced by the user or coming from higher level shape information. Figure 4.1 illustrates the lip localization process using the snakes approach.

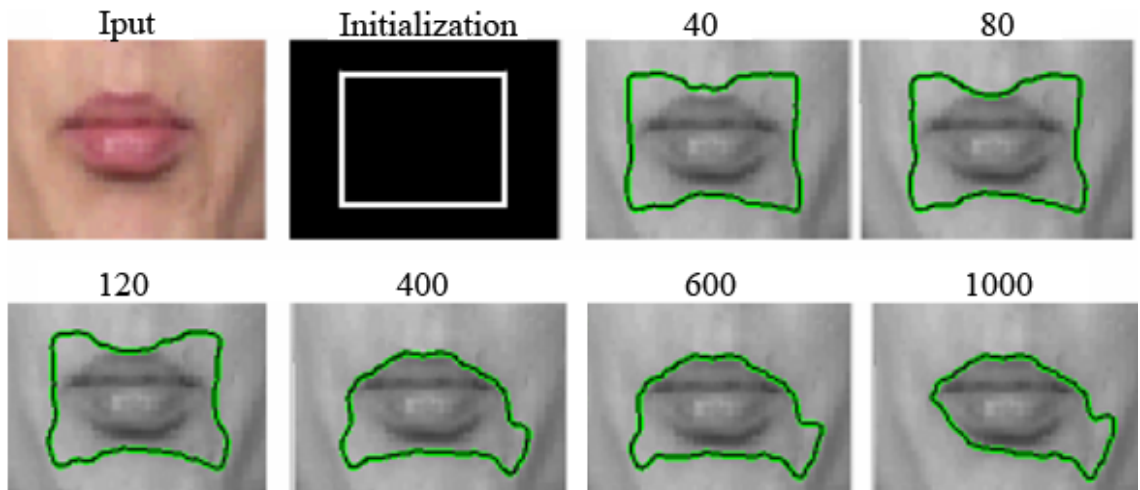

**Figure 4.1.** Lip localization (segmentation) using Snakes approach on input image. The number above each image shows the number of iterations needed for snake parameters tuning.

The active contours (snakes) method has been applied efficiently for lips and mouth region detection by various researchers including Delmas et al. (1999). This method was evaluated using 120 images with 90% mouth good localization, and with good snake initialisation. Barnard et al. (2002) and Eveno et al. (2002a) reported good lip localization representative results for sequences of 6 and 14 different subjects respectively.

However, a number of problems can be associated with the use of the snake technique for lips and mouth region detection (see Eveno et al., 2002a). These problems can be summarised as follows:

1. They can fit to the wrong feature, such as the nose or the chin; especially if the initial position was far from the lip edges (see Figure 4.1).
2. Snakes like splines do not easily bend sharply so it is difficult for them to locate sharp curves like the corners of the mouth.
3. They can be fooled easily by facial hair (moustache and beard).

4. Sometimes (depending on the detected object) the tuning of snake parameters is very difficult to achieve, and takes a long time (several seconds!) (see Figure 4.1).

**4.1.1.2 Active Shape Models**

Active Shape Models (ASM) can be used successfully for facial features detection, particularly the lips and mouth region. ASM was originally proposed by Cootes and Taylor in 1992 (Cootes et al., 1992). ASM are statistical models of the shapes of objects, which iteratively adjust to fit to the detected object in digital images. A shape is represented by a set of n labelled "landmarks", where each point is represented by its x and y axis.

$$X=\{(x_1,y_1), (x_2,y_2), ...., (x_n,y_n)\} \quad \ldots(4.2)$$

Using a training set of a land-marked object in images, ASMs build a statistical shape model of the global shape variation from the training set. In this study the object is the lip. This model is used to fit a model to new occurrences of the trained object. A principal component analysis (PCA) is used to construct such a model (see Figure 4.2).

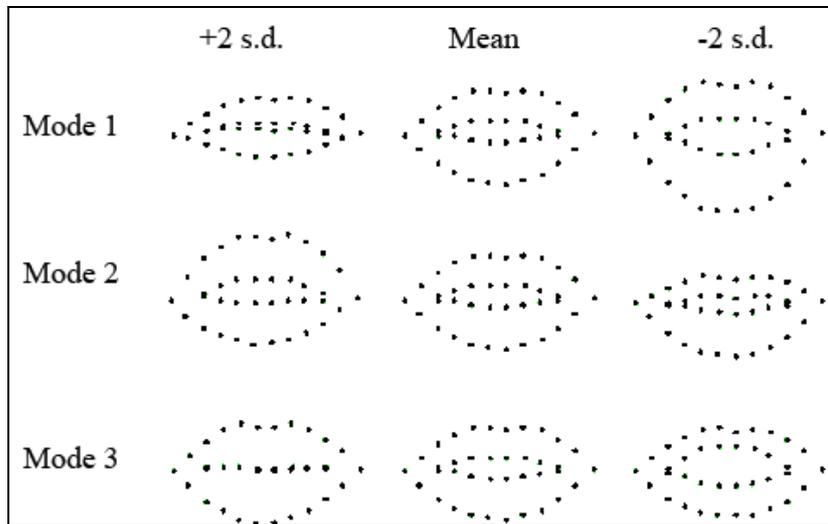

**Figure 4.2.** Mean shape and the first three modes of variation (by ±2standard deviation (s.d.)) of ASM proposed by Luettin, et al, (1996d).

Assuming that $i$ is the shape instance and $1 \leq i \leq m$, the mean shape $\bar{x}$ and covariance matrix $C$ are computed as follows:

$$C = \frac{1}{m-1}\sum_{i=1}^{m}(x_i - \bar{x})(x_i - \bar{x})^T \quad \cdots(4.3)$$

$$\bar{x} = \frac{1}{m}\sum_{i=1}^{m} x_i \qquad \cdots (4.4)$$

where *m* is the number of shapes, and *C* is used to compute the first *k* eigenvectors $\Phi_i$, $i = \{1, 2\ldots k\}$ and their corresponding Eigenvalues $\lambda_i$.

Any shape instance x can be approximated using the matrix **ϕ**:

$$\boldsymbol{\phi} = [\Phi_1,\ \Phi_2,\ \Phi_3 \ldots \Phi_k] \qquad \ldots(4.5)$$

as:

$$x \approx \bar{x} + \boldsymbol{\phi}\, b \qquad \ldots(4.6)$$

where b is a k-dimensional vector that represents the parameter space of the deformable model.

The ASM building algorithm is defined as follows:

1. Define shape representation through n landmarks.

2. Annotate M images (manual or automatic annotation for the shapes in the training data).

3. Align the shapes, using Procrustes shape analysis (translation $\Delta x$ and $\Delta y$, remove the scaling *s*, and rotation $\theta$)

4. Make PCA of deformations.

5. Make edge model.

6. Apply ASM search algorithm.

The ASM search algorithm is defined as follows:

1. Input image *y*

2. *b*=0

3. $x = \bar{x} + \boldsymbol{\phi}\, b$

4. Optimise the rigid body transform parameters $\Theta$ rigid = $(\Delta x,\ \Delta y,\ \theta,\ s)$ for the best fit between x and y.

5. Find *b* that makes x best fit y

6. If no convergence, go to 3, else exit. (Convergence can be achieved by a specific number of iterations, or using a measurement distance over a specific threshold.)

Obviously, the greater the number of iterations, the better the end result of the algorithm. A very recent study by Milborrow and Nicolls (2008) proposes an extended version of ASM for facial features detection; some of the extensions include fitting more landmarks, and the use of two dimensional landmarks templates instead of one dimension[1]. Their representative results illustrate better performance by adding more landmarks.

Caplier (2001) proposes an algorithm, which can extract lip shape over an image sequence. The algorithm works under natural lighting conditions, and uses an Active Shape Model to describe the mouth. The mouth model is iteratively deformed under constraints in the light of image energies. Caplier uses Kalman filters on the independent mouth parameters, to give an initial shape close to the final one, which speeds up the convergence time.

Mahoor and Abdel-Mottaleb (2006) worked on an enhanced version of ASM by employing colour information, which is used to localize the centres of the mouth and the eyes to assist the initialisation step. Their good representative results emphasized the power of colour as a rich source of information.

Some other researchers who have worked on ASM include Al-Zubi, (2004), Luettin, et al, (1996a, 1996b, 1996c), Luettin and Dupont, (1998), and Luettin, et al. (2001). Luettin, et al.'s (1996d) method was evaluated on a 96 image sequence of 12 different subjects, and 91.7% of the lips were detected correctly.

**4.1.1.3 Active Appearance Models**

Active Appearance Models (AAM) are the same as ASM, but instead of using the edge profile along a landmark, AAM are extended to include the grey scale information of the whole image along with the shape information. (Cootes, et al., 1998).

According to Cootes, et al., there are three major differences between ASM and AAM:

---

[1] State-of-art ASM free library, "MSAM" was developed by Milborrow and Nicolls (2008), and can be downloaded from the following link: http://www.milbo.users.sonic.net/stasm/download.html

1. The ASM uses models of the image texture around the landmark points, whereas the AAM uses a model of the image texture of the whole region.

2. The ASM searches around the boundary, whereas the AAM only samples the image under the current position.

3. The ASM seeks to minimize the distance between the model and the image points, whereas the AAM seeks to minimize the difference between the synthesized model image and the target image.

Figure 4.3 shows the first three modes at ±2 standard deviations about the mean of the combined appearance model trained on a specific database.

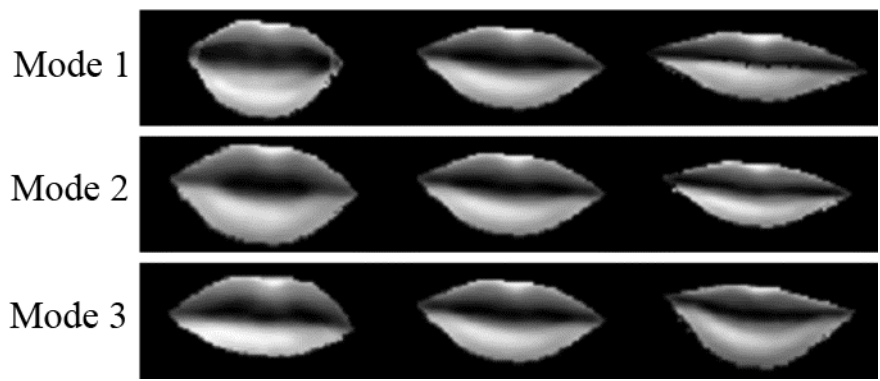

**Figure 4.3.** Combined shape and grey-level appearance model, first three modes of variation at ±2 standard deviations about the mean (Matthews et al., 2002).

Matthews et al. (2002) illustrated good lip detection results using AAM after 15 iterations. See Figure 4.4.

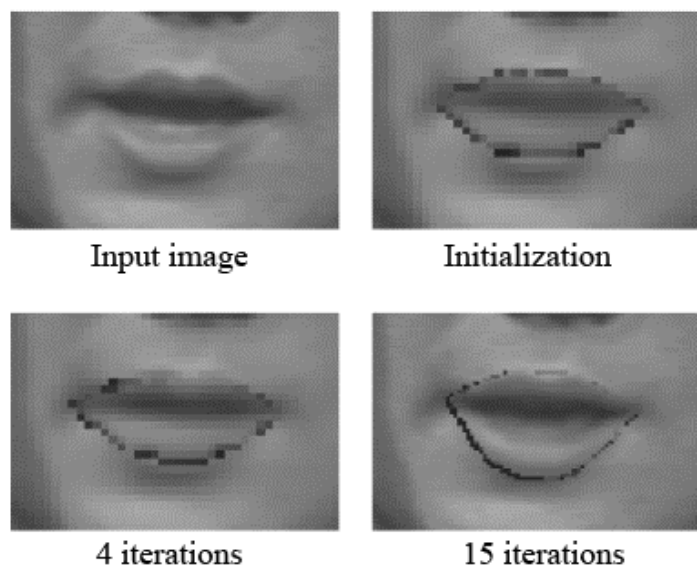

**Figure 4.4.** Example of AAM search, the model is initialised in the centre of the image, convergence took 15 iterations (Matthews et al., 2002).

Unlike the Snakes method, AAM converges using a small number of iterations because these models have some prior information about the shape, but they are not necessarily fast, because there are more parameters to be handled.

Another more recent work involving lip detection using AAM was done by Aubrey et al. (2007), who discovered that the AAM approach was more consistent with the detection of the non-speech sections containing complex lip movements.

**4.1.1.4 Deformable templates**

As introduced by Yuille et al. (1989), a deformable template is a parameterised mathematical model used to track the movements of a given object. It is *active* in the sense that it can adapt itself to fit the given object. The deformable lip template adjusts its shape according to the value of a number of integrals along the relevant contours (see Figure 4.5).

In a manner similar to that of snakes, an energy function is defined to link edges, peaks, and valleys in the input image to the corresponding parameters in the template. The best fit of the elastic model is found by minimizing an energy function of the parameters. (Yang, et al., 2002).

It is a useful shape model because of its flexibility, and its ability to both impose geometrical constraints on the shape and to integrate local image evidences (Fisher, 1999). The work done by Hennecke et al. (1994) employs a deformable template for lip tracking, making use of several configurational and temporal penalty terms, which keep erroneous template deviations under control. This method was evaluated using 50 image sequences; the reported observed results were satisfactory and stable under unfavourable lighting conditions.

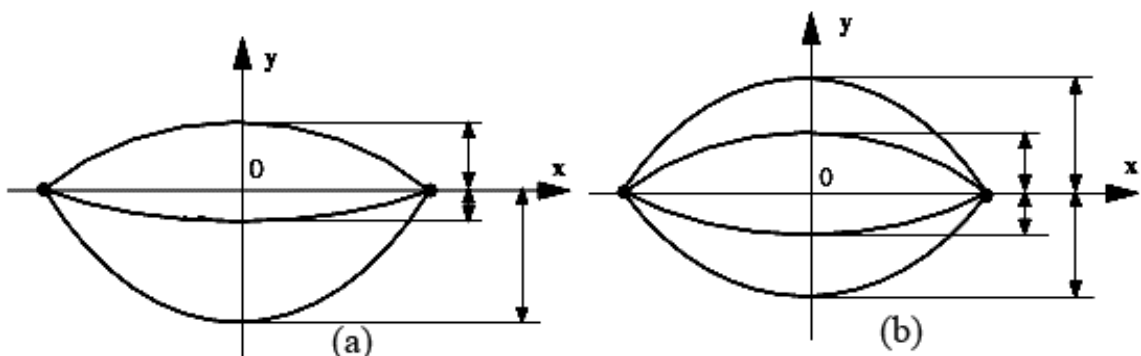

**Figure 4.5.** Mouth deformable templates as proposed by Zhang (1997); (a) mouth-closed deformable template, (b) mouth-open deformable template.

Other works that have dealt with lip detection and/or tracking based on a deformable template include that of Zhang (1997). In this contribution, the mouth features are represented by the mouth corners and the lip contour parameters, which describe the opening of the mouth. This model facilitates the development of an algorithm to determine whether the mouth is open or closed.

There are several drawbacks when using deformable templates, such as: high computational time complexity, the unexpected shrinking of the templates and rotation of the template. Wu et al. (2004) identified the aforementioned shortcomings and proposed some solutions to eliminate these drawbacks. For more information refer to the same article.

### 4.1.2 Image-based lips segmentation methods

Since there is a difference between the colour of lips and the colour of the face region around the lips, detecting lips using colour information attracted researchers' interest recently – simplicity, not time consuming, and the use of fewer resources e.g. low memory; allowing many promising methods for lip detection using colour information to emerge.

The most important information that researchers focus on are: using the red and the green colours in the RGB colour system, the hue of the HSV colour system, and the component of the red and blue in the YCbCr colour system. Some of these researchers also used more information from the lip edges and lip motion. A well known mixed approach is called the "hybrid edge" (Eveno, et al., 2002a).

**4.1.2.1 RGB approach**

Colours are seen as variable combinations of the so-called primary colours red (R), green (G), blue (B). The primary colours can be added to produce the secondary colours of light magenta, cyan, and yellow. The RGB colour system can be represented by a cube, where R, G and B values are at three corners. Cyan, magenta and yellow are at the other three corners, where black is at the origin, and white is at the corner farthest from the origin (Gonzales and Woods, 2002). In RGB space, skin and lip pixels have different components. The red is dominant for both, the green is more than the blue in the skin colour, and skin appears more yellow than lips (Eveno et al., 2002a).

Gomez et al. (2002) propose a method for lip detection in which the image is transformed by a linear combination of red, green and blue components of the RGB colour space. Then they apply a high pass filter to highlight the details of the lip in the transformed image, after which both of the generated images are converted to obtain a binary image. The largest area in the binary image is recognized as the lip area.

The Zhang et al. (2003) lip detection method is based on red colour exclusion and Fisher transform. Their approach works as follows: firstly, the face region is located using a skin-colour model, and motion correlation, then the lip region of interest (ROI) is defined as the area in the lowest part of the face. Secondly, the R-component in RGB colour space is excluded, then G-component and B-component are used as the Fisher transform vector to enhance the lip image. Finally, an adaptive threshold to separate the lip colour and the skin colour is set according to the normal distribution of the grey value histogram. Their system was evaluated using 120 images taken from video sequences of 4 persons, under normal light conditions (day light). 20 images for each person were selected for training the Fisher classification vector and 10 images for testing, and they reported 100% detection rate. Such a perfect result is expected using such a small number of images and subjects under normal light conditions.

**4.1.2.2 HSV approach**

The RGB colour system is most commonly used in practice, while the HSV is closer to how humans describe and interpret colour. Hue (H) represents the dominant colour as perceived by someone, so if someone calls an object green, blue, orange, or any other known colour, he/she is specifying its hue. Saturation (S) refers to the amount of white light mixed with a hue. The brightness (or intensity) is represented by the value (V). There are specific equations to convert from RGB to HSV and vice versa. For more information about these equations see Gonzales and Woods (2002).

Apparently, hue is a key feature for detecting lips using the HSV colour system, since the hue value of the lip pixels is smaller than that of the face pixels. Coianiz et al. (1996) use the filtering hue method to detect the lips in coloured images. The hue is filtered by the following weight function:

$$f(h) = \begin{cases} 1 - \dfrac{(h-h_0)^2}{w^2} & |h-h_0| \leq w \\ 0 & \text{otherwise} \end{cases} \quad \cdots (4.7)$$

where $h$ represents the hue value, $h_0$ is the centre of the filter, and w is a parameter controlling the distance from $h_0$.

Coianiz et al.'s (1996) mouth localization method works as follows:

- The hue-filtered image is sub-sampled to reduce the computational time, and to reduce the effect of the insignificant details.

- The resulting image is then thresholded; the value of the threshold can be determined empirically.

- A special distribution of pixels above the threshold is analysed.

Researchers normally detect the face then use it to detect the lips, while Jun and Hua, (2008) proposed a reverse technique, by which they detect the face using the lips. Their method determines skin area based on a threshold value in an HSV colour model, then it searches the rough lip area based on the skin chrominance adaptively. After that it employs an R/G filter to find the lip, depending on the geometrical relation between face and lip; the face is finally located.

**4.1.2.3 Hybrid edge**

The "hybrid edge" is the combination of the colour and luminance information. It was first produced by Eveno et al. (2002a). In this work Eveno et al. use a pseudo hue definition which demonstrates the difference between the lips and the skin colour. The pseudo hue can be computed as follows:

$$h(x, y) = \dfrac{R(x, y)}{R(x, y) + G(x, y)} \quad \cdots (4.8)$$

where R(x,y) and G(x,y) are the red and the green components of the pixel (x,y) respectively.

The use of the pseudo hue with gradient vector of the intensity of the upper, middle and lower sections of the lip, allows a reliable estimation of several key points positions.

Then 5 cubic polynomial curves are fitted using these points to represent the outer lips boundary (Eveno, et al., 2002a). Their method was evaluated using sequences of 14 different speakers, acquired under natural non-uniform lighting conditions. This study shows good representative lip localization results of 7 different subjects, 3 images each.

A more recent study by Cetingul et al. (2005) uses the hybrid edge as a part of a quasi-automatic system to extract and analyse lip-motion features for speaker identification.

**4.1.2.4 YCbCr approach**

This colour space is used in digital videos, where Y is the luminous component, and Cb and Cr are the blue-difference and red-difference chroma components respectively (Acharya and Ray 2005). The lip region has high Cr and low Cb values (Nasiri, et al., 2008).

Nasiri et al. (2008) propose a novel lip detection method using Particle Swarm Optimisation, which is used to obtain an optimised map. After mapping the image to the YCbCr colour space, two sets for lip and face pixels are created to be clustered by a clustering algorithm. This method was evaluated using 2 databases, the first one consisting of images of 114 persons with 7 different facial expressions, where the lip localization accuracy was 91.22% and the second database consisting of facial images of 50 persons, where 92% of the lips were detected accurately.

Jaroslav (2004), one of many researchers who use the YCbCr colour space for detecting lips, based his method on the fact that lips are more red than faces, so he focused on the Cr component using a specific equation to maximize the Cr and minimize the Cb. After using edge detection as a mask to remove any unnecessary information, the image output of the equation is thresholded, and the largest area is found to be the mouth area.

The previous approaches were mainly focused on classifying the image pixels according to their colours to lips or non-lips pixels. Some researchers add some other information such as edge and motion information, or even without using any colour information; instead, they use information such as the edges and contours of the lips.

Such an approach is the work done by Guitarte et al. (2003). Guitarte and his colleagues propose a real time method consisting of lip finding in the first frame, and to save processing time, they propose another method for tracking the lip in the other frames.

Lips are found among a small number of blobs, which share a common characteristic; in their case, the blobs belong to the same horizontal contour, which should fulfil geometric constraints.

Their simplicity, non-time consuming, and the use of fewer resources motivate the use of the image-based lips segmentation methods. However, several serious problems arise when we use this approach, such as:

- The colour representation of a lip acquired by a camera is influenced by many factors such as the ambient light, movement, etc.

- Different cameras produce different lip colour values, even for the same person's lips under the same lighting conditions.

- Human lip colour is different from race to race, and from person to person within the same racial group.

The previous problems were highlighted by Yang et al. (1998) to illustrate the problems of the use of colour approach for face detection, but here the author projects some of Yang et al.'s results on the lips instead of the face, because (in this case) what applies to the face definitely applies to lips.

Accordingly, using contemporary technology, it is not feasible to obtain good results for lip segmentation using a simple predefined threshold as a decision rule to classify pixels (to lip and non-lip) using their colour values. Thus if we want to continue using the colour approach for lip detection, we need to solve the previous problems.

## 4.2 The proposed lips detection approaches

The previous model-based lip-detection methods need a significant amount of processing time, and training time for those which need training steps, which makes them difficult to be applied for the online systems, or to be applied on low resources machines, like the PDAs, in which this study is more interested.

Having considered the various methods for lip detection, we found that the active shape models was the best performing technique that could be applied to lip detection, although it is affected by factors such as facial hair. However, we found that it does not meet some of the many functional requirements (e.g. in terms of speed), of our project.

"The implementation of Active Shape Model ASM was always difficult to run in Real-Time" (Guitarte, et al., 2003:3). The same thing applies to the AAM approach. In fact, AAM is slower than ASM.

In grey level images and under diffuse lighting conditions, the external border of the lip is not sharp enough (see Figure 4.6a), and this makes the use of techniques based on the information provided by these images alone ineffective (Coianiz, et al., 1996). Moreover, the ASM and the AAM algorithms are sensitive to the initialisation process. When initialisation is far from the target object, they can converge to local minima (see Figure 4.6a). Other problems, like the appearance of the moustache, beard, and unexpected accessories like mouth rings, contribute to the problems and challenges that model-based lip detection methods undergo. Figure 4.6b illustrates this problem for the ASM scheme.

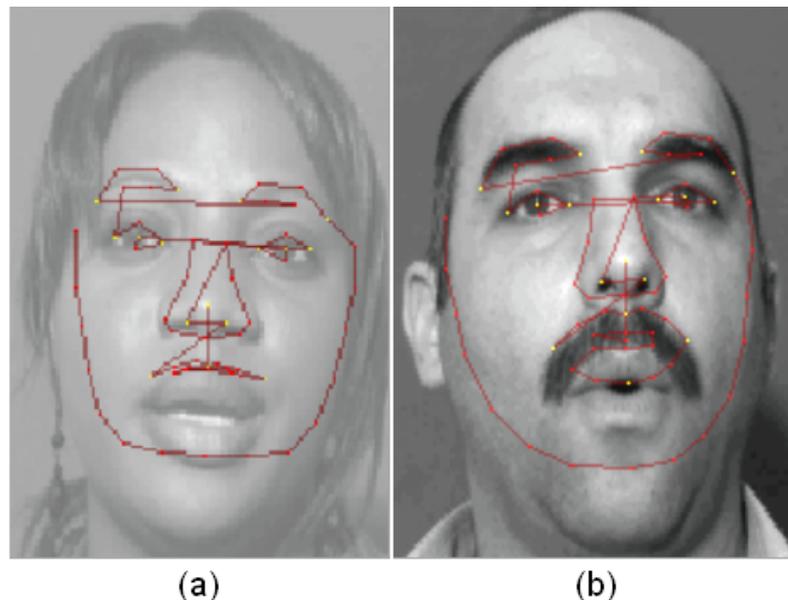

**Figure 4.6.** Facial features detection using MASM*, (a) lip detection converges to local minima, (b) the effect of facial hair on ASM convergence.

Since the visual speech recognition (VSR) problem needs several pre-processing steps, e.g. face and lips detection, it is vital for the VSR system to have faster solutions for these steps in order that the final solution can work in real time.

In order to overcome the above-mentioned difficulties, we shall incorporate colour information into our proposed scheme. In spite of being vulnerable to variations in light conditions, we shall see later that the use of colour for detection enhances performance for only modest increases in run-time. In other words, this is a typical time-accuracy dilemma: whether to use accurate methods that take up more time, or to use less accurate methods with less time.

---

* "MSAM" is a state-of-art ASM free library, developed by Milborrow and Nicolls (2008), and can be downloaded from the following link: http://www.milbo.users.sonic.net/stasm/download.html

Two lips localization methods using colour information are proposed in this study. The first method, which is called the "layer fusion" method, uses some of the previous image-based approaches, each approach (layer) voting for each pixel to be a lip or non-lip pixel. The second proposed method, which is called "the nearest colour" method, is based on using the YCbCr approach to find at least any part of the lip, then using the pixel values of this area to train the system, thus, the rest of the pixels will be clustered depending on the training data.

Both methods were programmed and evaluated along with some other image and model -based techniques for lip detection, using the same database, under the same circumstances.

### 4.2.1 Layers fusion

Both of the proposed algorithms rely on face detection pre-processing, because the face has many features, and is not subject to change like the lips, so it is easier and more accurate to detect it first, statistically. Human lips are located in the $3^{rd}$ tri-sector of the face, so the region of interest (ROI) for lip localization is the $3^{rd}$ tri-sector of the face.

Hue, RGB, YCbCr, edges and motion information: all have been used in this method; each one works alone by clustering the pixels using their specific information into two categories: lips and non-lips pixels. The clustering method that was used in this method is the k-mean clustering algorithm.

The output of the clustering method is a binary image of the ROI; the same thing is applied to the same area, but using different information: RGB, Hue, Cb, Cr, edges and motion. Similarly, several binary images are then obtained, each image is considered as a layer, and each layer votes by 1 for each lip pixel, or zero if not.

Fusing all the layers by taking the sum of the votes for each pixel produces a voting matrix. Starting with the highest values in the voting matrix, the algorithm starts to expand the lip area in all directions (left, right, up and down), and the algorithm stops expanding the lip area if the value of the voting matrix at the current location is less than a pre-defined threshold (e.g. threshold =2".) i.e. for each pixel to be considered as a lip pixel, it should be voted by at least 2 methods, and connected to the expanded area which contains the highest votes. After considering all the locations of the voting matrix, the largest area is recognized as the lip area.

The luminance and the colour information in the RGB are mixed together, to solve this problem. The tri-chromatic coefficients have been broadly used, and can be expressed by the following equations:

$$r = \frac{R}{R+G+B} \quad \cdots(4.9)$$

$$g = \frac{G}{R+G+B} \quad \cdots(4.10)$$

$$b = \frac{B}{R+G+B} \quad \cdots(4.11)$$

where $r + g + b = 0$ and $R+G+B \neq 0$.

Each value is taken as part of a feature vector, which represents each pixel, to be clustered using the k-mean algorithm.

The hue sector of the lips is located in the range [0, $\theta$] and [360 -$\theta$, 360] values in the HSV, so any clustering method will be confused by the low and high hue values, which belong to the same cluster (the lips). To solve this problem the lip hue should be warped; the following equation was used to warp the hue of the lip:

$$H = \begin{cases} h & 0 \leq h \leq 180 \\ 360 - h & \text{otherwise} \end{cases} \quad \cdots(4.12)$$

where H is the new warped hue and h is the original hue.

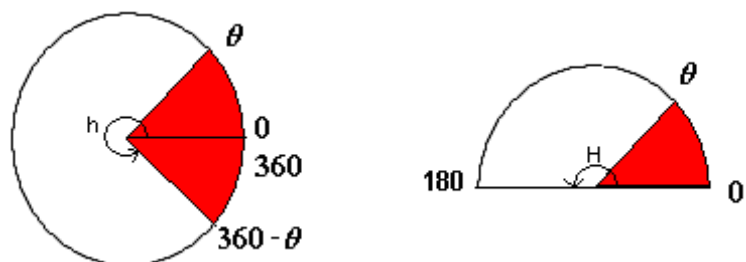

**Figure 4.7.** (left) Lip hue distribution in the HSV colour space, (right) the warped hue.

In YCbCr, chromatic and luminance components are separated, and lip pixels have high Cr and low Cb (Nasiri, et al., 2008). To foster this property the following equation is used:

$$Lip - Map = Cr^2 \left( Cr^2 - \eta \frac{Cr}{Cb} \right)^2 \qquad \cdots (4.13)$$

$$\eta = 0.95 \frac{\sum Cr^2}{\sum \frac{Cr}{Cb}} \qquad \cdots (4.14)$$

where $Cr^2$ and Cr/Cb all are normalized to the range [0,1], and Cb can be computed from the RGB using the following equation:

$$Cb = 128 - 0.168736 * R - 0.331264 * G + 0.5 * B \qquad \ldots (4.15)$$

and Cr:

$$Cr = 128 + 0.5 * R - 0.418688 * G - 0.081312 * B \qquad \ldots (4.16)$$

The "Lip-Map" formula (Equation 4.13) is designed to increase the values of the pixels where Cr is high and Cb is low (the case of the lips pixels); now the lip-map values of the ROI can be sent to the clustering algorithm, the higher values are lip-pixels, and the lower ones are non-lip pixels.

The pseudo hue, which was mentioned earlier, is also used in this method, clustered, and its votes given to the voting matrix.

The edges information is also used to vote for this method. Edges are detected to highlight the boundary of the lips, and the features of the mouth, and to destroy the texture of the face. The edge filter was the Sobel edge horizontal and vertical filters (Sobel and Feldman, 1968). Edge values are approaching 255 or more, while the texture values are around 0, or less. This will be done locally (the size of the filter 3x3), so clustering the output edge detection image depending on the intensity, will not be affected by the light conditions, and will cluster the pixels into two different groups: the edge groups (more likely lip pixel), and the texture group (more likely face pixel).

The motion information is used here to track the lips rather than localizing them. The ROI in the next frame is subtracted from the current ROI. If the difference is more than a specific threshold, this means that the mouth changes its location or status (open-close or close-open). In this case, the whole lip localization method should be triggered; in some cases the face detection method is also triggered.

After making use of all possible information, which increases the number of the votes for each pixel, i.e. increase or decrease the probability of each pixel in ROI to be a lip-pixel or non-lip pixel, the voting matrix looks like that in Figure 4.8.

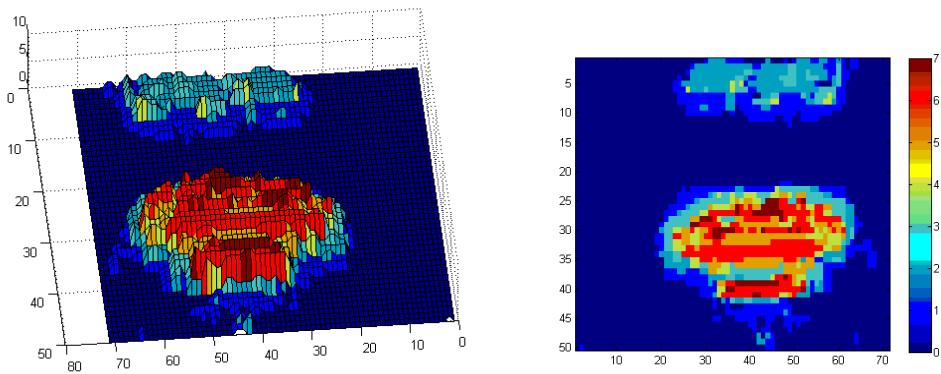

**Figure 4.8**. Representations of the voting matrix. 3d (left) and 2d (right).

The "Layer Fusion" algorithm works as follows:

1. For each pixel in ROI, get the r, g and b values from the RGB colour system.

2. For each pixel in ROI, cluster the vector {r, g, b} to lips or non-lips pixels and save the result in a binary image.

3. Repeat 1 and 2 but the cluster vector contains r/(r + g) instead of r, g and b.

4. Repeat 1 and 2 for the Hue instead of the r, g and b.

5. Repeat 1 and 2 for the Lip-Map instead of the r, g and b.

6. Apply Sobel edge detector on the ROI.

7. For each pixel in the edge image, cluster the intensity to lips/non-lips pixels and save the result in a binary image.

8. Create "voting matrix" with the size of ROI, and initialize its values to 0s.

9. For each binary image created in the previous steps, and for each pixel within that image, if the value is 1 (lips-pixel) increase the corresponding voting matrix value.

10. Go to the maximum values in the voting matrix, create output binary image, update the corresponding pixels to 1.

11. Spread in all directions over the voting matrix, and update the output binary image; stop spreading if the vote is less than 2.

12. Filter the output binary image using opening filter.

13. Label the largest area as the lips area.

The major drawback of this method is the time consumed in extracting all this information, and clustering them each time. In other words, the accuracy achieved of this method is not worth the amount of time consumed. This was the main motivation behind proposing the next lip localization method, which is called the "nearest colour" method.

## 4.2.2 The "nearest colour" lip localizing algorithm

The major contributor to the time consumed in the previous method is the clustering method, particularly using it several times. This method does not rely much on clustering, but instead uses one of the previous approaches to initially detect the lip pixels, then by taking all the available information about these pixels, the ROI is searched to find the nearest pixels to the initially detected pixels to emphasize and expand the area to include the whole detected lip area.

The question is, what method to use for the initial detection?

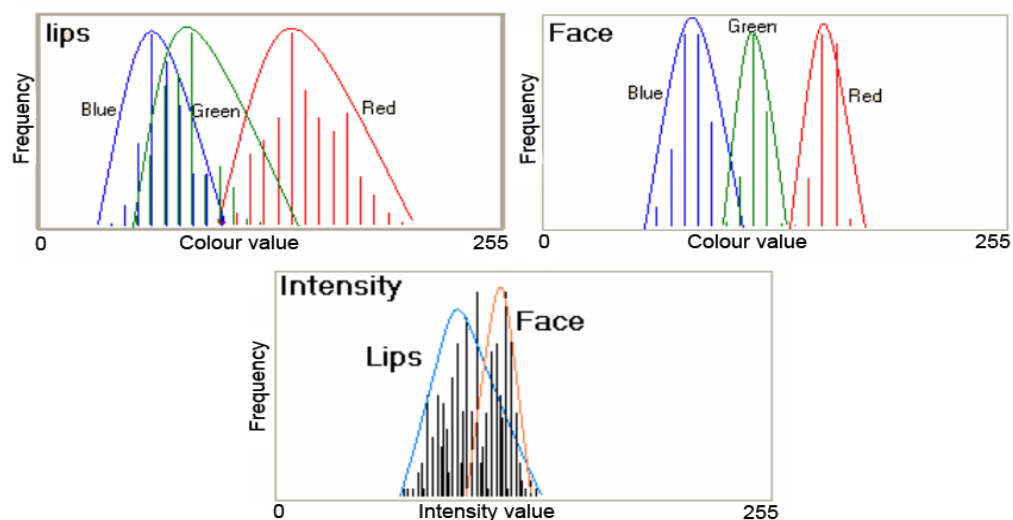

**Figure 4.9.** (left) RGB histogram for lip pixels, (right) RGB histogram for face pixels, (bottom) intensity histogram for both lips and face pixels.

For the RGB, luminance and chromatic information are mixed, and the RGB values of both the lip and the face are overlapped, i.e. it is difficult to extract the lip from the face using the RGB values of their pixels, see Figure 4.9.

For the Hue in the HSV, even after the warping process, the lip hue values are very close to the face hue values, see Figure 4.10.

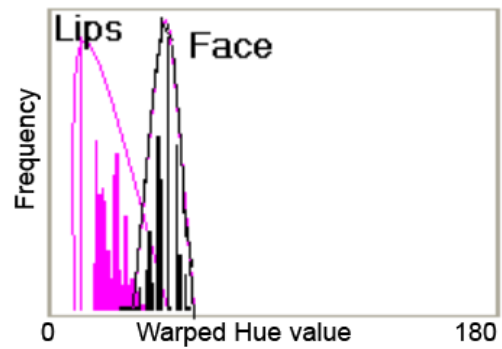

**Figure 4.10**. Warped Hue histogram for both lips and face pixels.

In YCbCr, chromatic and luminance components can be easily separated, lip pixels have high Cr and low Cb, and this is fostered by the Lip-Map equation, see Figure 4.11.

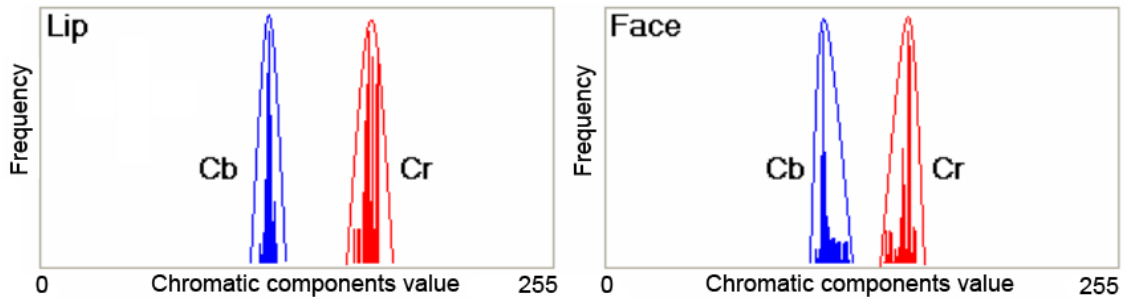

**Figure 4.11**. YCbCr chromatic components histogram. (left) lip pixels (right) face pixels.

The Cr component for the lips is slightly higher than that for the face, and the difference between Cr and Cb for the lip is higher than that for the face. The previous Lip-Map equation was designed to highlight these properties by increasing the output value where Cr is high and Cb is low; $Cr^2$ gives emphasis to pixels with higher Cr, and Cr/Cb completes the idea that lip regions have high Cr and low Cb values (Nasiri et al., 2008).

Although this formula does not work properly (it cannot detect the whole lip region) for various kinds of images, and it is not flexible under various lighting conditions, (Nasiri et al., 2008), (according to the experiments of this study) it is still better to be used as a starter for the proposed method rather than the other approaches. It can highlight at least (in the worst case) a small part of the lip, without being tricked by other details like the facial hair, which is enough for the initialization step of the proposed "nearest colour" algorithm.

For the reasons mentioned, the Lip-Map is used as an initial step for the "nearest colour" algorithm. The proposed "nearest colour" algorithm works as follows:

1. Apply the Lip-Map formula on the ROI pixels.

2. Initially, cluster the output to lip or non-lip pixels (see Figure 4.12 b).

3. For all the lip pixels calculate the mean of the r, g, b, warped hue, Cb, Cr, the X co-ordinate and the Y co-ordinate.

4. For each pixel in ROI, calculate the distance from the mean vector in step 3.

5. Sort the distances array, assuming that at least the highest half of the distances array belong to non-lip pixels.

6. For all the pixels of the highest distances, calculate the mean of the r, g, b, warped hue, Cb, Cr, the X co-ordinate and the Y co-ordinate.

7. Now we have 2 feature vectors, one representing the lip and the other representing the non-lip pixels.

8. For each pixel in ROI, calculate its distance to both feature vectors.

9. If the pixel is nearest to the lip feature vector then classify it as lip pixel, else classify it as face pixel.

10. Repeat 8 and 9 until no more pixels are clustered in ROI (see Figure 4.12c).

11. Create a binary image 1s representing the lip region, and apply the opening binary filter (erosion followed by dilation) to remove the noise (see Figure 4.13).

12. Consider the largest white area as the lips region.

13. In some cases where there are 2 large white neighbouring areas, consider both together as the lips region.

After executing the previous algorithm, the mouth box can be easily determined using the output binary image, see Figure 4.12d. The colour information calculated in step 3 is used to separate the lip pixels from the face pixels, depending on their colours, while the x and y coordinates are used to discard (reduce the effect of) the pixels that have the same colour as the lips, but far away from the segmented area, such as some of the nose and chin pixels.

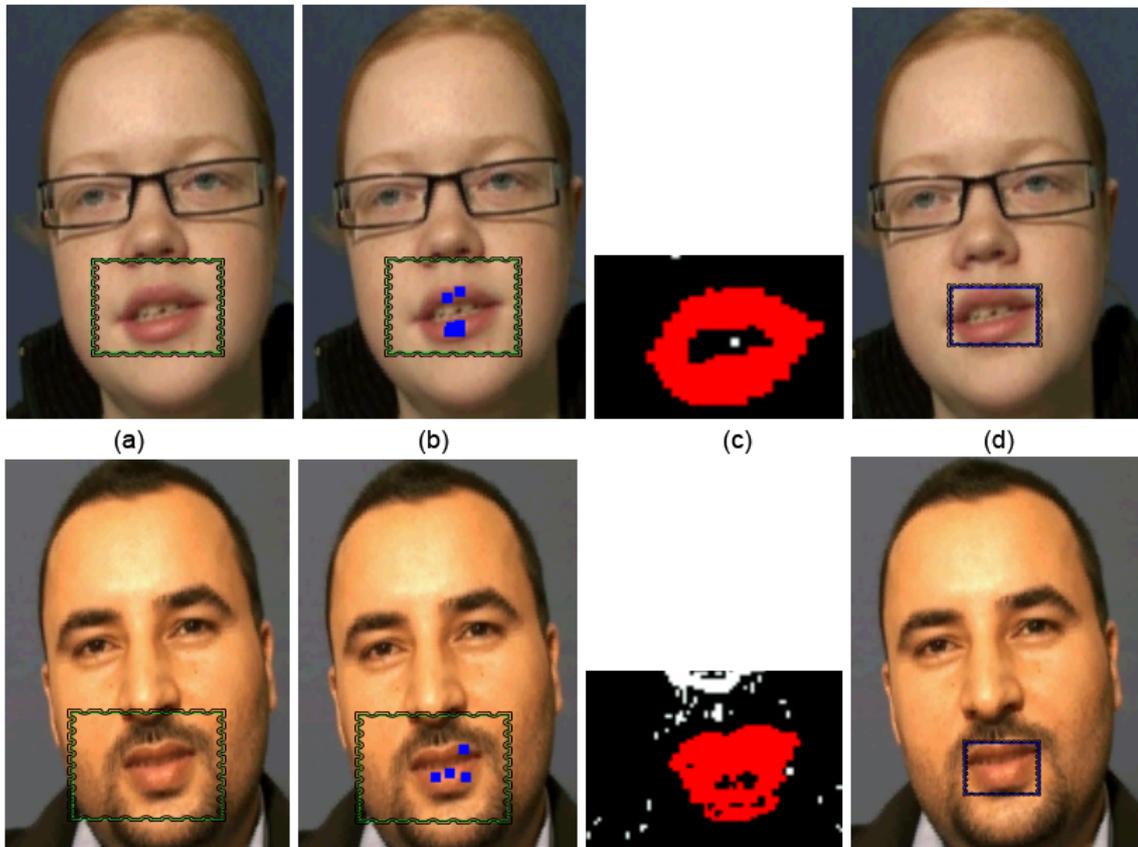

**Figure 4.12.** The different stages of the "nearest colour", a) face detection followed by ROI defining, b) initial clustering using the Lip-Map, c) binary image of ROI resulting from the nearest colour algorithm, d) final lip detecting.

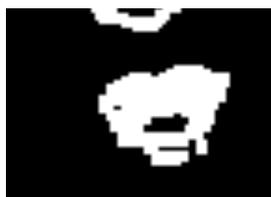

**Figure 4.13**. Applying the binary opening filter on the segmented ROI in Figure 4.12c.

### 4.2.3 K-mean clustering algorithm

The k-mean clustering method is an unsupervised learning method, which aims to partition a data set into *k* clusters or groups. Each data item is assigned to the group where its mean is the nearest to that data item. This method was proposed by Stuart Lloyd in 1957, and then published in 1982 (Lloyd, 1982).

The k-mean method works as follows:

1. Declare the number of clusters k, positive none zero integer.

2. Initialise centre point for each cluster.

3. Calculate the Euclidian distance for all the points from the centre points.

4. Label each point by the label of the minimum centre.

5. Update the centre points depending on the new distances, the centre point = the mean of all points of the same label.

6. Repeat 3-5 for a specific iteration number, or until no change is assigned to the centre points.

The main problem with this algorithm is the predefined number of clusters (k); if the number of clusters is not known, this method without modifications cannot be used. This method is used in this study to cluster pixels to lip and non-lip clusters, so the number of clusters is already known, which is k=2.

## 4.3 Experimental results and evaluation

The proposed methods and some of the mentioned approaches were programmed and tested to detect lips in a new-recorded video database (called the in-house database). This database was designed and recorded to evaluate the whole visual word recognition system. The results of this evaluation will enlighten us to choose the best approach for lip detection to be used within the proposed system.

### 4.3.1 In-house database

Since the beginning of this study, it was intended to use the PDADatabase for the evaluation of each of the proposed methods in this study. However, the PDADatabase was not created specifically to test the performance of lip-reading tasks. It was designed to test the performance of a biometric verification scheme using the fusion of audio-visual signals (as well as a handwritten signature) from a video clip of the clients while uttering short sentences. Due to the limited capabilities of the then commercially available PDA in recording and processing videos in real-time for more than a few seconds, the words were pronounced without leaving gaps between them, which causes the visual signal from several words to be merged into that of one word.

As the designers of the PDADatabase did not need to use lip reading when they created the database, we found it necessary to create a new video database to specifically help in designing and developing lip-reading schemes for VW recognition and eventually to evaluate such methods, rather than just recognizing the face, for instance.

The new database is called an "in-house" database. It was recorded and designed for the purpose of the VSR project, to be a substitute for the PDADatabase, which was not designed for this purpose.

26 participants of different races and nationalities (Africans, Europeans, Asians and Middle-Eastern volunteers), who were students and staff members from the Applied Computing Department in the University of Buckingham, took part in recording this database.

Each participant recorded 2 videos (sessions 1 and 2) at different times (about a 2 month period in time between the two recordings). The participants were 10 females and 16 males, distributed over different ethnic groups: 5 Africans, 3 Asians, 8 Europeans, and 10 Middle Eastern participants. The database contains 6 males with both beard and moustache, and 3 males with moustache only.

Each person in each recorded video utters 30 different words five times; the words are:

- Numbers from 0-9.

- Short look-alike words: (*knife, light, kit, night, fight*) and (*fold, sold, hold, bold, cold*).

- Long words: (*appreciate, university, determine, situation, practical*).

- For words that can be used as a security alert in video surveillance the author chose: (*bomb, kill, run, gun, fire*).

Video frame rate is 30 frames per second, and the average time for each video is about 8 minutes and 33 seconds. Each video contains about 15000 frames, so the total number of frames in the database ≈ 15,000 frame * 2 Sessions * 26 persons = 780,000 frames.

Total number of words = 30 different words * 5 times repeated * 2 Sessions * 26 persons = 7,800 words.

The videos were recorded inside a normal room, which was lit by a 500-watt light source, using Sony HDR-SR10E high definition (HD) 40GB Hard Disc Drive Handy-cam Digital Camcorder – 4 Mega Pixels. The HD 1920 x 1080 resolution video was designed to be played on TV screens, so there were no actual frames, but instead there were odd and even fields; the video signal is recorded each 1/60 second for the odd lines, and the same time for the even lines of the image; in each 1/60 second one of the fields is refreshed. It needs 1/30 second to get both fields to be refreshed, exactly like the TV broadcasting signal. So when this video is played on computers (which only understand frames language) for the sake of this study, the interlacing problem appeared, see Figure 4.14.

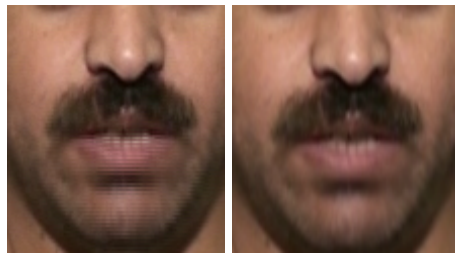

**Figure 4.14.** (left) shows the horizontal lines that reflect the movement of the lips as a result for the output interlaced video, (right) fixing the problem by de-interlacing (blending both fields).

The horizontal lines are the side effect of refreshing the odd and even fields; this problem appears when there is a movement, in our case it appears in the ROI where lips move. To solve this problem, a de-interlacing technique should be applied; there are different scenarios for de-interlacing such as:

- Blend both fields (odd and even) together.

- Duplicate the odd field and discard the even one.

- Duplicate the even field and discard the odd one.

- Discard the odd field leaving only the even one.

- Discard the even field leaving only the odd one.

In this study it was decided to use the first type of de-interlacing, because it is the best way to preserve as much information as possible by merging or blending both of the fields, see Figure 4.14 (right).

The video is then compressed using Intel IYUV codec, converted to AVI format, and resized to (320 x 240) pixels, because it is easier to deal with AVI format, and it is faster for training and analyzing the videos with smaller frame sizes, so the high definition HD images were not used in this study.

The previous process for the new database was partially done using the VirtualDub[*] open source licensed under a General Public Licence.

### 4.3.2 Experiments and results

No specific researcher image-based method was evaluated; instead, the whole approach, which is used by several researchers, has been evaluated by this study. For the image-based approaches, the evaluation of each was built on clustering the ROI pixels into lips or non-lips pixels; using the information obtained from that approach, the clustering method was the k-mean clustering algorithm.

For the model-based approach, ASM was evaluated on behalf of this approach. This time the whole method was applied as it is, using the MASM software, designed, and developed by Milborrow and Nicolls (2008). The method was the same as the one introduced by Cootes et al. (1992), but with adding more landmarks. Results are improved by fitting more landmarks, but search time increases linearly (Milborrow and Nicolls, 2008).

The method was tested to evaluate lip detection for about 780,000 video frames from the in-house database. The evaluation was done by human subjectively, by observing the detected lips allowing about 3 pixels plus/minus in all directions. If the mouth box was drawn to include the real mouth within this margin it is considered as true detection, otherwise, it is false detection.

It is a very difficult task to evaluate all the video frames manually; instead 10 random samples from each video was evaluated, each random sample of size 100 random video frames, and the result of the lip detection method is observed and approximated by human intelligence; the observer has to decide whether the lips are detected or not in each frame, then the observer averages the results for each video, then for each person.

---

[*] VirtualDub home page: http://www.virtualdub.org/index

Table 4.1 illustrates the results of detecting lips of all the subjects (26) in the in-house database, using all the evaluated lip detection methods.

**Table 4.1**. Results of lip detection using different approaches.

| Person | Hue | rgb | Hybrid Edge | YCbCr | ASM | Layers Fusion | Nearest Colour |
|---|---|---|---|---|---|---|---|
| *Female_01* | 67.74% | 45.83% | 97.56% | 84.09% | 22.73% | 90.65% | **100.00%** |
| *Female_02* | 64.37% | 39.02% | 71.76% | **100.00%** | 95.83% | 98.48% | **100.00%** |
| *Female_03* | 89.52% | 97.01% | 98.90% | 95.56% | 94.44% | **100.00%** | **100.00%** |
| *Female_04* | 92.11% | 88.46% | 85.29% | **100.00%** | 73.33% | 99.21% | **100.00%** |
| *Female_05* | 69.92% | 48.89% | 86.21% | **95.35%** | 90.48% | 93.33% | 89.86% |
| *Female_06* | 61.29% | 51.48% | 88.37% | 87.35% | **100.00%** | 92.86% | 87.80% |
| *Female_07* | 52.68% | 32.14% | **82.09%** | 43.90% | 54.17% | 68.61% | 61.90% |
| *Female_08* | 88.89% | 100.00% | 80.00% | 88.10% | 14.29% | **100.00%** | **100.00%** |
| *Female_09* | 93.88% | 86.00% | 93.26% | 88.10% | 50.00% | 79.55% | **100.00%** |
| *Female_10* | 85.00% | 73.91% | 88.68% | 78.57% | 83.33% | 99.24% | **100.00%** |
| *Male_01* | 62.50% | **86.89%** | 24.49% | 12.50% | 64.71% | 26.14% | 80.95% |
| *Male_02* | **100.00%** | 11.29% | 42.86% | 71.43% | 52.94% | 75.40% | **85.37%** |
| *Male_03* | 65.55% | 4.88% | 21.62% | 19.51% | **71.43%** | 30.00% | 68.18% |
| *Male_04* | **100.00%** | 98.48% | 92.86% | **100.00%** | 50.00% | **100.00%** | **100.00%** |
| *Male_05* | 72.66% | 47.73% | 54.88% | 85.71% | 4.35% | 91.13% | **92.86%** |
| *Male_06* | **89.89%** | 12.50% | 6.82% | 24.39% | 87.50% | 31.76% | 68.29% |
| *Male_07* | 70.97% | 2.50% | 13.79% | 10.00% | 6.25% | 59.02% | **85.37%** |
| *Male_08* | 85.71% | 36.17% | 89.36% | 96.30% | 93.75% | 81.61% | **100.00%** |
| *Male_09* | 91.20% | 25.64% | 40.00% | 52.50% | **100.00%** | 87.88% | **100.00%** |
| *Male_10* | 84.92% | 52.50% | 68.18% | 82.50% | 73.68% | 87.14% | **92.68%** |
| *Male_11* | 82.25% | 70.73% | 97.10% | 39.02% | 37.50% | 95.43% | **97.56%** |
| *Male_13* | 93.98% | 77.50% | 79.07% | 95.56% | 64.71% | **100.00%** | **100.00%** |
| *Male_14* | 96.25% | 20.93% | 86.05% | 32.50% | **100.00%** | 54.55% | 98.04% |
| *Male_15* | 79.07% | 33.33% | 75.00% | 27.50% | 84.21% | 66.37% | **100.00%** |
| *Male_16* | 14.06% | 70.00% | 80.20% | 25.00% | **100.00%** | 99.31% | 85.37% |
| *Male_17* | 63.79% | 8.89% | 40.44% | 43.59% | 0.00% | 58.43% | **75.61%** |
| average | **77.62%** | **50.87%** | **68.65%** | **64.58%** | **64.22%** | **79.47%** | **91.15%** |

As can be seen from Table 4.1, the "Layer Fusion" lip detection results are not much better than, for instance, the Hue approach; this small difference does not justify the time consumed, which is about 5 times more than the time consumed in the Hue approach. For both reasons, this method will not be used for the lip-detection step in the proposed VW system.

The in-house database, which was created to investigate the lip-reading problem, consists of several ethnic groups. Most of the investigated methods depend on the use of colour information as a key factor. Hence, the lip detection results should be affected by the variation in the skin and lip colour due to ethnicity. The results in Table 4.2 depict the performance of the different lips detection schemes (including ours) for the different ethnic groups.

Table 4.2. Categorical results of lip detection using different approaches

| Category / Method | Hue | RGB | Hybrid Edge | YCbCr | ASM | Layers fusion | Nearest Colour |
|---|---|---|---|---|---|---|---|
| **African** | 74.45% | 53.90% | **87.92%** | 72.44% | 42.30% | 78.72% | **87.60%** |
| **Asian** | 79.64% | 29.76% | 76.56% | 67.07% | 97.92% | 80.91% | **98.08%** |
| **European** | 73.95% | 86.44% | 78.17% | 83.01% | 62.67% | 95.60% | **96.34%** |
| **Middle Eastern** | 80.56% | 67.89% | 92.47% | 82.22% | 79.90% | 95.24% | **97.10%** |
| **Beard And Moustache** | 74.96% | 33.22% | 40.29% | 47.95% | 48.90% | 64.09% | **86.38%** |
| **Moustache** | 78.93% | 16.39% | 27.43% | 20.66% | 59.32% | 54.38% | **86.43%** |
| **Female** | 76.54% | 66.28% | 87.21% | 86.10% | 67.86% | 92.19% | **93.96%** |
| **Male** | 78.30% | 41.25% | 57.04% | 51.13% | 61.94% | 71.51% | **89.39%** |
| **Average** | **77.62%** | **50.87%** | **68.65%** | **64.58%** | **64.22%** | **79.47%** | **91.15%** |

Both tables show the superiority of the proposed "**nearest colour**" algorithm, with accuracy of about 91% and a significant difference (12%) from its next rival algorithm. This method is qualified more than the others to detect the lips in 91% confidence for the final step of this study (the lip reading). Figure 4.16 illustrates some correctly detected lips of some subjects in the in-house database, and Figure 4.17 illustrates the false detection.

The nearest colour method makes use of the YCbCr (just for the initial step) to get some of the real lips pixels. The colours of these pixels may differ from person to person, which is not important for this method, because it compares these pixels with the same image pixels, trying to find other similar pixels to declare them as lip pixels. The difference between the lips and the skin colour does not matter, because each pixel is classified depending on its distance to either the lips averages, or to the skin averages, regardless of how close these averages are.

For the other approaches, the larger (apparent) the difference between the lip colour and the skin colour, the more accurate the clustering method will be. An interesting result is the relatively good results of the Hue approach (77.62%), which is better than both the YCbCr, and the Hybrid edge approaches results. The Hue approach performance is better because it is not much affected by the appearance of the facial hair (about 75% to 79%), while it is less than 50% for both of the YCbCr and the Hybrid edge approaches. Because the hue of the facial hair is nearest to the skin colour rather than to the lip colour, this is what makes the Hue approach the only approach which performs better with males (78.30%) rather than females (76.54%).

Nevertheless, the proposed "nearest colour" method is affected less than the Hue approach when it comes to the facial hair (about 86%), because the facial hair colour values (including their hues) are closer to the skin colour.

Another interesting result is the (87.92%) accuracy rate of the Africans by the "Hybrid Edge," which is slightly higher than the "nearest colour" (87.60%), which is the worst ethnic group result compared to the other ethnic groups (more than 96%). This is because the colour of the lips and the skin for the Africans is almost similar, and very overlapped (see Figure 4.15). This is not the case when it comes to the other ethnic groups.

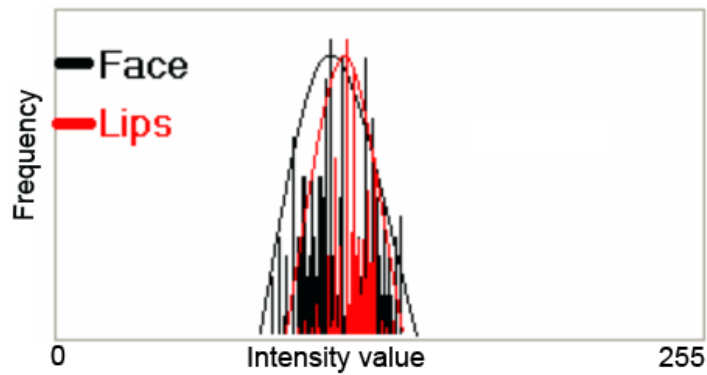

**Figure 4.15.** shows the intensity histogram of lips (red) and face (black) for African females.

The "Hybrid Edge" is backed up with the edges information as well as the hue information. So, using sources of information other than the overlapped colour, such as the edges, makes the results better for the Africans (87.92%).

In general, the female results are better for most of the methods, although most of the female participants were not wearing full makeup, but still the females' lips are more highlighted than males in general, and in the evaluation database females' lips are even highlighted by makeup or by nature. Another more important reason is the lack of facial hair in the case of females, which greatly affects the detection method.

The RGB approach gives the worst results, because luminance and colour information are mixed together in the 3 channels, the red, green and the blue. This makes such an approach greatly affected by light conditions, shadows, wrinkles and facial hair.

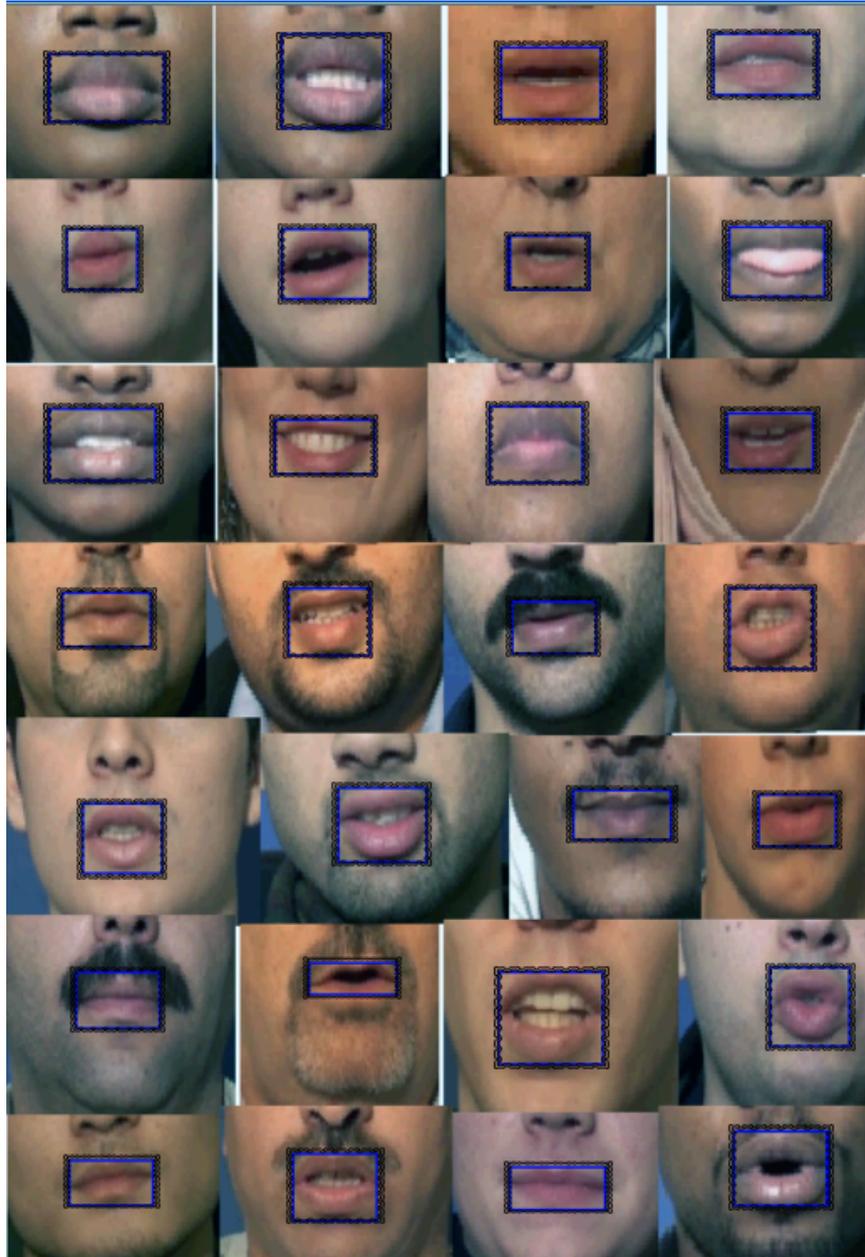

**Figure 4.16.** "Nearest Colour" algorithm accurate lips detection samples.

Representing the model-based approach, the ASM was tested and evaluated using the same database. Tables 4.1 and 4.2 show the results of the ASM facial features detection. Surprisingly, the performance of this method (64%) was much worse than the proposed methods (-27%), and worse than most of the image-based methods. Figure 4.18 illustrates accurate lip detection, and Figure 4.19 illustrates inaccurate lip detection for subjects from the in-house database.

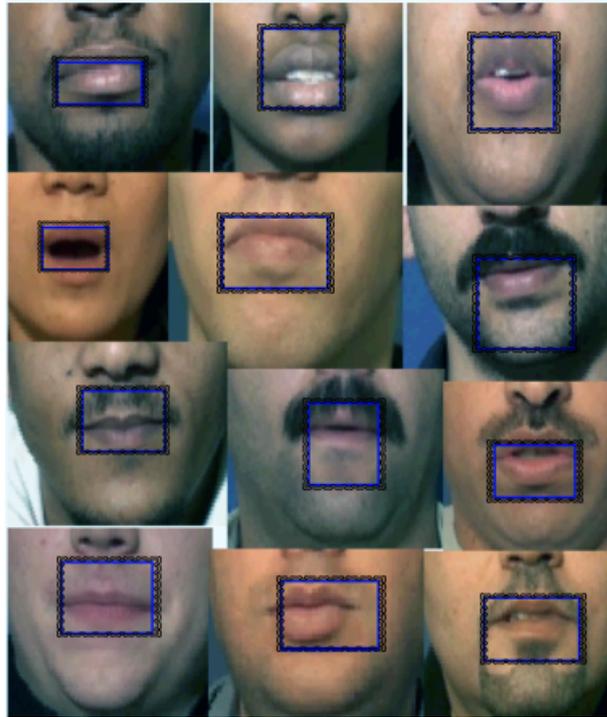

**Figure 4.17.** "Nearest Colour" algorithm inaccurate lips detection samples.

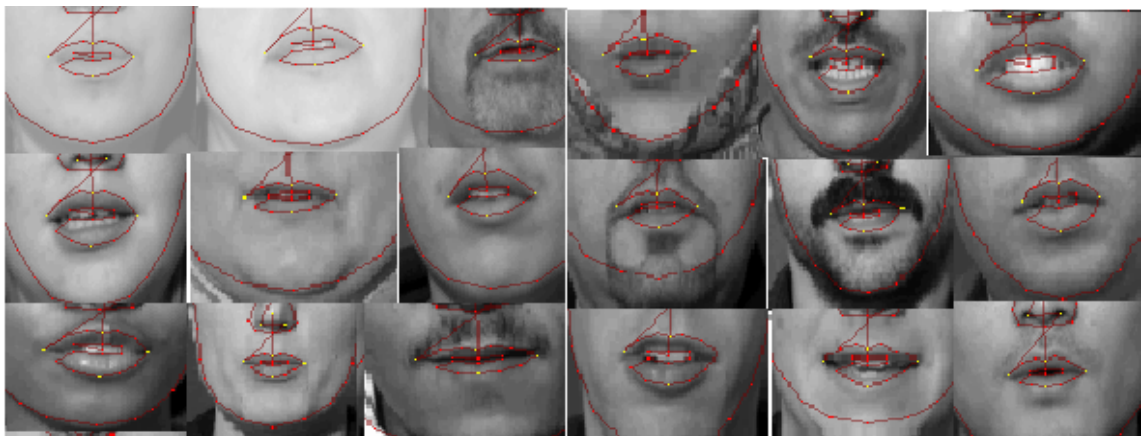

**Figure 4.18**. ASM accurate lip detection samples.

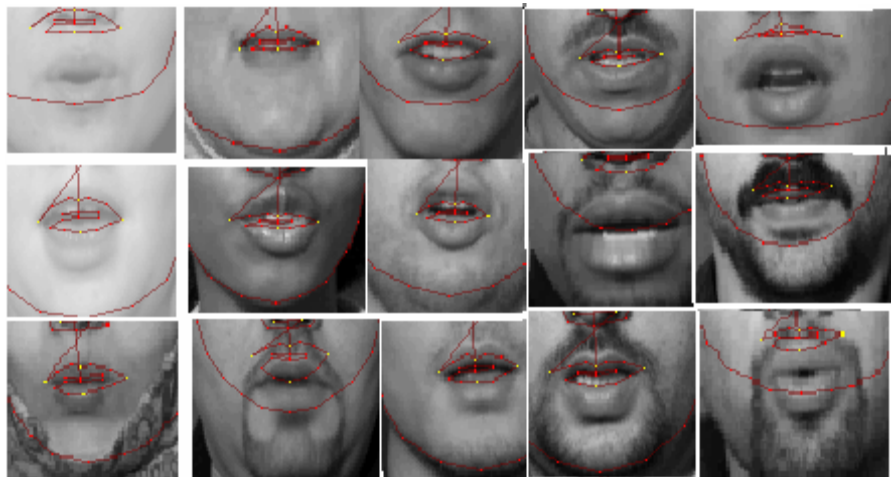

**Figure 4.19.** ASM inaccurate lip detection samples**.**

The bad performance for ASM is due to the problems associated with the use of model-based approaches in general, and the use of ASM in particular, which were mentioned earlier. This proves and justifies the use of an image-based approach instead, for the new proposed methods in this study.

The convergence to local minima is one of the most common problems of the ASM. This problem can be noted in Figure 4.19 where a local minimum happened to be one of the following:

1. The appearance of the facial hair, which makes some contours look like the lips contour.

2. or the inner contour of the mouth, especially when the mouth is widely opened.

3. or other facial features like the nose or the chin contour.

## 4.4 Summary

In this chapter we discussed some of the most well known state-of-the-art lips detection methods, divided into two main categories: the model-based methods, and the image-based methods. Some of these approaches were implemented and evaluated using a special database, which is recorded for the lip-reading problem investigation. We also proposed two new methods for lip detection: the "layer fusion", and the "nearest colour" method. The first did not perform very well and does take a long time. The results demonstrated that the nearest colour algorithm outperforms the state-of-the-art lips detection methods, and it can be applied in real time and online applications. As a result of this chapter, the author is opting to use this method to detect lips automatically for the lip-reading stage in the next chapter.

# Chapter 5

# Visual Word Recognition (VWR)

Most Visual Speech Recognition (VSR) systems referred to in the literature are based on visemes. In such a system each word is viewed and recognized by its visemes combinations. Using visemes in visual speech recognition systems raises several problems including the mapping of one viseme to many phonemes, the mapping of many visemes to one phoneme, and the difficulty of identifying visemes (see section 2.2.3 for more details).

To eliminate the aforementioned problems in VSR systems, we propose a new VSR approach dependent on the signature of the word itself, rather than the current trend that consists of recognizing visemes first then synthesizing these components to recognize the word. The proposed scheme uses a hybrid feature extraction method that combines geometric, appearance, and image transformed features. The new proposed VSR approach is termed "visual words" (see Figure 5.1).

The rest of this chapter is devoted to describing the proposed VSR approach in detail, and consists of two main parts: the proposed features extraction method that extracts the best features to represent a spoken word, and the classification procedure that recognizes a spoken word. The time complexity of the proposed method is discussed at the end of the chapter.

## 5.1 The proposed lip reading method

After detecting the face (see chapter 3) and localizing the lips (see chapter 4), the real task of the proposed visual words system begins. Like other VSR systems, the proposed visual words system consists of two main parts: (see Figure 5.1)

  1- Feature extraction / selection.

  2- Visual speech feature recognition.

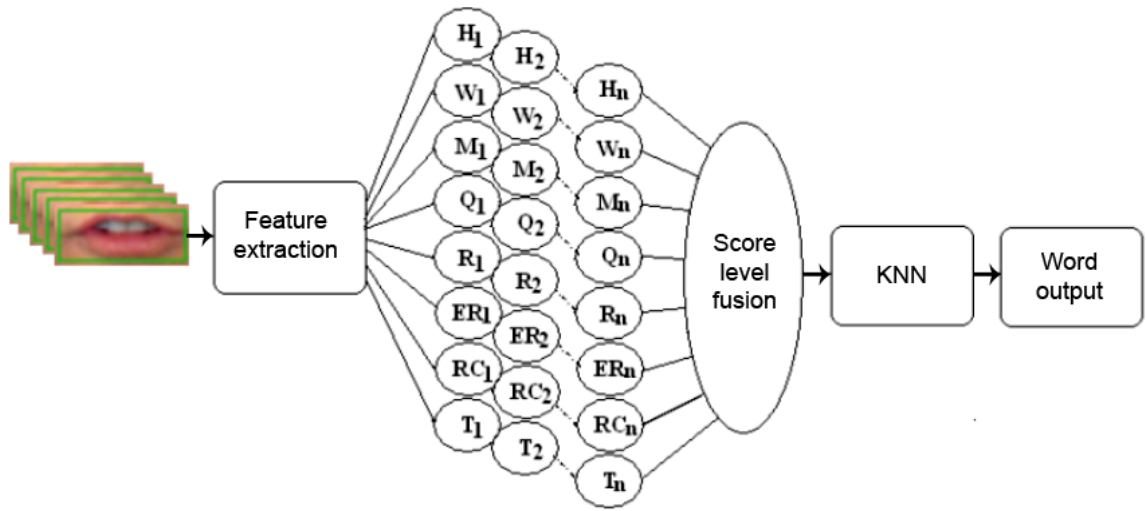

**Figure 5.1**. The proposed feature extraction and recognition method.

As can be noticed from Figure 5.1, for each spoken word, 8 feature vectors of length *n* (number of frames) are extracted, forming 8 different signals; *($H_1$ to $H_n$: the mouth Height's signal), ($W_1$ to $W_n$: the mouth Width's signal), ($M_1$ to $M_n$: the Mutual's signal), ($Q_1$ to $Q_n$: the Quality's signal), ($R_1$ to $R_n$: the DWT vertical to horizontal features ratio's signal), ($ER_1$ to $ER_n$: the Edge ratio's signal), ($RC_1$ to $RC_n$: the Red colour's signal)* and *($T_1$ to $T_n$: the Teeth's signal)*.

## 5.1.1 Feature extraction / selection

Visual speech/word recognition systems require the analysis of feature vectors extracted from the speech-related visual signals in regions of interest (ROI) in the sequence of the speaker face frames while uttering the spoken word/speech. Like other visual pattern recognition problems, VSR (and VWR) systems depend on extracting features from the sequence of ROIs in the video frames beyond their sets of "normalized" spatial/frequency coefficients. In fact, representing ROI's by their spatial/frequency content alone often results in several problems:

1. Large amounts of data need to be processed by the recogniser, requiring significant computational powers and resources (the "*curse of dimension*").

2. The relevant data will be hidden in piles of redundant data.

3. The large variety in the pixels values of the mouth ROI makes it very difficult to generalize, i.e. it would be very difficult to assign a sequence of images for each word, for instance.

4. The image of the mouth region is vulnerable to translation, rotation, scale, illumination, shadow and noise. A good feature extraction method should yield invariant features to such factors.

To alleviate the effect of these problems on the recognition rate, the ROI's appearance must not be taken as a whole. Instead, more specific and speech-related features should be extracted from the ROIs in the frame sequence associated with the spoken word, in order to obtain a meaningful and distinguishing representation of that word.

Ideally, the required feature representations of words must capture specific visual information that is closely associated with the spoken word, to enable the recognition of the word and distinguish it from other words. Unlike the visemic approach, the proposed visual words technique depends on finding a signature for the whole word, instead of recognizing each part (viseme) of the word alone. To find such a signature, or a signal for each word, we need to find a proper way of extracting the most relevant features, which play an important role in recognizing that word.

Feature extraction schemes used in VSR systems can be classified as appearance-based, shape-based or hybrid schemes that extract a combination of appearance and shape features. The appearance-based feature extraction approach assumes that visual speech features are encapsulated in the pixel values, or their corresponding frequency information, within the ROI of the mouth region alone. Shape-based feature extraction schemes, on the other hand, assume that visual speech features are embedded in the shapes of the speaker's lips or within the wider facial features. Hybrid schemes that concatenate both previous types of features, such as Active Appearance Models (Neti, et al., 2000) have also been proposed to improve the performance of VSR systems.

For VSR systems, researchers have reported different feature extracting approaches: the geometric-based features, such as the height and the width of the mouth, appearance-based features, such as the appearance of the tongue and the teeth, image-transformed-based features, such as transforming the ROI to the discrete wavelet domain (DWT), and hybrids of the aforementioned features.

An appearance-based approach to visual speech feature extraction ignores the fact that mouth appearance varies from one person to another (even when two persons speak the same word). Thus, using the appearance-based feature extraction alone does not take individual differences into consideration, and leads to inaccurate results. Moreover, appearance-based feature extraction methods mostly lack robustness in certain illumination and lighting conditions (Jun and Hua, 2009).

In this thesis we adopt the hybrid-based approach and we expand on the list of features beyond traditionally adopted ones such as the height and width of the speaker's lips. Indeed there is valuable information encapsulated within the ROI that has a significant association with the spoken word, e.g. the appearance of the tongue and teeth in the image during the speech. The appearance of the teeth (for instance) occurs while uttering specific phonemes (the dentals and labio–dentals).

At the same time, focusing only on the image-based features (appearance and transformed-based features) yields image-specific features, and it is sometimes difficult to generalize about those features on other videos or speakers. These results are backed up by Jun and Hua (2009).

The visual signal associated with a phoneme is rather short and hence their visual features are extracted from "representative" image frames. However, the visual signals associated with words are of longer duration involving tens of frames that vary in many ways. Hence the need to supplement/modify the set of features used in a visemic system by including some features relating to variation of frames along the temporal axis. There are many ways to represent such features, but we shall include two seemingly obvious features: an image quality parameter that measures the deviation/distortion of any frame from its predecessor, as well as the amount of mutual information between a frame and its predecessor. Such features are expected to compensate for the fact that many words do share some phonemes.

The list of features adapted in this thesis is by no mean exclusive, but it was limited out of a desire to minimize the number for efficiency purposes and to a manageable set of features for which their impact on the accuracy of the intended VW system can be estimated experimentally. The following is the proposed list of features that will be extracted from the sequences of the ROIs of the mouth areas during the uttering of the word (see Figure 5.1):

1. The height (H) and width (W) of the mouth, i.e. ROI height and width (geometric-based features).
2. The mutual information (M) between consecutive frames ROI in the DWT domain (image-transformed-based features based on temporal information).
3. The image quality value (Q) of the current ROI with reference to its predecessor measured in the DWT domain (image-transformed-based features based on temporal information).
4. The ratio of vertical to horizontal features (R) taken from DWT of ROI (image-transformed-based features based on temporal information).
5. The ratio of vertical edges to horizontal edges (ER) of ROI (image-transformed-based features).
6. The amount of red colour (RC) in ROI as an indicator of the appearance of the tongue (image-appearance-based features).
7. The amount of visible teeth (T) in the ROI (image-appearance-based features).

The proposed feature extraction method produces 8 signals for each uttered word, creating 8-dimensional feature space. Those signals maintain the dynamic of the spoken word, which contains a good portion of information; on the contrary, the visemic approach does not take into consideration the dynamic movement of the mouth and lips to produce a spoken word (Yu, 2008). Accordingly, for each word we would extract a time-series of 8-dimensional vectors. The main difficulties in analysing of these time-series stem from the fact that their lengths not only differ between the spoken words, but also differ between different speakers uttering the same word and between the different occasions when the same word is uttered by the same speaker.

In what follows we describe each of the 8 features. We assume that for each frame of a video, the speaker face is first localized using the method proposed in chapter 3, and then the third tri-sector of the detected face region is tested for the nearest colour lip localization method as described in chapter 4 to determine the ROI (of the mouth and the lips) from which these features are extracted (see Figure 5.2).

**5.1.1.1 The height and width features of the mouth**

Some VSR studies used lip contour points as shape features to recognise speech. For example Wang et al. (2004) used a parameter set of a 14 points ASM lip to describe the outer lip contour. In addition, Sugahara et al. (2000) employed a sampled active contour model (SACM) to extract lip shapes. Determining the exact lips contour is rather problematic due to the little differences in the pixel values between the face and the lips. Here, we argue that *it is not necessary (redundant) to use all or some of the lip's contour points to define the shape of the lips, where the height and width of the mouth backed up with a bounding ellipse is enough to approximate the real outer contour of the lips* (see Figure 5.3).

In our proposal, the height and width of the mouth area is determined by an approximation of the minimal rectangular box that contains the mouth area. The corners of the detected mouth box will be eliminated using the assumption that the mouth shape is the largest ellipse inside the minimal box. This assumption also helps in reducing redundant data when extracting the other features in the scheme.

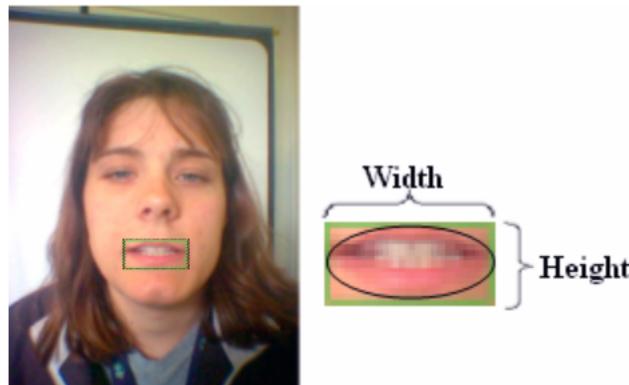

**Figure 5.2**. Lips geometric feature extraction; width and height.

Sometimes lips are not horizontally symmetric due to different ways of speaking, the ellipse assumption forces such symmetries, and alleviates such differences between individuals (see Figure 5.3).

Therefore, the exact ROI where the relevant computation of the features is conducted is actually the area inside the ellipse. The width of the detected box is the width (feature) of the mouth, and the height of the detected box is the height (feature) of the mouth. Here, we are assuming that all faces are nearly upright.

The width and height of the mouth are variables whose values depend on the uttered phoneme; some phonemes akin to (the central open vowel) [a], the mouth is opened to the maximum, i.e. the height expands. Other phonemes akin to (the front half closed unrounded vowel) [e], the width expands to the maximum, and some other phonemes like (the back half closed rounded vowel) [o] make the mouth shape circular, i.e. the width and the height of the mouth are almost equal.

These changes in height and width of the mouth create the two signals (W and H) that represent changes during the time of uttering a specific word. These two signals are values added to the signature of each word in the feature space.

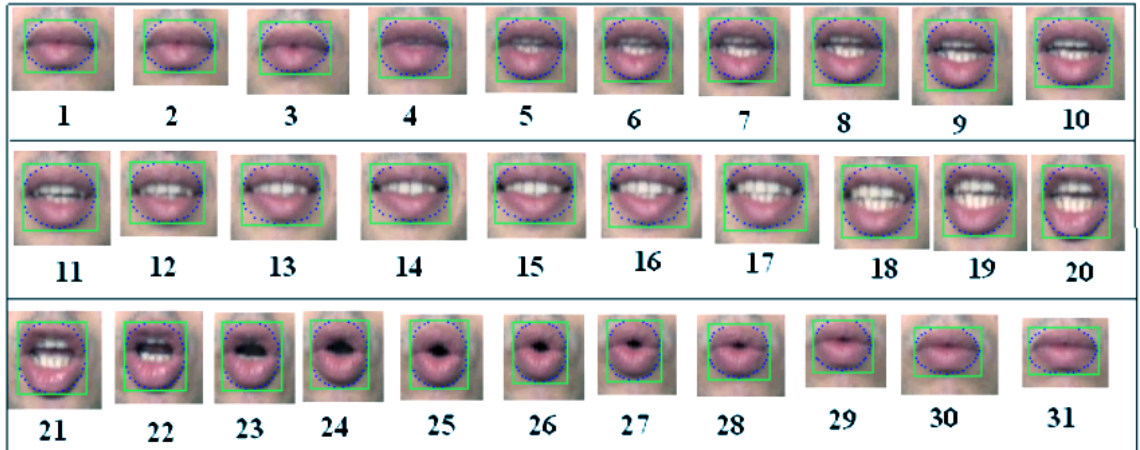

**Figure 5.3.** The change of the mouth shape while uttering the word "*Zero*", the blue dotted ellipse shows the approximated lip contour using the ellipse assumption.

The inner width and height of the mouth are more informative than the outer width and height, because some times these values change while speaking and the outer values do not change. But this is not significant, and occurs less frequently. We found that it is much easier and more accurate to detect the outer contour of the lips rather than the inner contour which is highly affected by the appearance of the tongue and the teeth.

**5.1.1.2 The mutual information (M) feature**

Mutual information between 2 random variables X and Y defines the dependency of these variables, i.e. mutual information reveals how much X contains information about Y, and *vice versa*.

If X and Y are independent from each other, their mutual information vanishes, but if they are identical then the mutual information between them will be high. Therefore, the value of the mutual information between any two random variables is in the range of zero to a maximum value. Mutual information can be utilized to quantify the temporal correlation between frames of a video sequence, so it can calculate the amount of redundancy between any two frames (Khayam, 2003). The temporal change of the appearance of ROI is caused by uttering a new/different phoneme. For example, the mouth appearance will change while switching from phoneme [e] to phoneme [d] when uttering the word "*feed*". Therefore, it is sensible to use the mutual information to measure some aspects of the change in the mouth area between consecutive ROIs.

The mutual information *M* between two random variables X and Y is defined by:

$$M(X;Y) = \sum_x \sum_y p(x,y) \log\left(\frac{p(x,y)}{p(x)p(y)}\right) \quad \cdots (5.1)$$

where *p(x,y)* is the joint probability mass function (PMF) of random variables *X* and *Y* (in our case mouth image (ROI) in current frame *X*, and previous mouth image in frame *Y*), *p(x)* and *p(y)* represent the marginal PMF of *X* and *Y* respectively.

In other words, *p(x)* and *p(y)* are the normalized histogram for frames *X* and *Y* respectively, and *p(x,y)* is the normalized joint frequency histogram of the two frames *X* and *Y* (Khayam, 2003).

To use the mutual information formula, the size of both of the random variables must be the same, but because the height and width of ROI are changing over time while uttering different phonemes, consecutive ROIs might not be of the same size. To solve this problem, the smaller ROI is stretched to fit the bigger ROI, or both ROIs are scaled to a predefined size, say 50 x 50 pixels.

Computing the mutual information in the spatial domain is inefficient and is influenced by many factors including the presence of noise and variation in lighting conditions. Instead, measuring the mutual information in the frequency domain provides a more informative mechanism to model changes between successive ROIs in different frequency sub-bands. Here we apply the DWT on both the current and the previous ROI. The mutual information formula is applied 4 times, one for each wavelet sub-band, and the average of the four values is taken as the mutual information feature for that frame or ROI (see Figure 5.4). Transforming both ROIs into the wavelet domain helps to reduce the effect of noise and variation in lighting conditions. For simplicity and efficiency, the DWT decomposition of the ROIs is implemented using the Haar filter.

### 5.1.1.3 The quality measure (Q) feature

There are many image quality measures proposed in the literature. Most of them attempt to find the amount of distortion in one image by referring to another image. Unlike the mutual information measurement, which attempts to measure the amount of dependency or similarity between two images, quality measures attempt to measure how different one image is from another.

Thinking again of the consecutive ROIs, a quality measure between them can tell something about change/distortion occurring due to an uttered phoneme. Therefore, any distortion in the current ROI, as compared to the previous ROI, is an indicator of changes in the structure of the mouth region. The amount of distortion can be measured by a quantitative quality measure, and considered as a feature at that frame or ROI.

This study utilizes a universal image quality index proposed by Wang and Bovik (2002) because it is a fast mathematical quality measure, and models image distortion as a combination of loss of correlation, luminance distortion, and contrast distortion (Wang and Bovik, 2002). The combined distortion of these three factors can be computed using one formula, and gives a clue to the changes in the consecutive ROIs, particularly the loss of correlation. The quality measure Q is given by:

$$Q = \frac{4\sigma_{xy}\bar{x}\bar{y}}{(\sigma_x^2 + \sigma_y^2)[(\bar{x})^2 + (\bar{y})^2]} \qquad \cdots(5.2)$$

where $Q \in [-1,1]$,

$$\bar{x} = \frac{1}{N}\sum_{i=1}^{N} x_i \quad , \quad \bar{y} = \frac{1}{N}\sum_{i=1}^{N} y_i \quad , \quad \sigma_x^2 = \frac{1}{N-1}\sum_{i=1}^{N}(x_i - \bar{x})^2 \quad ,$$

$$\sigma_y^2 = \frac{1}{N-1}\sum_{i=1}^{N}(y_i - \bar{y})^2 \quad , and \quad \sigma_{xy}^2 = \frac{1}{N-1}\sum_{i=1}^{N}(x_i - \bar{x})(y_i - \bar{y})$$

The best value for Q is when there is no distortion in the current ROI compared to the previous ROI; the value then is equal to 1 or -1 and the maximum distortion is measured when Q = 0.

This formula is very sensitive to luminance, because it models image distortion as luminance distortion, as well as loss of correlation and contrast distortion. This problem is solved by using the same approach that was used for the mutual information feature, i.e. using the DWT decomposed ROIs. Again, 4 quality measures (Q) are computed, one for each wavelet sub-band (the HH, HL, LH, and the LL). Then the average of the four values is taken as the quality measure feature for that frame or ROI (see Figure 5.4). For compatibility, we also use the Haar filter and the ROIs are both scaled to 50 x 50 pixels.

The average of both the mutual M, and the quality Q features are defined by:

$$M_i = \frac{M(LL_i;LL_{i-1}) + M(HL_i;HL_{i-1}) + M(LH_i;LH_{i-1}) + M(HH_i;HH_{i-1})}{4} \cdots (5.3)$$

$$Q_i = \frac{Q(LL_i;LL_{i-1}) + Q(HL_i;HL_{i-1}) + Q(LH_i;LH_{i-1}) + Q(HH_i;HH_{i-1})}{4} \cdots (5.4)$$

where $M_i$ and $Q_i$ are the mutual and quality features at frame $i$ respectively, $LL_i$, $HL_i$, $LH_i$ and $HH_i$ are the wavelet sub-bands of the current ROI, and $LL_{i-1}$, $HL_{i-1}$, $LH_{i-1}$ and $HH_{i-1}$ are the wavelet sub-bands of the previous ROI.

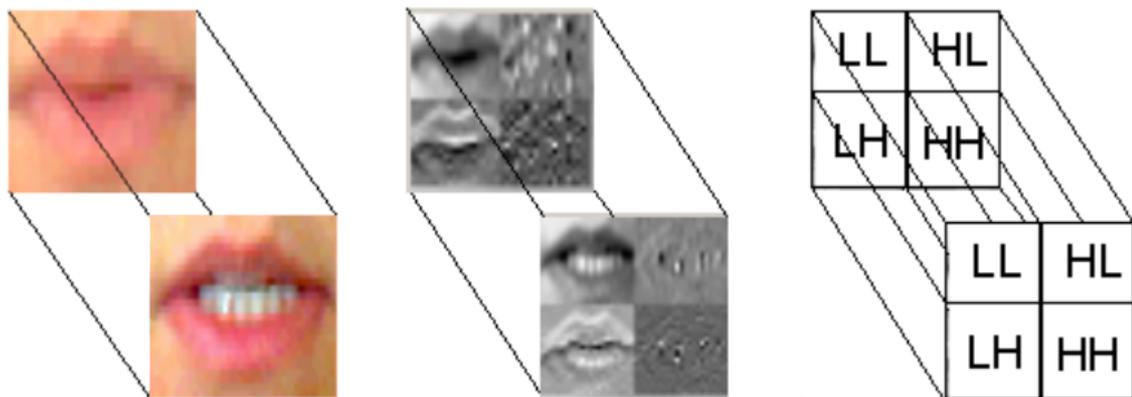

**Figure 5.4**. (1st row) The previous mouth and its Haar wavelet, (2nd row) Current mouth and its Haar wavelet.

### 5. 1.1.4 The ratio of vertical to horizontal features (R)

The DWT of an image I, using any wavelet filter, then histogram of the approximation sub-band LL approximates that of the original image while the coefficients in each of the three other sub-bands have a Laplacian distribution with 0 means, (Al-Jawad, 2009). Moreover, in each non-LL-sub-band the further away from the mean a coefficient is, the more likely it is associated with a significant image feature such as edges/corners. In other words, in each of the non-LL sub-bands the redundant data are located on and around the sub-band's Laplace distribution mean. This fact has been used by Al-Jawad to classify coefficients in non-LL sub-bands as **significant** if they are away from the mean by more than the sub-band standard deviation, **non-significant** otherwise. The LH wavelet sub-band highlights the horizontal features in images, while the HL wavelet sub-band highlights the vertical features in images (see Figure 5.5).

Here we adopt the above approach to identify feature-related pixels as the significant coefficients in the Non-LL sub-bands, i.e. the feature points are the ones with values greater than (median + standard deviation), or less than (median – standard deviation). The ratio (R) of the vertical features obtained from wavelet sub-band HL to the number of the horizontal ones gained from the LH is given by:

$$R = \frac{V}{H} \quad \cdots (5.5)$$

where V = number of vertical features, and H = number of horizontal features.

Accordingly, by substituting V and H in equation 5.5, we get equation 5.6.

$$R = \frac{\sum_x \sum_y \begin{cases} 0 & (HL_{median} - \sigma_{HL}) \leq HL(x,y) \leq (HL_{median} + \sigma_{HL}) \\ 1 & otherwise \end{cases}}{\sum_x \sum_y \begin{cases} 0 & (LH_{median} - \sigma_{LH}) \leq LH(x,y) \leq (LH_{median} + \sigma_{LH}) \\ 1 & otherwise \end{cases}} \quad \cdots (5.6)$$

where $HL_{median}$ and $LH_{median}$ are the medians of the wavelet sub-band *HL* and *LH* respectively, *HL(x,y)* and *LH(x,y)* the intensity value at location (x,y) in both *HL* and *LH* wavelet sub-bands, $\sigma_{HL}$ and $\sigma_{LH}$ are the standard deviation in both of the mentioned sub-bands.

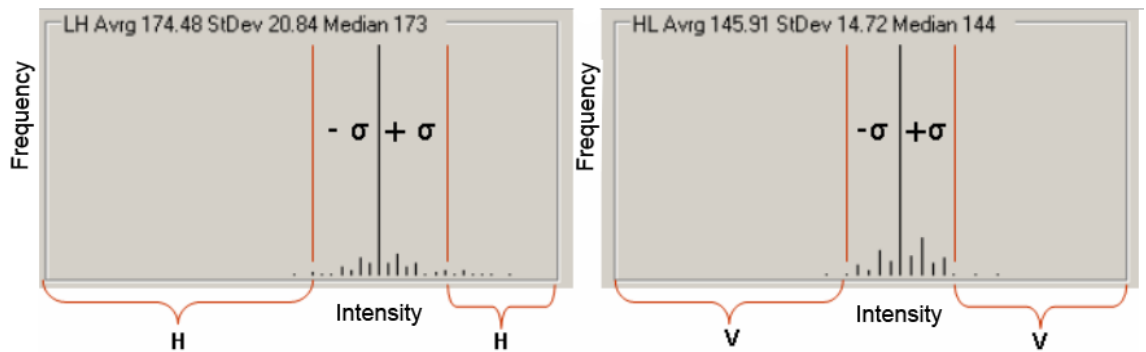

**Figure 5.5**. (Left), LH wavelet sub-band histogram, H represents the horizontal features. (Right), HL wavelet sub-band histogram, V represents the vertical features.

Figure 5.6 demonstrates the correlation between the mouth appearance and its ratio (*R)* property while speaking.

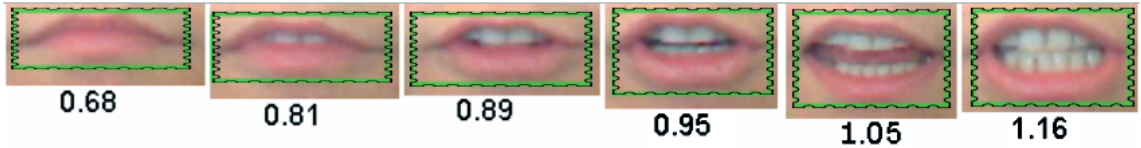

**Figure 5.6**. The co-relation between the mouth appearance and its ratio (*R*).

As can be seen from Figure 5.6, the ratio R is high when there are a lot of vertical features of ROI compared to the horizontal ones (the mouths on the right), and R is low when vertical features are low and/or horizontal features are high (the mouths on the left).

### 5.1.1.5 The ratio of vertical edges to horizontal edges (ER)

The ratio of vertical edges to horizontal edges (ER) of ROI is obtained by using the Sobel edge detector mentioned in chapter 3. This detector has two filters; the horizontal filter ($S_h$) highlights the horizontal edges of ROI, while the vertical filter ($S_v$) highlights the vertical edges of ROI.

The output value of any of the edge filters in a specific location demonstrates whether the pixel in that location is an edge pixel or not; the higher the filter output is, the more likely it will be an edge. The summation of the absolute values of the vertical filter demonstrates the amount of vertical edges in the ROI. In addition, the summation of the absolute values of the horizontal filter demonstrates the amount of horizontal edges in the ROI.

The ratio of the vertical edges to the horizontal ones is given by:

$$ER = \frac{\sum_{x=1}^{W}\sum_{y=1}^{H}\sum_{i=-1}^{1}\sum_{j=-1}^{1}\left|ROI(x+i, y+j)(S_v(i+1, j+1))\right|}{\sum_{x=1}^{W}\sum_{y=1}^{H}\sum_{i=-1}^{1}\sum_{j=-1}^{1}\left|ROI(x+i, y+j)(S_h(i+1, j+1))\right|} \quad \cdots (5.7)$$

where ROI(x,y) is the intensity value at the location (x,y) of the mouth region, W is the width of ROI, H is the height of ROI. $S_v$, and $S_h$ are Sobel vertical and horizontal filters respectively.

When the mouth is stretched horizontally, the amount of horizontal edges increases, so ER decreases. When the mouth is opened, the amount of vertical edges tends to increase, and this increases the ER. The appearance of the teeth contributes to both the vertical edges through the edges between teeth, and contributes to the horizontal edges through the horizontal teeth lines. In addition, the appearance of the tongue contributes to this kind of measurement (see Figure 5.7).

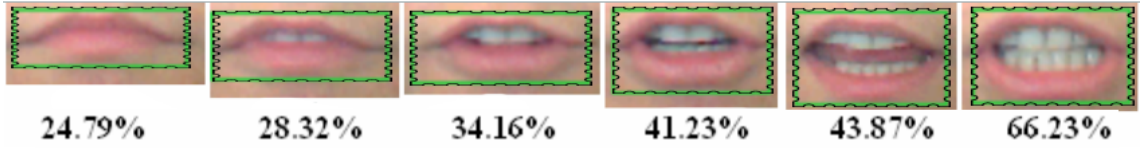

24.79%   28.32%   34.16%   41.23%   43.87%   66.23%

**Figure 5.7**. The co-relation between the mouth appearance and the ER feature.

Therefore, ER reveals something about the appearance of the mouth at a particular time. That is the reason why it is used as a feature for the proposed visual words for lip reading method.

### 5.1.1.6 The appearance of the tongue (RC)

Some phonemes like [T] involve the appearance of the tongue, i.e. moving the tongue and showing it helps to utter such phonemes. Therefore detecting the tongue in the ROI reveals something about the uttered phoneme and, by implication, the visual word.

However, it is difficult to model the tongue; the only available cue is its red colour. Therefore, the amount of red colour (RC) in the ROI will be taken to represent the appearance of the tongue, as well as the lip colour. Since the lip is captured within the ROI, the change in the amount of the red colour is then a cue for the appearance of the tongue. The different size of the tongue and lip from person to person is not problematic, hence all the features are scaled to the range [0,1], and the ratio of the red colour to the size of ROI is considered. This ratio can be calculated using the following equation:

$$RC = \frac{\sum_{x=1}^{W}\sum_{y=1}^{H} \text{Re}d(ROI(x,y))}{(W)(H)} \quad \cdots (5.8)$$

where Red(ROI(x,y)) is the red component value of the RGB colour system at the location (x,y) of the mouth region, W is the width of ROI, H is the height of ROI.

### 5.1.1.7 The appearance of the teeth (T)

Some phonemes like [s] incorporate the appearance of the teeth, i.e. showing teeth helps to utter such phonemes. Therefore detecting teeth in the ROI is a visual cue for uttering such phonemes and enriches the visual words signatures.

The major characteristic that distinguishes teeth from other parts of the ROI is the low saturation and high intensity values (Goecke et al., 2000). By converting the pixels values of ROI to 1976 CIELAB color space ( L*, a*, b*) and 1976 CIELUV color space (L*, u*, v*), the teeth pixel has a lower a* and u* value than other lip pixels (Liew, et al., 2003). A teeth pixel can be defined by:

$$t = \begin{cases} 1 & a^* \leq (\mu_a - \sigma_a) \\ 1 & u^* \leq (\mu_u - \sigma_u) \\ 0 & otherwise \end{cases} \quad \cdots (5.9)$$

where $\mu_a, \sigma_a$ and $\mu_u, \sigma_u$ are the mean and standard deviation of a* and u* in ROI respectively. The appearance of the teeth can be defined by the number of teeth pixels in ROI. Therefore, the amount of teeth in ROI is given by:

$$T = \sum_{x=1}^{W}\sum_{y=1}^{H} t(x, y) \quad \cdots (5.10)$$

Figure 5.8 demonstrates the amount of the teeth measure (T) compared to the appearance of ROI. The relation between T and the actual appearance of the teeth can be noticed.

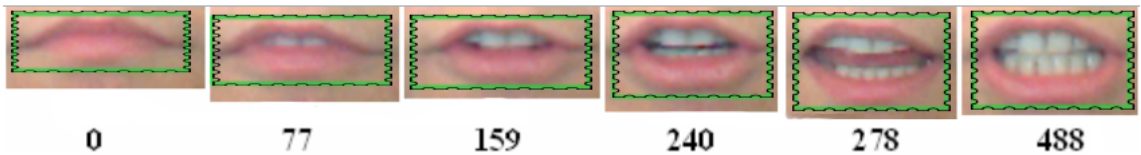

**Figure 5.8**. The co-relation between the teeth appearance and T feature.

As can be noticed from Figure 5.8, the second and third ROIs have almost the same shape, width, and height, except the teeth appearance, which is more obvious in the 3rd one. The same thing applies to the last two ROIs. This shows how important is the T feature for describing the appearance of the ROI while uttering different phonemes, particularly when other features do not make any difference.

## 5.1.2 The classification process

All the previous features are normalized to the range [0,1] to alleviate the individual differences, and different scales of mouth caused by different distances from the camera, i.e. the different sizes of ROIs. For each property, a feature vector (a signal) is obtained to represent the spoken word from that feature perspective (see Figure 5.9).

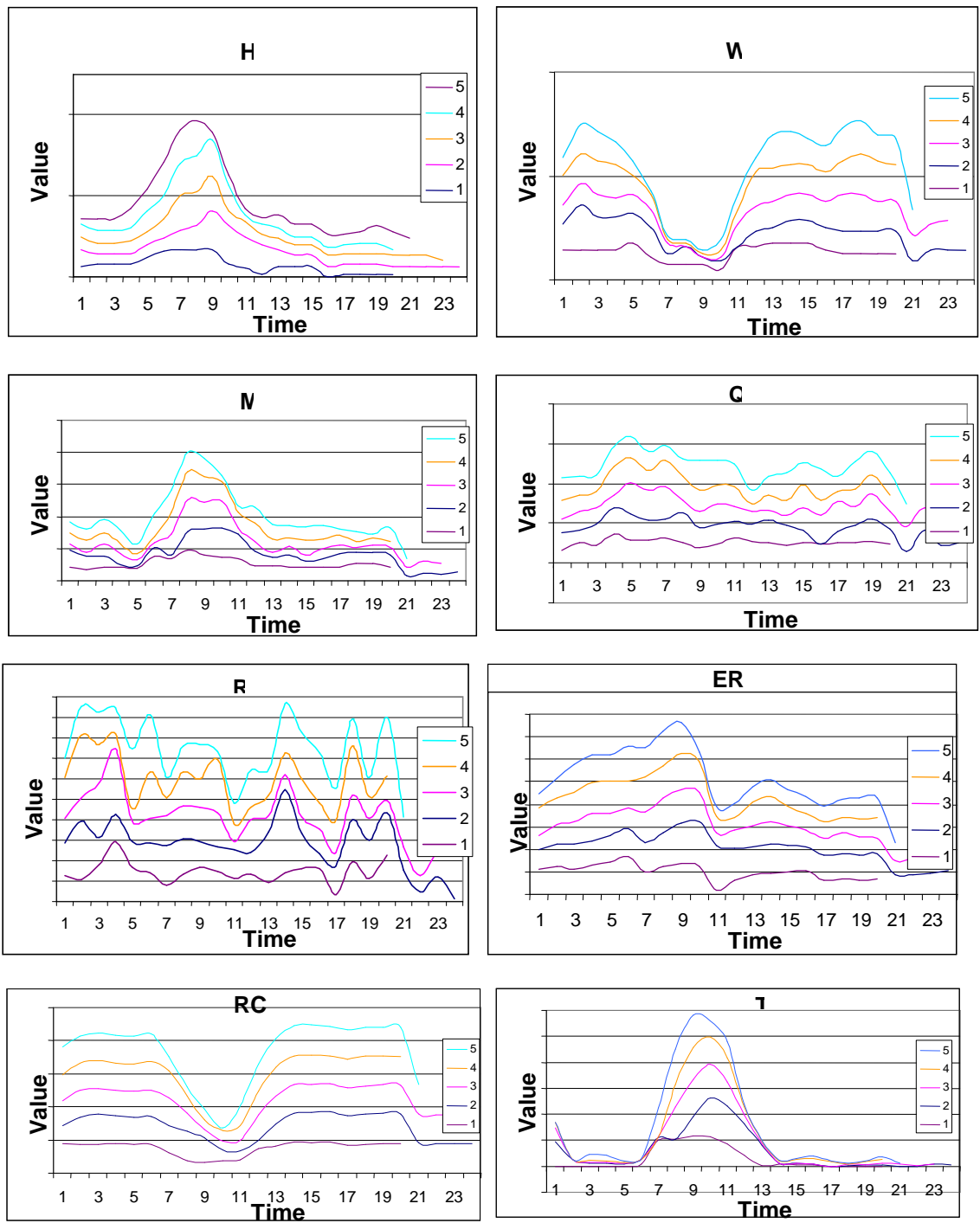

**Figure 5.9**. Stacked line charts, shows the related signatures for the word "Bomb" uttered 5 times by the same speaker.

Consequently, for each spoken word we get a feature matrix (see Table 5.1). The feature matrix has a fixed number of columns, but with a different number of rows, depending on the uttered word, and on the different speed of uttering words.

To compare signals with different lengths, we use the Dynamic Time Warping (DTW) method, and linear interpolation.

Table 5.1. Sample feature matrix for uttering the word 'Bomb'.

| Frame | H | W | M | Q | R | ER | RC | T |
|---|---|---|---|---|---|---|---|---|
| 1 | 0.26 | 0.57 | 0.44 | 0.33 | 0.25 | 0.56 | 0.89 | 0.00 |
| 2 | 0.32 | 0.57 | 0.35 | 0.50 | 0.21 | 0.63 | 0.87 | 0.00 |
| 3 | 0.32 | 0.57 | 0.40 | 0.45 | 0.35 | 0.56 | 0.88 | 0.00 |
| 4 | 0.32 | 0.57 | 0.42 | 0.74 | 0.58 | 0.64 | 0.89 | 0.00 |
| 5 | 0.47 | 0.71 | 0.38 | 0.57 | 0.33 | 0.72 | 0.86 | 0.00 |
| 6 | 0.63 | 0.43 | 0.75 | 0.56 | 0.28 | 0.84 | 0.87 | 0.04 |
| 7 | 0.68 | 0.29 | 0.67 | 0.63 | 0.16 | 0.49 | 0.61 | 0.51 |
| 8 | 0.68 | 0.29 | 0.94 | 0.49 | 0.27 | 0.63 | 0.37 | 0.52 |
| 9 | 0.68 | 0.29 | 0.79 | 0.39 | 0.33 | 0.69 | 0.32 | 0.58 |
| 10 | 0.32 | 0.21 | 0.71 | 0.49 | 0.29 | 0.61 | 0.38 | 0.56 |
| 11 | 0.21 | 0.64 | 0.69 | 0.63 | 0.23 | 0.10 | 0.41 | 0.43 |
| 12 | 0.05 | 0.64 | 0.45 | 0.55 | 0.27 | 0.33 | 0.73 | 0.22 |
| 13 | 0.26 | 0.71 | 0.46 | 0.52 | 0.19 | 0.43 | 0.83 | 0.01 |
| 14 | 0.26 | 0.71 | 0.41 | 0.43 | 0.28 | 0.47 | 0.94 | 0.05 |
| 15 | 0.26 | 0.71 | 0.41 | 0.51 | 0.32 | 0.50 | 0.93 | 0.02 |
| 16 | 0.00 | 0.57 | 0.40 | 0.47 | 0.31 | 0.53 | 0.95 | 0.02 |
| 17 | 0.05 | 0.50 | 0.43 | 0.50 | 0.06 | 0.32 | 0.83 | 0.00 |
| 18 | 0.05 | 0.50 | 0.53 | 0.49 | 0.39 | 0.34 | 0.87 | 0.00 |
| 19 | 0.05 | 0.50 | 0.53 | 0.54 | 0.23 | 0.32 | 0.90 | 0.00 |
| 20 | 0.05 | 0.50 | 0.42 | 0.49 | 0.45 | 0.35 | 0.88 | 0.00 |

For the fusion of the aforementioned features, researchers reported two types of fusion: feature-based fusion, score level fusion and decision-based fusion (Potamianos et al., 2003). Simple features concatenation is an example of feature-based fusion methods, where all the features vectors concatenated in one feature vector, which is passed as it is to the recogniser, or is transformed using appropriate transformation then passed to the recogniser. Applying such fusion in our approach is problematic for several reasons; first, the signal length will be 8 times longer than the normal length, so comparing signals becomes time inefficient, particularly when using DTW, whose time complexity is $O(n^2)$. Secondly, all the features will contribute equally to the final results, and while they are not equally representative, some features are not as representative as others.

In contrast, score level fusion methods include the use of each feature vector alone, using a weighting technique to give different weights for the features, to capture the reliability of each feature vector, depending on how informative they are. This type of fusion is adopted in this study (see Figure 5.1).

To find the proper weight for each feature, the proposed VSR system was tested on a random sample of 10 subjects chosen from the in-house database using only one feature at a time. Two different types of experiments were conducted (speaker-dependent (SD) and speaker-independent (SI)) using leave-one-word out cross-validation. The word recognition rate was different each time, depending on the strength/weakness of the tested feature. Table 5.2 shows the results of the SD experiments and the calculated weights using both SD and SI experiments.

There are many weighting schemes discussed in the literature such as equal weighting, relative weighting, variance reduction technique, root mean squared distance weighting, the average distance weighting, etc. The relative weighting scheme is chosen for this study for its simplicity and is assigned to each feature to indicate its relative importance to the recognition process. This importance is determined imperially depending on the strength of each feature measured by the word recognition rate using that feature alone. The relative weight of each feature is given by:

$$Weight_i = \frac{\overline{WRR_i}}{\sum_{j=1}^{8} \overline{WRR_j}} \quad \cdots (5.11)$$

where $\overline{WRR_i}$ is the word recognition rate for feature $i$.

**Table 5.2**. The weight of each feature depending on its recognition rate.

| Subject | H | W | M | Q | R | ER | RC | Teeth | all |
|---|---|---|---|---|---|---|---|---|---|
| Subject 1 | 21% | 11% | 13% | 4% | 5% | 20% | 21% | 39% | 51% |
| Subject 2 | 92% | 64% | 64% | 28% | 16% | 64% | 100% | 84% | 96% |
| Subject 3 | 56% | 55% | 11% | 14% | 14% | 38% | 55% | 77% | 90% |
| Subject 4 | 52% | 60% | 60% | 68% | 36% | 64% | 88% | 76% | 76% |
| Subject 5 | 72% | 44% | 40% | 56% | 44% | 76% | 56% | 76% | 76% |
| Subject 6 | 36% | 40% | 32% | 48% | 28% | 44% | 44% | 52% | 56% |
| Subject 7 | 72% | 52% | 32% | 64% | 32% | 80% | 72% | 76% | 76% |
| Subject 8 | 84% | 56% | 44% | 36% | 36% | 64% | 60% | 68% | 92% |
| Subject 9 | 22% | 23% | 9% | 11% | 9% | 37% | 43% | 62% | 68% |
| Subject 10 | 26% | 18% | 8% | 16% | 10% | 53% | 31% | 44% | 67% |
| **Average** | **53%** | **42%** | **31%** | **34%** | **23%** | **54%** | **57%** | **65%** | **75%** |
| Std | 0.26 | 0.19 | 0.21 | 0.23 | 0.14 | 0.19 | 0.24 | 0.16 | 0.15 |
| SD weights | **15%** | **12%** | **9%** | **10%** | **6%** | **15%** | **16%** | **18%** | |
| SI weights | **15%** | **14%** | **9%** | **10%** | **7%** | **9%** | **12%** | **23%** | |

As can be noted from the above table, the appearance-based features, ER, RC, and T properties are the most reliable features, particularly for the speaker-dependent approach. This result was backed up by Jun and Hua (2009). The image transform-based features M, Q and R are the weakest properties. This weakness may come from the used transform, which is the DWT. Although DWT provides a good tool for signal and image processing, it has some drawbacks: 1) shift sensitivity, 2) poor directionality and 3) lack of phase information (Fernandes et al., 2003). Transform-based features nevertheless contribute well to the final result when they are fused in the score level. The reliability of the geometric features H and W is in the middle. They are not affected either by the drawbacks of the transformation method, or by the *bias* of the ROI appearance

Notice the last column in Table 5.2, which shows the result using all the features without weighting. These results assume that each feature contributes the same to the recognition algorithm while the results in the table show the opposite. We expect to get better results when using such weights as described in the next chapter.

Despite their contributing differently to the results, all the extracted features are important to the recognition rate. Some features tend to contribute better for speaker-dependent VWR while others contribute better for the speaker-independent scenario. Appearance-based features (ER, RC and T) are expected to contribute less to the results of speaker-independent, because mouth appearance is different from one speaker to another. This justifies the reduced weights in the SI weighting approach for both ER and RC, but does not explain the increased weight for T (teeth)! Because this feature is strongly associated with the utterance of specific phonemes such as [s] and [T], this argument opens the door for future investigation of these features, to enhance them and find more new descriptive features for both SD and SI experiments.

For each signal, the distances are measured with other signals from the training data, using DTW or Euclidian distance after linear interpolation, to overcome the different signal lengths. The fusion in the score level is achieved by using the weighted average of the distances of the eight signals; the weighted average is given by:

$$WA = \frac{\sum_{i=1}^{8} w_i D_i}{8} \quad \cdots (5.12)$$

where $w_i$, $D_i$ are the weight and the distance of the $i^{th}$ feature respectively.

According to the K-Nearest-neighbour (KNN), the minimum k weighted averages are considered to predict the class (word) by announcing the maximum occurrence class in the nearest k as the predicted class.

### 5.1.2.1 Dynamic Time Warping

Since humans speak in different ways and at different speeds, this not only yields visual words with different lengths (even for the same word), but also yields signals with shifted phases (see Figure 5.10: left). A direct comparison one-one method, such as the Euclidean distance, will not work well in such circumstances, and yields unexpected results (see Figure 5.10: right).

Unlike normal Euclidean distance, DTW "warps" the signals non-linearly in the time dimension, i.e. it determines a measure of their similarity autonomous from certain non-linear variations in the time dimension (see Figure 5.10: left).

Dynamic Time Warping (DTW) is a well-known quadratic time algorithm designed to measure similarity between two numerical sequences, which may vary in length. It is used here to measure similarity between visual words signals, which normally vary in length, because people vary in their speech speed.

DTW was first introduced by Vintsyuk (1968), and used for speech recognition for the first time by Sakoe and Chiba (1978).

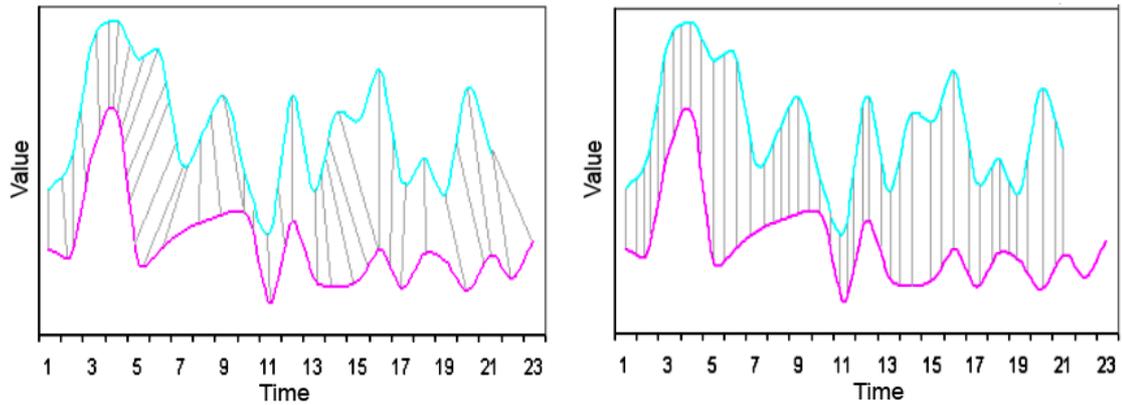

**Figure 5.10**. Two different signals for the same word "Bomb", (left) warping the time to alleviate phase change, (right), no time warping, direct comparison.

DTW has several different algorithms. The following algorithm was used in this study:

```
Given two time series X and Y
// x_i is a point in a test feature vector
X = x_0, x_1, …, x_{N-1}
// y_i is a point in a training feature vector
Y = Y_0, Y_1, …, Y_{T-1}

// initialize cost matrix
For i=0 to N-1
   For j=0 to T-1
      Cost(i,j) = √((x_i − y_j)²)
// find the optimal path which minimizes the cost
DTW[0,0] = 0
For i=1 to T-1
   DTW[0,i] = ∞
For i=1 to N-1
   DTW[i,0] = ∞
For i=1 to N-1
   For j=1 to T-1
      DTW[i,j]=Cost(i,j)+ Minimum{DTW[i,j-1],
                                  DTW[i-1,j],
                                  DTW[i-1,j-1]}
Return (DTW[N-1,T-1])
```

### 5.1.2.2 The K-nearest Neighbour (KNN)

KNN is a simple machine learning algorithm used for classifying objects based on the closest (nearest) training examples in the feature space. An object is classified by a majority vote of its neighbours, with the object being assigned to the class most common amongst its k nearest neighbours.

Let P={$X_1$, $X_2$, $X_3$,..,$X_n$} a set of n labelled examples, and let X' Є P be the example nearest to a test point X, then the nearest-neighbour (NN) rule for classifying X is to assign it the label of X'. (Duda et al., 2001).

The k-nearest-neighbour rule is an extension of the NN rule, and it classifies X by assigning it the label of the most frequent example in the k-nearest examples set (Duda et al., 2001) (see Figure 5.11).

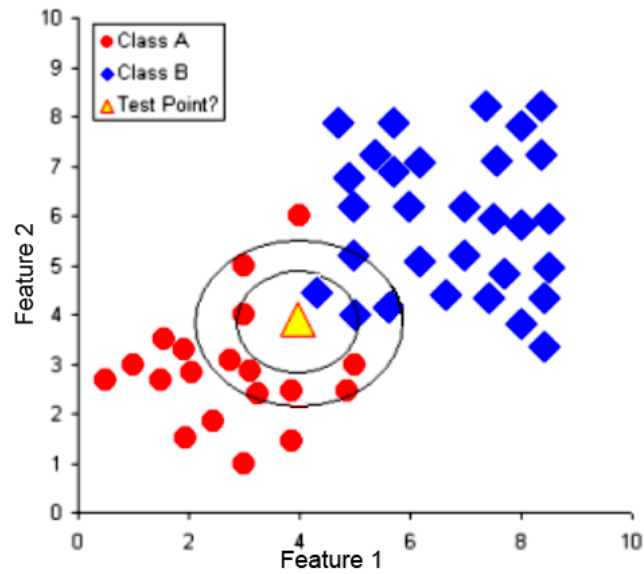

**Figure 5.11**. The effect of the number of neighbours (k) on KNN decision.

For example, considering Figure 5.11, assume that there are two classes, A and B, and a test point to be classified. If k = 3 (the points in the small circle in Figure 5.1), then the test point (the yellow triangle) will be classified as class B; because 2 points (blue) belong to class B, and only 1 point (red) belong to class B. Whereas, if k = 12 (the points in the big circle), then the test point will be classified as class A; because the nearest 12 points are divided into 8 reds and 4 blues, and the red points belong to class A.

The time complexity of the naïve version of the KNN algorithm is O(dn), while d is the dimension of an example, and n is the total number of examples. This complexity comes from measuring the distance of the test example with all the examples in the training set.

However, this complexity can be reduced to O(1) using one of three general approaches: 1) computing partial distances, 2) pre-structuring, and 3) editing the training examples. Partial distances methods depend on measuring the distance of part of the test example with part of the training example. If the result was greater than a specific threshold it stops, and considers another example, assuming that this example will be far away from the test example (Duda et al., 2001).

In pre-structuring methods, a new structure like the search tree is built, and all the training examples are linked together according to their similarity. A test example is compared to the first level, then carries on the comparison with only the linked examples to the nearest example. In the editing methods, some training examples are meant to be eliminated to reduce the number of examples in the training set, which leads to complexity reduction. The method needs to search for useless examples, which can lead to errors in the classification method and remove them before the test process starts. A simple method is to remove the examples that are surrounded by examples of the same class, leaving the decision boundaries (Duda et al., 2001). The problem with KNN is that it is sensitive to outliers. To solve this problem we propose to weight each neighbour by its distance from the test example, thus assigning importance not only to the nearest neighbours but also to their distances.

### 5.1.2.3 The Weighted K-nearest Neighbour

This study proposes a weighting technique for the KNN algorithm. This method is the same as the normal KNN, but rather it considers a weight for each neighbour depending on its distance from the test point. We need to keep the effect of the maximum number of the same class, at the same time maintaining the effect of the distance weight. The weights are calculated using the following equation:

$$W_i = \frac{D_i}{m_i} \quad \cdots (5.13)$$

where $i=1$ to $n$, where $n$ is the number of different classes, $D_i$ is the distance of the nearest example of class $i$, $m_i$ is the repeated number of class $i$ within the nearest k examples, and $W_i$ is the weight of class $i$.

This equation is calculated only for the nearest k examples, and not for all the examples. The test point is classified by:

$$Predected\ Class = \arg\min(w_i) \quad \cdots (5.14)$$

Accordingly, the minimum the distance to the test point, the minimum $W_i$ will be and the more repeated time of a class in the first minimum k examples, the minimum $W_i$ will be. This method is evaluated and compared with the normal KNN.

## 5.2 Time complexity

For any computational algorithm it is important to discuss its time and space complexity, particularly if that algorithm/method is intended to be applied using limited resources machine such as the PDAs, and mobile phones, where processors' speeds are not too fast, and memories are not too big compared to normal computers. Algorithm complexity can be measured using the Big O notation, for example an algorithm of O(n) has a linear complexity, where n is the maximum number (in the worst case) of steps carried out by the algorithm to finish its task.

The proposed Visual words system (Vwords) complexity (time and space) involves the following:

- Time complexity for both H and W features is O(1).
- Time complexity for M, and Q is $O(4n^2)$ because there are 4 wavelet sub-bands.
- Time complexity for R, T, and RC is $O(n^2)$ because the size of ROI is W x H.
- Time complexity for ER is $O(18n^2)$ because ER uses filter of 3 x 3 for the horizontal features and the same for vertical features.
- Time complexity for KNN is $O(8n)$ because we have 8 features before the fusion process.
- Time complexity of the face and lip localization is $O(n2)$.

This means that the overall time complexity is:

$O(1)+ O(1).+ O(4n^2)+ O(4n^2)+ O(n^2)+ O(n^2)+ O(n^2)+ O(18n^2)+ O(8n)+ O(n^2)+ O(n^2)$

which can be approximated to $O(n^2)$. Since $n^2$ came from the ROI size which is W x H, however, this size is relatively small, compared to the face, for instance.

## 5.3 Summary

This study proposes a new VSR approach dependent on the signature of the word itself. This is done by proposing a hybrid feature extraction method, depending on geometric, appearance, and image transform features. We found that there is no need to use all or some of the lip's contour points to define the shape of the lips, where the height and width of the mouth backed up with a bounding ellipse is enough to approximate the real outer contour of the lips. The new proposed VSR approach is termed the "visual words" (see Figure 1).

The appearance-based features ER, RC and T properties are the most reliable features, particularly for the speaker-dependent approach, while the image transform-based features M, Q and R are the weakest properties, and the reliability of the geometric features H and W is in the middle. All the proposed features nevertheless contribute well to the result when they are fused in the score level.

Weighted average of the distances using all features is used as a score level fusion, which is passed to the KNN classifier. Different signal lengths are dealt with using either DTW, and/or linear interpolation. A weighted KNN classifier is proposed to enhance the word recognition rate. The proposed system was programmed, applied, and evaluated using the new database mentioned in chapter 4. For the results, test protocol and evaluation process, see chapter 6.

# Chapter 6

# Evaluation of the Proposed VW System

This chapter is concerned with performance testing of the VW scheme proposed in the previous chapter. We will present and analyse the result of several sets of experiments that have been conducted for this purpose and to compare its performance to that of other approaches (visemes). We shall also attempt to shed some light on several areas where the visual words can be applied, including speaker identification and verification (the visual passwords), and security surveillance.

The evaluation is done in two stages: the initial stage and the substantive stage. The aim of the initial stage, also termed the pilot study, was to determine whether a visual word should be represented by the ROI feature vectors for the sequence of frames as a time series (i.e. a matrix of nfx8 where nf is the number of frames) or use its conventional representation as a sequence of visemes. The pilot study includes several experiments, and uses a selection of videos from the PDADatabase (see section 6.2). The pilot study will demonstrate that the matrix representation approach provides promising results for the VSR problem. The substantive stage will demonstrate that the adopted approach outperforms the results of the visemes approach and compares well with the human perception of lip reading.

Speakers in the PDADatabase utter a sequence of words or numerals in a continuous manner without a gap between words, and it was very difficult to read some lips in some cases. Therefore a more credible and fair assessment of the performance of VW schemes required a new database that was created for the specific purpose of this study, which would form the platform for the more substantive experiments.

The rest of this chapter is organised so that we first describe the test protocols to be followed throughout. In section 6.2 we present and analyse the VW pilot study experiments. The subsequent 3 sections 6.3-6.5 are devoted to the analysis of substantive experiments. In section 6.6 a comparison of performance with existing schemes is discussed and in section 6.7 we test the viability of using the Hidden Markov Model (HMM) approach for visual speech recognition. We close the chapter by discussing a possible approach to cluster English language words using a visual signature similarity measure.

## 6.1 Test protocol experimental settings

In the last chapter, we proposed the VW recognition system whereby a visual word sample is represented by the time series of the 8-dimensional vectors extracted from the mouth ROIs in the frames of the video where the speaker utters the given word. The input to every experiment referred to in the rest of this thesis consists of extracted features corresponding to every video and word involved in the experiment.

At this point we need to recognise an inherent difficulty in the automatic visual word recognition problem. Different persons speak at different speeds and therefore produce visual word features of different lengths. It is also known that the same person might utter the same word at a different speed at different times. Within reasonable limits, humans can understand each other regardless of the speed with which the words are spoken. However, this presents a major challenge to the automatic visual word system, since machines are then required to compare two visual signals of different lengths even though the two signals represent the same spoken word.

During the pilot stage and the subsequent investigations, several experiments were designed to achieve the main objectives of the study, which are to gain better understanding of the automatic visual speech recognition problem, and to propose solutions (computer system) to improve the performance of the recognition process as well as evaluating the proposed visual speech system.

The following main types of experiments were designed and carried out:

1. **Speaker-dependent experiment**: this was conducted on each subject alone, all the test examples and the training examples pertaining to the same subject (person). The main goal of this experiment is to test the way of speaking unique to each person, and each one's ability to produce a visual signal that was easily read. These experiments use **leave-one-word-out cross-validation** protocol; each word sample for each subject is considered as a test example, and the other examples of the same word, and the other examples of the other 29 words, are considered as a training set. After testing each example, another example is removed from the training set and joins the test set, and the current tested example joins the training set, and so on, until no more examples are tested for each subject, and for all subjects in the database.

2. **Speaker-independent experiment**: in this type of experiment the computer evaluates a group of persons, and each time one person gets out of the training set and is tested against the remaining persons in the group, the leave-one-subject-out cross-validation comes into play to evaluate the methods, neglecting the personal differences in the way of speaking. The training set does not contain any examples belonging to the tested subject, i.e. the training set contains all the subjects from the database, except the tested subject. Each time, after testing each subject, the current test subject joins the training set, and another subject is removed from the training set and assigned as a new test subject (test set), and so on, until no more subjects need to be tested. The average is then taken to verify the accuracy of the experiment.

The aforementioned experiments were tested and evaluated using the Dynamic Time Warping (DTW) algorithm as a distance function (see chapter 5) to overcome the problem of different lengths. The normal Euclidean distance is also used after performing linear interpolation to obtain signals of the same lengths.

The nearest k neighbours (KNN) and the weighted-KNN (WKNN) were used as majority decision rules. The results of the proposed system were compared with other systems, and also compared with the visemes approach using the Hidden Markov Model (HMM).

The leave-one-out cross-validation technique is used to evaluate the system despite the higher load, processing time and memory size employed by this method as compared to the multi-fold cross-validation, for instance. This is motivated by the fact that if the proposed system is applied online in real life, the words will input the system one by one, and can be tested against a pre-defined and trained database in the same manner as these experiments were designed.

The beginning and the end of each visual word were determined for the purposes of the experiments using the audio signal, which is analysed using an open source library called WaveSurfer[*] – an Open Source tool for sound visualization and manipulation designed by Sjölander and Beskow (2000).

As the human voice lies within a specific frequency range, WaveSurfer uses a specific threshold to detect the human voice and distinguish it from noise so it was easy to determine the limits of all the visual words. The automatic detection of the words' limits was edited manually to make sure that the boundaries (the beginning and the end frames) were chosen correctly. This was achieved by allowing 3 frames (approximately) at the beginning before the audio signal started, because the visual signal appears before the audio signal, as the face muscles start to move to produce the audio signal. See Figure 6.1.

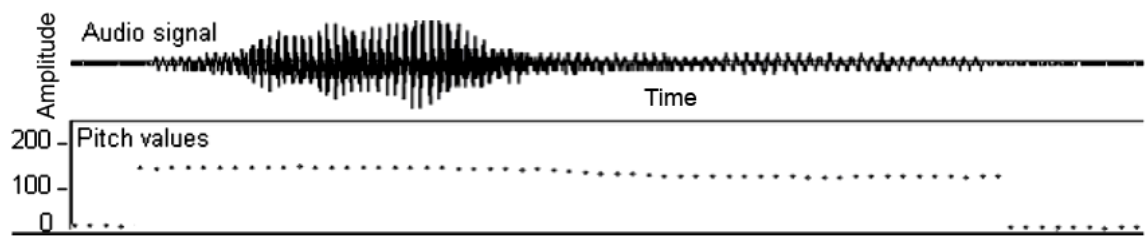

**Figure 6.1**. Upper part, the audio signal for the word "one", and the lower part, the pitch values of the audio signal, the dots high values represent the human voice, and the low values represent silent and non-human voices.

After examining the database, it was found that some people did not produce a complete visual signal while talking, i.e. they rarely moved the visual parts of their speech production systems (lips and mouth region). Consequently, they produced almost the same signal for different words, which makes it almost impossible to visually recognize what they have said, either by machine or human intelligence. As an extra aim of the SD experiments, and by performing the experiments on all the participants in the database, it becomes an easy task to identify those (visual-speech-impaired-people, or visual-*speechless-persons*) (VSP) – a good indication to identifying them is that they will be the participants whose speaker-dependent-results appear to be very low and less than the others' results.

However, this study is concerned with normal people, those who are willing to talk properly, by producing at least the visual signals clearly. Nonetheless, they did not have to pay extra attention to producing a better visual signal by opening their mouths to a greater extent than normal.

---

[*] WaveSurfer can be downloaded from: http://www.speech.kth.se/wavesurfer

## 6.2 The pilot study experiments

Two experiments were conducted as part of the pilot study to evaluate the visual words approach. The first was a speaker-independent (SI), and the second was a speaker-dependent (SD) experiment.

In the speaker-independent experiments we randomly selected 48 PDADatabase videos (23 different females and 25 different males) in each of which the person utters the same sequence of words {6,1,2,9,7}, but without gaps between the individual words.

Leave-one-speaker-out cross validation is used to evaluate the method. Each time, all the words spoken by each subject became the test set, and the rest of the words spoken by the rest of the subjects formed the training set. Using the same methods mentioned in chapter 5, the KNN with either Euclidean distance (ED) or Dynamic Time Warping (DTW) was used to recognize the words spoken in the test set.

**Table 6.1**. Speaker-independent experiment: primary results of the VW.

|        | K=1 ED | K=1 DTW | K=4 ED | K=4 DTW | K=8 ED | K=8 DTW |
|--------|--------|---------|--------|---------|--------|---------|
| **Male**   | 32.80% | 33.60%  | 36%    | 39.20%  | 36%    | **39.20%** |
| **Female** | 38.26% | 59.13%  | 31.30% | 54.78%  | 35.65% | **60%**    |
| **Both**   | 36.25% | 45.83%  | 36.67% | 50.83%  | 38.75% | **56.25%** |

As can be observed from Table 6.1, the best word recognition rate (56.25%) is when the DTW is used as the measurement distance. This is explained by the power of this algorithm when it deals with signals of different lengths, unlike the ED, which compares only the digital signal opposite points and cannot compare the rest of the longer signal.

Also, choosing k=8 in the k-NN algorithm gives better results than 4 or 1. This is normal in SI experiments, because more neighbours (bigger k), allows more chances to have optimal solutions (that come from different users) among the k nearest neighbours.

The second experiment was speaker-dependent. This tested 12 videos for two persons, a male and a female (6 videos for each), each one saying different words from the domain: {1,2,3,4,5,6,7,8,9}. This is similar to the previous experiment, but this time one word (example) is left out, as a test, and the rest of the examples from both speakers are trained to find the nearest K neighbours, then a majority decision rule is used to determine the class (the word). This was repeated for all the examples, and the average was taken at the end. The results of this experiment are depicted in Table 6.2.

**Table 6.2**. Speaker-dependent experiment: primary results of the Vwords.

|  | K=1 | | K=4 | | K=8 | |
|---|---|---|---|---|---|---|
|  | ED | DTW | ED | DTW | ED | DTW |
| **Male** | **72.00%** | 68% | 56.00% | 28% | 28.00% | 16% |
| **Female** | 45.71% | **48.57%** | 31.43% | 31.43% | 25.71% | 31.43% |
| **Both** | 43.33% | **56.67%** | 45.00% | 53.33% | 43.33% | 46.67% |

This experiment produced unexpected results; according to the first experiment, the female results (word recognition rate 48.57%) are expected to be better than the male (72%). Also, the smaller the value of k, the better the result. Unlike the previous experiment, this kind of experiment relies on the nearest single neighbour, rather than a group of neighbours, because taking more than one neighbour increases the contribution of the potential misclassified words in the majority decision rule.

By investigating the male and the female videos, which were used in this experiment, the male participants were seen to be speaking clearly and revealing much visual information while talking. In contrast, the female participants were not providing much visual information while talking. Thus the male participant in this experiment was more easily read using the proposed method, as opposed to the female. This explains why the male results were better than the female. However, it was still very early to draw conclusions using such primary experiments at this stage.

A large number of experiments were then conducted to further investigate and evaluate the proposed method, but this time using the in-house database (see chapter 4). The PDADatabase is not oriented to lip-reading, as the designers of the PDADatabase did not consider lip reading when they created the database. For example, there is not much vocabulary, and the words are pronounced without leaving a space between words (continuous speaking), which causes several words to be merged into one which most likely results in missing some visual feature values at the beginning and/or the end of the spoken words. Moreover, the PDADatabase was recorded with a PDA, which had a web-quality camera, and the video capturing is not stable due to hand vibration. These were the main incentives for recording the new (in-house) database using an HD camera mounted on a stable stand and speakers, which would utter each word separately.

## 6.3 The new database (in-house) – revisited

The new database was designed for the specific purposes of this study.

| Table 6.3. Types of words used in the database, indexed by number under each word. | | | | | | |
|---|---|---|---|---|---|---|
| Category | No# words | No# repetition | No# sessions | No# subjects | Total | The words |
| *Numeric* (Nu) | 10 | 5 | 2 | 26 | 2600 | (Zero, one, two, … nine) ( 0, 1, 2, …. 9 ) |
| *look-alike 1* (LAL1) | 5 | 5 | 2 | 26 | 1300 | (Fold, Sold, Hold, Bold, Cold) ( 15, 16, 17, 18, 19 ) |
| *look-alike2* (LAL2) | 5 | 5 | 2 | 26 | 1300 | (Knife, Light, Kite, Night, Fight) ( 10, 11, 12, 13, 14 ) |
| *Long General* (LG) | 5 | 5 | 2 | 26 | 1300 | (Appreciate, University, Determine) ( 20, 21, 22 ) (Situation, Practical) ( 23, 24 ) |
| *Security* (Sec) | 5 | 5 | 2 | 26 | 1300 | (Bomb, Kill, Run, Gun, Fire) ( 25, 26, 27, 28, 29 ) |
| Total | 30 | | | | 7800 | |

The words spoken were categorized into 5 different classes: the numeric (*Nu*), look-alike 1 (*LAL1*), look-alike 2 (*LAL2*), long general words (*LG*), and some words that relate to security issues (*Sec*) (see Table 6.3). Each speaker (who uttered 30 different words) was recorded twice: session 1 and session 2. Each word was repeated 5 times for each recording; with 26 participants, the total number of examples is 7800 (see Table 6.3).

The appearance of facial hair affects the extracted features, particularly the appearance-based features, and this in turn affects the accuracy of detecting the lips, as well as the final results of the lip reading. The new database contains some subjects with facial hair. Table 6.4 shows some information about the database subjects, including their ethnic group, whether they have facial hair, and being a native or non-native speaker.

As can be seen in Table 6.4, the 26 subjects are divided into 4 ethnic groups: Africans, Asians, European, and Middle Eastern. Also 9 males have facial hair. This kind of information is useful for results interpretation.

**Table 6.4.** Subject Grouping: ethnicity, facial hair status, and native or non-native speakers.

| Subject | Ethnic group | Native speaker | Has beard | Has moustache |
|---|---|---|---|---|
| Female_01 | African | | N/A | N/A |
| Female_02 | Asian | | N/A | N/A |
| Female_03 | Middle Eastern | | N/A | N/A |
| Female_04 | European | ✓ | N/A | N/A |
| Female_05 | Middle Eastern | | N/A | N/A |
| Female_06 | European | | N/A | N/A |
| Female_07 | African | | N/A | N/A |
| Female_08 | European | ✓ | N/A | N/A |
| Female_09 | African | | N/A | N/A |
| Female_10 | Middle Eastern | | N/A | N/A |
| Male_01 | Middle Eastern | | ✓ | ✓ |
| Male_02 | Middle Eastern | | ✓ | ✓ |
| Male_03 | Middle Eastern | | ✓ | ✓ |
| Male_04 | European | | | |
| Male_05 | European | ✓ | ✓ | ✓ |
| Male_06 | African | | | ✓ |
| Male_07 | Middle Eastern | | | ✓ |
| Male_08 | Middle Eastern | | | |
| Male_09 | Middle Eastern | ✓ | ✓ | ✓ |
| Male_10 | European | | | |
| Male_11 | Middle Eastern | | | |
| Male_13 | European | ✓ | | |
| Male_14 | Asian | | | |
| Male_15 | Middle Eastern | | | ✓ |
| Male_16 | European | ✓ | | |
| Male_17 | African | | ✓ | ✓ |

## 6.4 Speaker-dependent experiments

The SD experiments were conducted under the same test protocol that was discussed in section 6.1. For each subject in the database, different recognition algorithms and different distance measures were used (see Table 6.5 for the results).

The results of the SD experiments show that lip-reading results depend very much on the individual speaker rather than the reader, and whether the reader was human or machine. Hence the results were in the range of (35% to 91%).

It can be noticed from Table 6.5 that the weighted k nearest neighbour (WKNN) outperforms the classical KNN, but not by too much; as the difference is only 1, this might not be worth the processing time calculating the weights for the k neighbours, whereas for other applications, and/or other databases, WKNN might make a significant difference.

**Table 6.5**. Speaker-dependent experiments results.

| Subject | KNN | | | | WKNN | |
|---|---|---|---|---|---|---|
| | No interpolation | | Interpolation | | Interpolation | |
| | ED | DTW | ED | DTW | ED | DTW |
| Female_01 | 15% | 57% | 62% | 63% | 62% | **67%** |
| Female_02 | 57% | **91%** | 90% | **91%** | 90% | **91%** |
| Female_03 | 39% | 77% | 82% | 80% | 82% | **83%** |
| Female_04 | 21% | 65% | **67%** | 63% | **67%** | 63% |
| Female_05 | 35% | 67% | **77%** | 71% | **77%** | 73% |
| Female_06 | 43% | 62% | 66% | 65% | 66% | **68%** |
| Female_07 | 22% | 70% | 77% | 73% | **81%** | 75% |
| Female_08 | 47% | 77% | 80% | 79% | 80% | **81%** |
| Female_09 | 30% | 62% | **71%** | 65% | **71%** | 67% |
| Female_10 | 21% | 74% | 80% | 79% | 80% | **81%** |
| Male_01 | 17% | 59% | 71% | 66% | **74%** | 69% |
| Male_02 | 16% | 38% | **47%** | 38% | **47%** | 41% |
| Male_03 | 8% | 29% | 37% | 32% | **39%** | 35% |
| Male_04 | 18% | 39% | **52%** | 45% | **52%** | 46% |
| Male_05 | 12% | 67% | 71% | 72% | **73%** | 72% |
| Male_06 | 11% | 47% | 52% | 51% | 56% | **57%** |
| Male_07 | 22% | 72% | **80%** | 76% | **80%** | 79% |
| Male_08 | 27% | 57% | **63%** | 59% | **63%** | 59% |
| Male_09 | 23% | 79% | **83%** | 82% | **83%** | 82% |
| Male_10 | 53% | 69% | **78%** | 71% | **78%** | 72% |
| Male_11 | 32% | 75% | **78%** | 77% | **78%** | 77% |
| Male_13 | 33% | 55% | **61%** | 59% | **61%** | 59% |
| Male_14 | 15% | 39% | 45% | **48%** | 45% | **48%** |
| Male_15 | 19% | 47% | 55% | 53% | **57%** | 53% |
| Male_16 | 41% | 77% | 72% | **80%** | 72% | **80%** |
| Male_17 | 21% | 43% | **51%** | 49% | **51%** | 49% |
| Average | 27% | 61% | 67% | 65% | **68%** | 66% |

This table also shows the effect of using linear interpolation, especially when using the ED as a distance measure (67% word recognition rate with interpolation, compared to 27% with no-interpolation). This is obvious, because of the different lengths of the words spoken, interpolation stretches the smallest signal to be the same as the longer one, which makes the comparison more realistic. While this is not the case for the DTW, interpolation has no big effect due to the nature of the DTW algorithm, which is not affected much by the different lengths of the signals (see chapter 5).

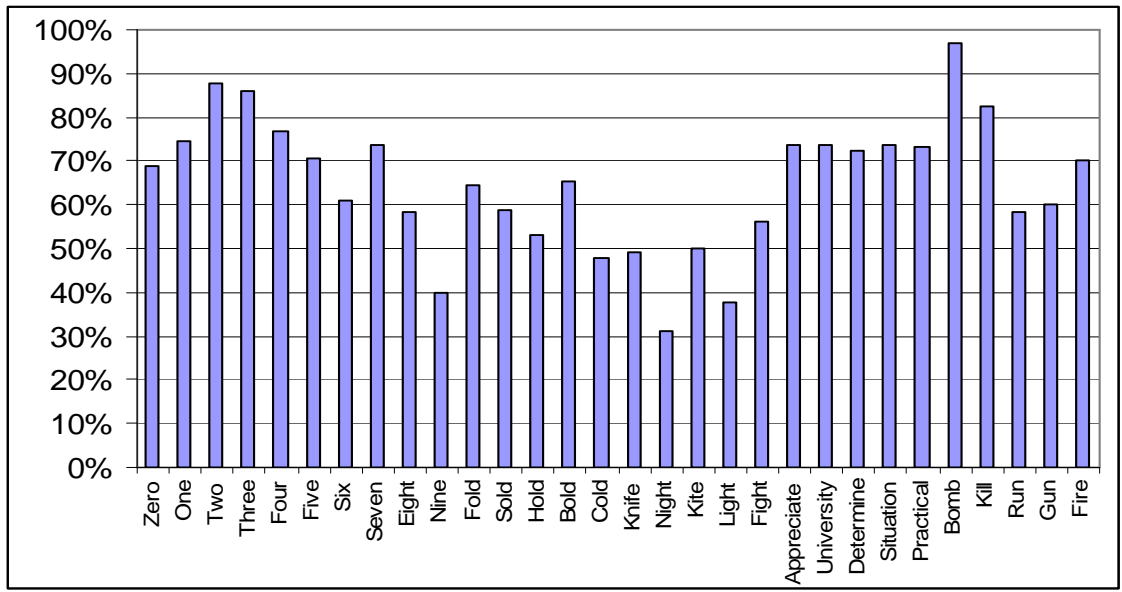

**Figure 6.2**. Speaker-dependent recognition rate for each word.

This figure illustrates that the word "Bomb" is the best recognized word (97%), perhaps because most of the participants started and ended the word with the phoneme [b], which is a labial phoneme (see chapter 2, section 2). A labial phoneme is a consonant articulated using both lips, and nothing is clearer and more obvious than lips in the human speech production system. Moreover, [b] is a plosive phoneme, which is a consonant articulated by stopping the airflow in the vocal tract, both lips are closed while the air flow is topping, then they open, showing all the dynamic movements needed for recognition.

Thus, the word recognition rate is greatly affected by its phoneme components, because producing these sounds involves the movement (dynamic) of some parts of the speech production system. It will be seen that the nearer the articulated sound to the lips, the more visual the dynamic will be, or the more obvious the sound is. For example, the recognition rate of the word "Nine" (40%) is one of the worst, because it starts and ends with the phoneme [n], which is an alveolar consonant (articulated with the tongue against the roof of the mouth), and it is also a nasal sound which allows the air to flow through the nose, so some people can produce this sound without even opening their mouths. If all the sounds of the words were uttered by the lips, they would have approximately the same recognition rate.

By looking at the middle of Figure 6.2, the look-alike words (LAL1 and LAL2) seem to be less recognized than the other words because their signals are more likely to be similar. For example, the signature of the word "Night" is almost similar to the word

"Light", because the phonemes [n] and [l] are both alveolar consonants, and the remaining parts of the words are the same, i.e. "Night" might be recognized as "Light" and vice versa. To test whether this applies to other words in both groups (LAL1 and LAL2), the speaker-dependent experiment was repeated again, but this time if any word from LAL1 (or LAL2) was recognized as any other word in the LAL1 (or LAL2) group, then a true acceptance was declared. The results are presented in Table 6.6, below, which confirm our observation. It can be noticed that the word recognition rate is in the range of (44%-97%), and the average word recognition is increased by about 10% (compared to Table 6.5), as a result of considering the look-alike words in each category as one word.

**Table 6.6**. Speaker-dependent experiment using WKNN, ED and linear interpolation, and considering the look-alike words to be treated as one word in each group.

| Subject | Max | K=1 | K=2 | K=3 | K=4 | K=5 | K=6 | K=7 | K=8 | K=9 |
|---|---|---|---|---|---|---|---|---|---|---|
| Female_01 | 69% | **69%** | **69%** | 67% | 68% | 65% | 64% | 65% | 66% | 65% |
| Female_02 | 97% | **97%** | **97%** | 95% | 95% | 93% | 92% | 91% | 89% | 88% |
| Female_03 | 87% | **87%** | **87%** | 83% | 85% | 85% | 81% | 77% | 78% | 75% |
| Female_04 | 81% | **81%** | **81%** | 79% | 79% | 80% | 78% | 76% | 73% | 73% |
| Female_05 | 83% | **83%** | **83%** | 82% | 80% | 79% | 82% | 78% | 78% | 77% |
| Female_06 | 75% | **75%** | **75%** | 72% | 69% | 68% | 64% | 66% | 66% | 62% |
| Female_07 | 82% | **82%** | **82%** | **82%** | **82%** | **82%** | 79% | 79% | 78% | 75% |
| Female_08 | 85% | **85%** | **85%** | 82% | 81% | 77% | 75% | 75% | 73% | 73% |
| Female_09 | 81% | **81%** | **81%** | 75% | 77% | 74% | 73% | 75% | 75% | 71% |
| Female_10 | 88% | **88%** | **88%** | 86% | 85% | 83% | 81% | 81% | 81% | 78% |
| Male_01 | 79% | 77% | 77% | **79%** | **79%** | **79%** | 77% | 75% | 76% | 70% |
| Male_02 | 61% | **61%** | **61%** | 57% | 55% | 51% | 49% | 49% | 49% | 51% |
| Male_03 | 44% | **44%** | **44%** | 43% | 42% | 43% | 40% | 37% | 37% | 35% |
| Male_04 | 63% | **63%** | **63%** | **63%** | 62% | 56% | 60% | 56% | 55% | 55% |
| Male_05 | 85% | 84% | 84% | **85%** | **85%** | 82% | 83% | 80% | 75% | 72% |
| Male_06 | 65% | 63% | 63% | 59% | **65%** | 63% | **65%** | 62% | 64% | 58% |
| Male_07 | 84% | **84%** | **84%** | 81% | 79% | 75% | 75% | 75% | 70% | 69% |
| Male_08 | 75% | **75%** | **75%** | 71% | 71% | 67% | 67% | 69% | 66% | 63% |
| Male_09 | 92% | **92%** | **92%** | 91% | 89% | 87% | 85% | 83% | 81% | 80% |
| Male_10 | 88% | **88%** | **88%** | 83% | 83% | 83% | 79% | 77% | 75% | 69% |
| Male_11 | 84% | 83% | 83% | 83% | 83% | **84%** | 83% | 81% | 79% | 79% |
| Male_13 | 69% | **69%** | **69%** | 67% | 66% | 63% | 60% | 58% | 57% | 56% |
| Male_14 | 53% | **53%** | **53%** | 50% | 49% | 48% | 43% | 41% | 43% | 45% |
| Male_15 | 69% | 67% | 67% | **69%** | 65% | 61% | 61% | 59% | 60% | 59% |
| Male_16 | 83% | **83%** | **83%** | 80% | 81% | 75% | 75% | 72% | 70% | 67% |
| Male_17 | 62% | **62%** | **62%** | 60% | 59% | 59% | 56% | 54% | 54% | 53% |
| Average | 76.38% | 76% | 76% | 74% | 74% | 72% | 70% | 69% | 68% | 66% |

| | |
|---|---|
| **83%** | Females only |
| **72%** | Males only |
| **77%** | Excluding subjects with moustache & beard |
| **71%** | Subjects with moustache & beard only |
| **78%** | Excluding VSP |

Furthermore, the results become higher when the number of neighbours (k) is at the lower end of the scale, particularly when 1≤ k ≤4. This can be explained by the fact that there are only 4 examples in the training set for each subject, and one example as a test example, so that taking more than 4 neighbours increases the probability of falsely accepting other words. Thus, for speaker-dependent experiments, the smaller the value of k is, and the less unrelated words occur in the nearest decision rule, which leads to a higher word recognition rate.

The speaker-dependent experiments also show that females provide better visual signals than males while talking, and the average word recognition rate for female speakers is (83%), which is significantly higher than it is for male speakers (72%). This could be due to several factors, the main one being the clear appearance of the female lip area (the lack of facial hair on female faces) and the use of makeup. This allows for more lip-detection accuracy, and more detail appears on the facial features. This contrasts with the appearance of facial hair in males, where some details are hidden by this obstacle. For example, the moustache of Male-3 totally hides his upper lip (see Figure 6.3) and that is probably why the word recognition rate of Male-3 is the worst (44%).

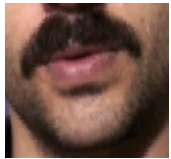

**Figure 6.3**. Facial hair hiding mouth details (Male-3 from the new database).

Generally, the word recognition rate of the subjects with moustache and beard only is 71%, while it increases to 77% when excluding the results of these subjects. These results show the effect of facial hair on the lip-reading accuracy.

Values in columns 3 and 4 in Table 6.6 are exactly the same; when K=1 or K=2 they are the same for WKNN, because if there are 2 neighbours, the algorithm will consider the closest of them, which is the same value in the k=1 case, e.g. if the nearest 2 distances were 2.1 and 2.5 and if k=2, the algorithm will consider the class of the distance 2.1, no matter what the class of the distance 2.5 and the same class will be considered if k=1, so the results will be the same.

We define and use the term **visual speechless person** (VSP) for persons who do not provide visual signals while talking, or at least provide incomplete visual signals, causing the lip-reading process to misunderstand the speech. Figure 6.4 depicts some visual signals produced by VSP subjects (Male-2 and Male-14), compared to a non-VSP subject (Female-2). Note that from the result in Table 6.6, Male-2 and Male-14 have the lowest word recognition rate, 61% and 53% respectively (exception Male-3 whose result was low because of the facial hair, see Figure 6.3). By examining all the videos and the visual signals for these two subjects, it was found that they are VSP, i.e. they do not provide a meaningful visual signal while talking.

| Word | Speaker | Visual word |
|------|---------|-------------|
| **Two** | Male 2 | 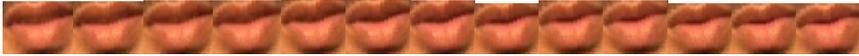 |
|  | Male14 | 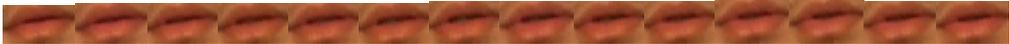 |
|  | Female2 | 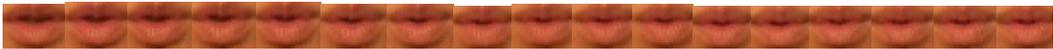 |
| **Cold** | Male 2 | 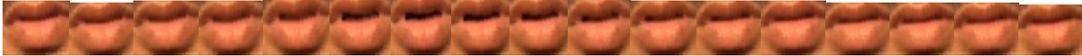 |
|  | Male14 | 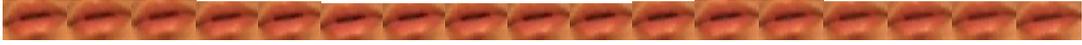 |
|  | Female2 | 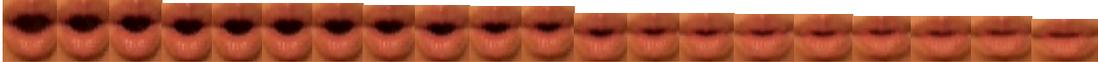 |

**Figure 6.4**. Illustrating Visual speechless person concept.

As can be seen from the previous figure, the mouth dynamic of both words ("two" and "cold") for both VSP subjects (Male-2 and Male-14) did not exhibit big change in the shape and the appearance of the mouth along the video frames. This means that different words can produce similar signals, and that leads to errors and reduces the word recognition rate for such subjects. VSPs are not all the same, they may have different abilities to produce the visual signal. When Male-2 (for instance) utters the word "cold", some differences – especially in the middle of the sequences – can be noticed, unlike Male-14 uttering the same word, which makes Male-2's ability to produce the visual signal slightly better than Male-14's. The results in Table 6.6 back up this conclusion; hence the word recognition rate for Male-2 and Male-14 are 61% and 53% respectively.

The leave-one-sample-out test protocol, which was used in the SD previous experiments, may lead to biased results, because these experiments were mainly conducted by training samples from a video, while testing a sample from the same video. This increases the effect of the video characteristics such as illumination, video digitising, the appearance of the face at that time and any other video-specific characteristics. These video-specific characteristics may affect the visual words features, by giving different features for the same word that comes from different videos, particularly the image-based features.

Although these factors are the same in the same video, they will not necessarily help any word to be better recognized, but as this is not the case in the real world, the system is trained using database videos, and the test samples come from other videos.

For that reason, new SD experiments (called SD2) were designed to evaluate the system in a more realistic way. The same approach leave-one-out cross validation is used, but this time by training a speaker's video from session 2, and testing the speaker's video from session 1, so the training set contains samples for the same person (session 2), and the test set contains samples for the same person, but from a different video (session 1) (see Table 6.7).

According to De Land (1931), Bauman (2000) and the human lip-reading experiment (see chapter 2), only half (or less) of English sounds can be seen and the reader must supply mentally (logically) those words that he/she has missed. For the computer to do so, some kind of information source should be added to aid the recognition process. A language model can be used to supply such information.

Such a model can be constrained by syntax and semantic fixed rules, which can be used to discard the predicted words, if they were not following such rules. For instance, a normal English sentence contains a *subject* followed by a *verb* followed by an *object.*

For example: "*The cat eats the mouse*". If the verb came before the subject (*eats the cat the mouse*), such a sentence is not likely to be said in the English language, and it breaks one of the syntax rules. Also it is less likely (in any language) to say: "*The mouse eats the cat*", because this sentence also breaks one of the semantic rules which might be: "*Mice cannot eat cats*".

It is difficult to design language rules using the in-house database, which was used for this study, because there are only 30 words in the database. It is also difficult to investigate such a model in the limited time of this PhD study, but the above results certainly suggest that developing a language model with the proposed visual words is a promising approach that we would follow up as a future work. Alternatively, and just to simulate the effect of incorporating such a model within the visual words system, we opted for using only one language rule, which says: "if the predicted word did not belong to the test word's group, then discard that word, and check the second nearest word", and so on until the predicted word is one of the test word's group (Nu, LAL1, LAL2, LG, or SEC).

**Table 6.7.** The new speaker-dependent (SD2) experiments using WKNN, DTW, and interpolation; the second column shows the effect of using a language model.

| Subject | SD2 | Language Model simulation |
|---------|-----|---------------------------|
| Female01 | 55% | 73% |
| Female02 | 68% | 81% |
| Female03 | 65% | 73% |
| Female04 | 41% | 57% |
| Female05 | 17% | 40% |
| Female07 | 29% | 55% |
| Female08 | 52% | 68% |
| Female09 | 50% | 57% |
| Female10 | 62% | 70% |
| Male01 | 44% | 64% |
| Male02 | 41% | 51% |
| Male03 | 31% | 45% |
| Male04 | 13% | 41% |
| Male05 | 36% | 57% |
| Male06 | 23% | 50% |
| Male07 | 64% | 71% |
| Male08 | 12% | 37% |
| Male09 | 69% | 81% |
| Male10 | 46% | 68% |
| Male11 | 70% | 75% |
| Male13 | 53% | 77% |
| Male14 | 39% | 49% |
| Male15 | 49% | 58% |
| Male16 | 57% | 72% |
| Male17 | 43% | 55% |
| **Average** | **45%** | **61%** |

This is not the case in the real world (English language model), but it is just a simulation, to show how much extra (a priori) information can be beneficial to the system. The accuracy increased by 16% on average using the language model simulation (see Table 6.7).

The accuracy of the system using SD2 dropped by about 20% compared to the SD results in Table 6.5. This drop is accounted for by use of image-based features such as the appearance of the tongue and teeth in the mouth image. These features vary from video to video due to several factors such as: illumination, video digitising, the appearance of the face at that time including face pose and rotation, etc. Better features and pre-processing methods for Vwords will be studied in the future.

## 6.5 Speaker-independent experiments

The SI experiments were conducted under the same test protocol that was discussed in section 6.1. The visual feature data for each subject in the database was a test set, and those for the rest of the subjects formed the training set each time. WKNN was used as a recognition algorithm. Linear interpolation was used to stretch the smaller signal to the size of the longer one before the test. The experiment was repeated 36 times for each subject, 9 times using ED as a distance measure, each time with a different number of the nearest neighbours (k=1 to 9); 9 times using DTW as a distance measure, 9 times using ED considering the language model, and 9 times using ED considering the language model. The results of the experiments are presented in Table 6.8, below.

Unlike the SD experiments, the best results were found when k is greater than 1, because in the SI experiments there are many examples of the same tested word; the more nearest neighbours there are, the more likely it is to find the optimal word. Despite the results being relatively very low in the SI experiments, the female word recognition average rate (36.29%) was better than the male (30.14%) – using DTW. This backs up the SD results, where the female results were found to be better, for the same reasons discussed in the SD experiments section.

Although the word recognition rate alone cannot be used to identify the VSPs (visual speechless persons), as some subjects results might be low because of other factors such as the video quality, for instance. In this study we did not develop an automatic method to identify them apart from the word recognition rate, which is still a good indicator. These experiments (SI) again show that VSPs' word recognition rates were the lowest (14.67%) for our VSPs (Male-2 and Male-14) which agrees with the SD results (see Table 6.6), where a VSP's ability to produce detectable visual signals while talking is less than others.

Furthermore, it can be noticed from Table 6.8 that the overall result of the SI experiments (average is 32.51%-DTW) is much less than that of the SD (76.38%- Table 6.6). These low results can be justified by the "bad examples" (outliers), which were fed to the training set. These "bad examples" were produced by subjects with facial hair (moustaches and beards), and VSP. Training the computer to take account of unrelated, contradicted and heterogeneous examples (bad examples) seem to mislead the VW algorithm.

**Table 6.8**. Speaker-independent experiments using WKNN, ED or DTW, and interpolation. Taking the maximum result of (k=1 to 9).

|            | Without language model | | Using language model | |
| --- | --- | --- | --- | --- |
| **Subject** | ED | DTW | ED | DTW |
| Female_01 | 23.33% | 27.33% | 50.00% | 48.67% |
| Female_02 | 44.00% | 48.67% | 62.67% | 62.00% |
| Female_03 | 40.67% | 35.33% | 63.33% | 57.33% |
| Female_04 | 40.00% | 38.67% | 62.67% | 62.00% |
| Female_05 | 23.33% | 26.00% | 54.00% | 52.00% |
| Female_06 | 42.95% | 42.95% | 55.70% | 60.40% |
| Female_07 | 20.67% | 31.33% | 45.33% | 49.33% |
| Female_08 | 24.00% | 29.33% | 48.00% | 47.33% |
| Female_09 | 33.33% | 41.33% | 50.00% | 54.67% |
| Female_10 | 37.33% | 42.00% | 56.00% | 60.67% |
| Male_01 | 46.98% | 42.95% | 67.79% | 62.42% |
| Male_02 | 14.67% | 14.67% | 40.00% | 36.67% |
| Male_03 | 20.67% | 23.33% | 41.33% | 35.33% |
| Male_04 | 12.67% | 16.00% | 38.67% | 39.33% |
| Male_05 | 37.58% | 36.24% | 57.05% | 57.05% |
| Male_06 | 18.00% | 19.33% | 42.67% | 41.33% |
| Male_07 | 30.20% | 35.57% | 57.05% | 54.36% |
| Male_08 | 24.00% | 27.33% | 46.00% | 42.00% |
| Male_09 | 36.00% | 41.33% | 50.00% | 50.67% |
| Male_10 | 30.00% | 31.33% | 48.67% | 49.33% |
| Male_11 | 57.33% | 53.33% | 70.00% | 64.67% |
| Male_13 | 27.33% | 26.00% | 54.67% | 52.00% |
| Male_14 | 13.33% | 14.67% | 44.00% | 48.67% |
| Male_15 | 32.00% | 33.33% | 56.67% | 54.00% |
| Male_16 | 34.67% | 39.33% | 57.33% | 64.00% |
| Male_17 | 24.83% | 27.52% | 53.69% | 56.38% |
| **Average** | **30.38%** | **32.51%** | **52.82%** | **52.41%** |
| Females | 32.96% | 36.29% | 54.77% | 55.44% |
| Males | 28.77% | 30.14% | 51.60% | 50.51% |

The training set contains native and non-native speaker subjects, and each of the non-native speakers has his/her own way of uttering English words, for example the word "determine" is pronounced in 3 different ways by the non-native subjects, "di-tur-min", "de-teir-main" and "de-tir-men". This gives the training set different signatures for the same word, which confuses the recognition algorithm and contributes to the "bad examples" pool. Moreover, the training set contains different ethnic groups, African, Asian, Middle Eastern and European; these groups are very different in appearance. Furthermore, there are also differences in the appearance of males versus females, and the differences between age groups, i.e. different colours and shapes of the lips and mouth region. This variety leads to different features being extracted from the same word, which again contributes to the "bad examples" pool and leads to unexpected results.

Even if we consider only one ethnic (race) group, only native or non-native, only males or females, and only one age group, individuals (from the same category) will still have differences in the mouth appearance, and in the way they talk. Everyone has his/her own distinctive appearance and way of speech, which emphasizes that the visual speech recognition problem is a **speaker-dependent problem**. This can be beneficial, because this phenomenon can be used effectively for **speaker identification and/or verification**.

The aforementioned problems decrease the accuracy of the speaker-independent experiments; as well as the main problem of any VSR, which is the lack of information provided by the visual aspect of speech. This can be alleviated using a language model. As in these experiments, the word recognition rate (again) increased by about 20% on average using the language model simulation.

Some researchers such as Jun and Hua (2009) reported the same large difference in the word recognition rate between the speaker-dependent and speaker-independent, 65.9% and 23% respectively. Their justification was based on the feature extraction approach that they used, since they used the DCT of the ROI to extract the features, which are, in this case, image-transformed-based features. These features are easily affected by the appearance of the mouth and the lighting conditions, giving different values when different persons utter the same word.

However, in our investigations we used a hybrid of geometric, appearance and transformed features, which leave little effect on the appearance, and have more effect on the real signature of the uttered word, regardless of the speaker. Thus Jun's and Hua's (2009) justification for this phenomenon gives a partial applied explanation. Nevertheless, their results back up the claim that VSR is a speaker-dependent problem.

In summary, we argue that each person's style of speech has a more distinguishable visual speech than others due to individual differences, not only in appearance, but also, and most importantly, in the way that people talk. Accordingly, we predict that a speaker can be identified using his/her own spoken visual words. This argument will be further investigated in chapter 7, as an example of one application of the proposed VW system among other related applications.

## 6.6 Comparison with other studies

The work reported in this thesis was evaluated using a different database from the ones in the literature, so the comparison with other researchers' work should be considered in this context. Table 6.9, below, presents the results of our experiments alongside those from other methods, and for a more meaningful comparison, the table includes the population size in the databases used in each study as well as the corresponding number/type of words.

**Table 6.9**. Results of some VSR studies.

| Study | SD WRR | SI WRR | Database subjects | Database words |
|---|---|---|---|---|
| The proposed Vwords -average Depend on the speaker in the range of | 76.38% 44-97% | 32.51% 14-55% | 26 | 30 |
| Human perception experiment (Ch2) | 53% | N/A | 4 | 9 Digits |
| Jun and Hua (2009) | 65% | 23% | 1 person SD 40 persons SI | 10 for SD 100 SI |
| Yu (2008) | 80%-90% | N/A | 2 | 50 and 20 |
| Lucey and Sridharan (2008) | 47% | N/A | 33 | Digits |
| Werda, et al. (2007) | N/A | 72.73% | 42 | French vowels |
| Belongie and Weber (1995) | 98% | 47% | 2 | 7 digits |

The performance of the proposed automatic lip-reading system is significantly better than the human perception, despite the larger database used (more subjects and more words) for evaluating the automatic system, compared to the small database used for the human experiment. This may be due to the fact that the computer has access to more visual information than human: each video produces around 30 frames per second, which makes it very difficult for the human eye to extract information from each frame at such a speed. According to Bauman (2000), humans can make approximately 13 to 15 movements per second while talking, although their eyes can only follow 8 or 9 of these movements.

As can be noticed in Table 6.9, the results of the different systems depend very much on the database used; the less words used and subjects participating, the more accurate their method will be. Therefore, it is probably meaningless to compare the proposed method with other researchers' work. Hence "experimental results are only valid for a given dataset and that the comparison of two algorithms should be done on the same data using the same testing protocol" (Cardinaux, 2005:21).

It is important for this study to be compared with the most frequently used approach for VSR, which is the visemes approach. Using the same data and test protocols, visemes were mostly classified using hidden Markov models (HMM). HMM is also used to recognize the words using their visemes sequences. This is the subject of the next section.

## 6.7 Hidden Markov Model (HMM)

Hidden Markov Models are statistical models consisting of a number of nodes that represent hidden states, the nodes being connected by links recounting the conditional probabilities of the transitions between these states. For each hidden state there exists a probability set of emitting visible states (Duda et al., 2001). All the HMM probabilities (parameters) can be learned from a sample sequence (training set), and then HMM can be used for classifying an input (sequences).

HMMs were first introduced by Baum and collaborators (Baum and Petrie, 1966) (Baum et al, 1970). Since then they have been used extensively for speech, DNA sequences and optical character recognition, and later for visual speech recognition. HMMs' great success in these applications demonstrates their power in modelling sequences. A typical HMM can be defined as a 5-tuple $\lambda$,

$$\lambda = (S, V, A, B, \pi)$$

where,

$S = \{s_1, s_2, s_3, \ldots, s_n\}$ set of all the hidden states.

$V = \{v_1, v_2, v_3, \ldots, v_m\}$ set of all observations (observed states).

$A = \{a_{ij}\}$ transitions' matrix ($1 \leq i, j \leq n$), the probability to move from a hidden state to the other. A's size = $n^2$.

$B = \{b_{jk}\}$ observations' matrix $(1 \leq j \leq n)$ and $(1 \leq k \leq m)$, the probability of a hidden state to emit a particular observed state. B's size = $nm$

$\pi = \{\pi(i)\}$ start probabilities vector $(1 \leq i \leq n)$. It contains prior probability of the model, which is given by an external source, for example if HMM models a language, like the speech recognition problem, the prior probability might be defined by the syntax or the semantic rules of the modelled language; if there is no such information, this part can be neglected (Duda, et al., 2001).

$n$ is the number of the hidden states, and $m$ is the number of the observations.

An example of an HMM showing all the mentioned parameters is shown in Figure 6.5.

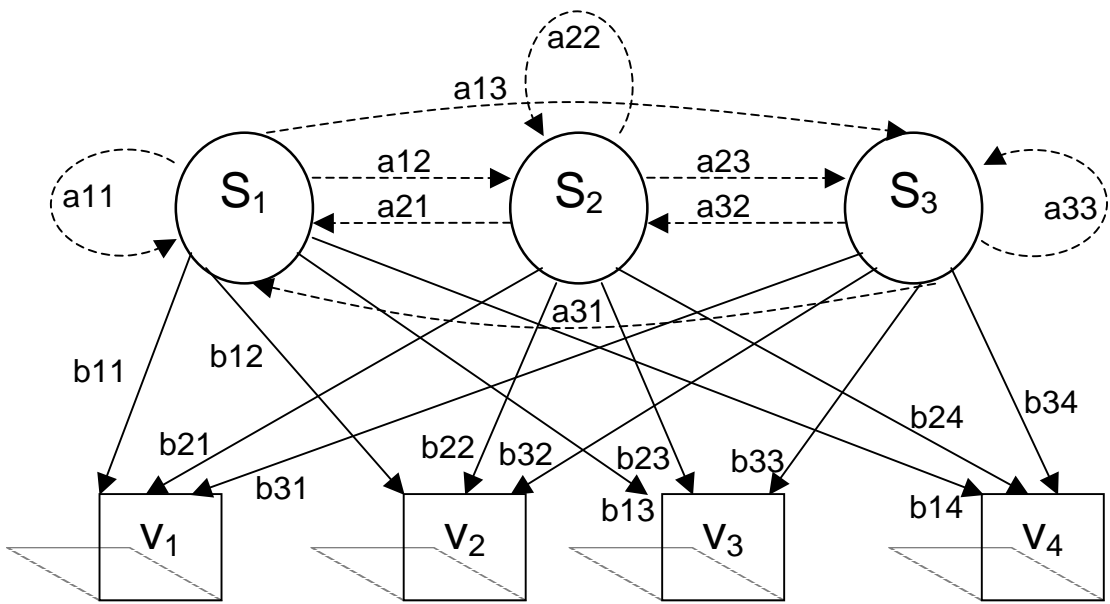

**Figure 6.5**. HMM of 3 states & 4 observations, dotted lines depict the transitions probabilities, and the solid lines depict the probabilities of the emitted observations.

The probability of the current state is calculated by using the probability of the previous state (using conditional probability, e.g. $P(S_2|S_1)$, which is called first order HMM, but if both of the previous states were needed to determine the current state, then it is called second order HMM, e.g. $P(S_3|S_1,S_2)$, and so on. However, the more the previous states are needed, the more complex the system will become.

HMM can be trained using a proper training set (observation sequences), to be ready to be used in an application, either to find the best model that fits an observation sequence, or to generate the best (most likely) sequence from the hidden states that is indicated by an observation sequence. These problems are called HMM training, evaluation and decoding problem respectively (Rabiner, 1989).

- *Training Problem*; given observation-Sequence *O*. What is the most likely HMM $\lambda$ for *O*? This question can be answered using the Baum-Welch-Algorithm to generate an HMM given a sequence of observations.

- *Evaluation Problem*; given an HMM $\lambda$, and observation-Sequence *O*. What is the probability that $\lambda$ has created *O*? This question can be answered using the Forward-Backward-Algorithm, to find the probability of an observed sequence, giving an HMM and hence choose the most probable HMM that best fits the input observed sequence.

- *Decoding Problem*; given HMM $\lambda$, and observation-Sequence *O*. What is the most likely sequence of hidden states for *O*? This question can be answered using the Viterbi-Algorithm to find the sequence of hidden states that most likely generates an observed sequence.

More information about HMM and their associated algorithms can be found in (Rabiner, 1989) and (Duda, et al., 2001) references.

## 6.7.1 HMM for visual speech recognition

Most visual speech recognition systems in the literature use HMM for both visemes classification and word recognition. Among these publications is the work of: Bregler and Konig (1994), Luettin et al. (1996c), Potamianos et al. (2003), Foo and Lian (2004), Hazen et al. (2004), Gurban and Thiran (2005), Leszczynski and Skarbek, (2005), Arsic and Thiran (2006), Sagheer, et al. (2006), Alizadeh et al. (2008), Yu (2008), and Lucey and Sridharan (2008).

Generally, the sequence of images that represent a particular viseme trains a 3 states HMM, each state representing a part of that viseme. The initial state represents the image sequences when the viseme starts to develop from silence, or from another viseme. The articulation state is part of the viseme when it takes its actual shape, and describes the largest variation in lip dynamic and the final state, which includes the visemes' image sequence, when it starts to decline to silence (relax), or to start another viseme (Yu, 2008).

Table 6.10. Viseme Model of MPEG-4 standard for English (Yu, 2008)[*].

| Viseme Number | Mapped Phonemes | Example Words | Vowels or Consonants | Image sequence example (from the new database) |
|---|---|---|---|---|
| 1 | [b],[p],[m] | Bomb, Practical, Determine | consonants | B P M |
| 2 | [s], [z] | Six, Zero | consonants | S Z |
| 3 | [ch], [dZ] | Situation, | consonants | Ch |
| 4 | [f], [v] | Four, Five | consonants | F V |
| 5 | [t], [d] | Kite, Fold | consonants | T D |
| 6 | [k], [g] | Kill, Gun | consonants | K G |
| 7 | [n],[ l] | Nine, Light | consonants | N N L |
| 8 | [Th] | Three | consonants | Th |
| 9 | [r] | Run Four | consonants | R R |
| 10 | [I] | Fire | vowel | I |
| 11 | [A:] | Appreciate | vowel | |
| 12 | [e] | Seven | vowel | E E |
| 13 | [O] | One Fold | vowel | O O |
| 14 | [U] | Two University | vowel | U U |

---

[*] The phonemes symbols in tables 6.10 and 6.11 are not compatible with the SAMPA or with the IPA standards, because they were taken from the viseme model of MPEG-4 standard for English.

For each *viseme* in the database, a 3 states HMM was designed, and the output (most likely) visemes sequences is recognized as a word by means of another HMM. For each *word* in the database, an HMM with a different number of states is designed, the number of the states being the number of the visemes appearing in a specific word. The MPEG-4 standard visemes set for English language was adopted for this study (see Table 6.10). Accordingly, each word is represented by a sequence of visemes. The following table (6.11) shows all the visemes and phonemes in the dataset of words that we have used in our database.

**Table 6.11**. Phonemes and visemes representation of the in-house database words.

| Word | Phonemes | Visemes |
|---|---|---|
| Zero | [z], [e], [r], [O] | 2, 12, 9, 13 |
| One | [O],[n] | 13, 7 |
| Two | [t], [U] | 5, 14 |
| Three | [Th], [r], [e] | 8, 9, 12 |
| Four | [f], [O], [r} | 4, 13, 9 |
| Five | [f], [I], [v] | 4, 10, 4 |
| Six | [s],[e], [k], [s] | 2, 12, 6, 2 |
| Seven | [s], [e], [v], [e], [n] | 2,12,4,12,7 |
| Eight | [A:], [t] | 11, 5 |
| Nine | [n], [I], [n] | 7, 10, 7 |
| Fold | [f],[O],[l],[d] | 4, 13, 7, 5 |
| Sold | [s],[O],[l],[d] | 2, 13, 7, 5 |
| Hold | [O],[l],[d] | 13, 7, 5 |
| Bold | [b],[O],[l],[d] | 1, 13, 7, 5 |
| Cold | [k],[O],[l],[d] | 6, 13, 7, 5 |
| Knife | [n],[I],[f] | 7, 10, 4 |
| Night | [n],[I],[t] | 7, 10, 5 |
| Kite | [k],[I],[t] | 6, 10, 5 |
| Light | [l],[I],[t] | 7, 10, 5 |
| Fight | [f],[I],[t] | 4, 10, 5 |
| Appreciate | [A:], [p], [r], [e], [ch], [e], [A:], [t] | 11, 1, 9, 12, 3, 12, 11, 5 |
| University | [U], [n], [e], [v], [A:], [r], [s], [e], [t], [e] | 14, 7, 12, 4, 11, 9, 2, 12, 5, 12 |
| Determine | [d], [e], [t], [r], [m], [e], [n] | 5, 12, 5, 9, 1, 12, 7 |
| Situation | [s], [e], [t], [U], [A:], [ch], [e], [n] | 2, 12, 5, 14, 11, 3, 12, 7 |
| Practical | [p], [r], [A:], [k], [t], [e], [k], [A:], [l] | 1, 9, 11, 6, 5, 12, 6, 11, 7 |
| Bomb | [b], [O], [m], [b] | 1, 13, 1, 1 |
| Kill | [k], [e], [l] | 6, 12, 7 |
| Run | [r], [A:], [n] | 9, 11, 7 |
| Gun | [g], [A:], [n] | 6, 11, 7 |
| Fire | [f], [I], [r] | 4, 10, 9 |

The HMM library which was used for this study was designed by Murphy (2009) using Matlab® software. Again two experiments were conducted using HMM to classify the 14 visemes, and then to recognize the specific word, the speaker-dependent experiment, and the speaker-independent experiment. The evaluation protocol for each experiment is the same as the one for the visual words speaker-dependent (leave-one-viseme-example-out and train the rest of the visemes of the same speaker), and the one for the speaker-independent experiments (leave-one-speaker-out and train visemes from the other speakers). Table 6.12 contains the results of the speaker-dependent experiment for visemes recognition, while Table 6.13 contains the results of the speaker-independent experiment.

**Table 6.12.** Speaker-dependent visemes recognition rate using HMM

| Visemes Subject | All | 1 | 2 | 3 | 4 | 5 | 6 | 7 | 8 | 9 | 10 | 11 | 12 | 13 | 14 |
|---|---|---|---|---|---|---|---|---|---|---|---|---|---|---|---|
| Female_01 | 0.17 | 0.22 | 0.09 | 0.33 | 0.36 | 0.11 | 0.00 | 0.00 | 0.00 | 0.14 | 0.83 | 0.07 | 0.04 | 0.43 | 0.00 |
| Female_02 | 0.25 | 0.33 | 0.09 | 0.33 | 0.71 | 0.07 | 0.00 | 0.10 | 1.00 | 0.21 | 0.75 | 0.14 | 0.04 | 0.79 | 0.25 |
| Female_03 | 0.27 | 0.56 | 0.55 | 0.67 | 0.07 | 0.21 | 0.00 | 0.03 | 1.00 | 0.07 | 0.83 | 0.07 | 0.12 | 1.00 | 0.00 |
| Female_04 | 0.23 | 0.56 | 0.82 | 0.00 | 0.21 | 0.25 | 0.00 | 0.10 | 0.00 | 0.14 | 0.50 | 0.07 | 0.19 | 0.21 | 0.25 |
| Female_05 | 0.32 | 0.56 | 0.09 | 0.67 | 0.64 | 0.46 | 0.27 | 0.03 | 1.00 | 0.29 | 0.50 | 0.57 | 0.04 | 0.50 | 0.25 |
| Female_06 | 0.15 | 0.22 | 0.27 | 0.33 | 0.07 | 0.11 | 0.00 | 0.06 | 0.00 | 0.00 | 0.08 | 0.07 | 0.42 | 0.14 | 0.50 |
| Female_07 | 0.23 | 0.44 | 0.18 | 1.00 | 0.71 | 0.04 | 0.00 | 0.03 | 0.00 | 0.07 | 0.42 | 0.07 | 0.12 | 0.86 | 0.25 |
| Female_08 | 0.27 | 0.89 | 0.09 | 1.00 | 0.14 | 0.14 | 0.27 | 0.16 | 0.00 | 0.14 | 0.92 | 0.07 | 0.04 | 0.57 | 0.50 |
| Female_09 | 0.22 | 0.56 | 0.45 | 0.00 | 0.29 | 0.25 | 0.18 | 0.03 | 0.00 | 0.07 | 0.75 | 0.29 | 0.08 | 0.21 | 0.00 |
| Female_10 | 0.28 | 0.44 | 0.55 | 0.33 | 0.79 | 0.21 | 0.00 | 0.10 | 0.00 | 0.14 | 0.75 | 0.07 | 0.00 | 0.71 | 0.00 |
| Male_01 | 0.19 | 0.44 | 0.55 | 0.00 | 0.21 | 0.04 | 0.18 | 0.06 | 1.00 | 0.00 | 0.58 | 0.00 | 0.00 | 0.64 | 0.25 |
| Male_02 | 0.20 | 0.56 | 0.09 | 0.67 | 0.21 | 0.32 | 0.00 | 0.13 | 0.00 | 0.00 | 0.92 | 0.07 | 0.00 | 0.14 | 0.00 |
| Male_03 | 0.26 | 0.67 | 0.27 | 0.00 | 0.36 | 0.50 | 0.18 | 0.06 | 0.00 | 0.14 | 0.42 | 0.21 | 0.04 | 0.29 | 0.50 |
| Male_04 | 0.17 | 0.44 | 0.00 | 1.00 | 0.21 | 0.00 | 0.09 | 0.06 | 0.00 | 0.07 | 0.83 | 0.14 | 0.00 | 0.43 | 0.00 |
| Male_05 | 0.15 | 0.78 | 0.82 | 0.33 | 0.36 | 0.04 | 0.00 | 0.00 | 0.00 | 0.00 | 0.33 | 0.00 | 0.00 | 0.14 | 0.00 |
| Male_06 | 0.19 | 0.22 | 0.00 | 0.00 | 0.14 | 0.07 | 0.00 | 0.03 | 1.00 | 0.14 | 0.33 | 0.00 | 0.54 | 0.64 | 0.00 |
| Male_07 | 0.20 | 0.44 | 0.55 | 0.33 | 0.14 | 0.11 | 0.27 | 0.03 | 0.00 | 0.07 | 0.50 | 0.21 | 0.08 | 0.36 | 0.50 |
| Male_08 | 0.19 | 0.00 | 0.00 | 1.00 | 0.14 | 0.07 | 0.00 | 0.10 | 0.00 | 0.00 | 0.83 | 0.00 | 0.08 | 0.86 | 0.50 |
| Male_09 | 0.22 | 0.11 | 0.36 | 0.00 | 0.64 | 0.29 | 0.18 | 0.03 | 1.00 | 0.00 | 0.67 | 0.14 | 0.08 | 0.36 | 0.00 |
| Male_10 | 0.21 | 0.67 | 0.64 | 0.67 | 0.14 | 0.14 | 0.09 | 0.10 | 1.00 | 0.00 | 0.83 | 0.00 | 0.00 | 0.14 | 0.50 |
| Male_11 | 0.31 | 0.67 | 0.00 | 1.00 | 0.57 | 0.57 | 0.27 | 0.19 | 0.00 | 0.07 | 0.83 | 0.07 | 0.00 | 0.21 | 0.75 |
| Male_13 | 0.16 | 0.33 | 0.00 | 0.00 | 0.21 | 0.32 | 0.09 | 0.03 | 1.00 | 0.07 | 0.17 | 0.14 | 0.04 | 0.43 | 0.00 |
| Male_14 | 0.14 | 0.67 | 0.00 | 0.00 | 0.29 | 0.04 | 0.00 | 0.03 | 0.00 | 0.00 | 0.83 | 0.07 | 0.00 | 0.14 | 0.50 |
| Male_15 | 0.28 | 0.67 | 0.09 | 0.00 | 0.57 | 0.61 | 0.18 | 0.10 | 1.00 | 0.07 | 0.83 | 0.00 | 0.08 | 0.07 | 0.50 |
| Male_16 | 0.16 | 0.22 | 0.00 | 0.67 | 0.14 | 0.07 | 0.27 | 0.00 | 1.00 | 0.00 | 0.83 | 0.07 | 0.19 | 0.14 | 0.00 |
| Male_17 | 0.13 | 0.00 | 0.00 | 0.00 | 0.00 | 0.78 | 0.00 | 0.00 | 0.00 | 0.00 | 0.08 | 0.00 | 0.00 | 0.14 | 0.00 |
| Average | 0.21 | 0.45 | 0.25 | 0.40 | 0.32 | 0.22 | 0.10 | 0.06 | 0.38 | 0.07 | 0.62 | 0.10 | 0.08 | 0.41 | 0.23 |
| Females | 23.85% | | | | | 25.13% | | | | | | | 28.90% | | |
| Males | 19.69% | | | | | Consonants | | | | | | | Vowels | | |
| | 21.84% | Excluding moustache & beard | | | | | | | | | | | | | |
| | 20.25% | Moustache & beard | | | | | | | | | | | | | |

Despite having a relatively high visemes recognition rate (32%) for Female 5, there is still a very low rate (13%) for Male 17 (for instance), and the overall average is only (21%). These low results were due to the general problems associated with the visemic approach (discussed earlier in chapter 2). For example, we can find the same viseme with different signatures depending on its location in the word; a viseme following silence (for instance) becomes different when it follows another viseme, or when it is followed by silence. So this viseme should be categorized as 2, 3 or more different visemes if it gives different signatures. This raises the question whether it is sufficient to use only 14 visemes or should we add extra versions of these 14 visemes. The proposed visual word approach overcomes these problems, but to stay within the viseme concept, another solution proposed by Yu (2008) was using 60 visual speech units as an alternative to visemes. This solves the problem of the small number of visemes, but his visual speech units can be interpreted as visemes, which takes this solution back to the other problems associated with visemes.

**Table 6.13**. Speaker-independent results for visemes recognition using HMM.

| Visemes Subject | All | 1 | 2 | 3 | 4 | 5 | 6 | 7 | 8 | 9 | 10 | 11 | 12 | 13 | 14 |
|---|---|---|---|---|---|---|---|---|---|---|---|---|---|---|---|
| Female_01 | 5.44% | 0.03 | 0.00 | 0.00 | 0.00 | 0.00 | 0.03 | 0.01 | 0.00 | 0.00 | 0.67 | 0.09 | 0.00 | 0.00 | 0.00 |
| Female_02 | 6.43% | 0.00 | 0.00 | 0.00 | 0.00 | 0.00 | 0.14 | 0.01 | 0.00 | 0.05 | 0.67 | 0.07 | 0.01 | 0.02 | 0.00 |
| Female_03 | 9.88% | 0.00 | 0.00 | 0.00 | 0.16 | 0.00 | 0.09 | 0.03 | 0.60 | 0.11 | 0.74 | 0.02 | 0.00 | 0.20 | 0.00 |
| Female_04 | 7.74% | 0.07 | 0.03 | 0.00 | 0.05 | 0.04 | 0.11 | 0.03 | 0.00 | 0.02 | 0.72 | 0.00 | 0.00 | 0.07 | 0.00 |
| Female_05 | 11.04% | 0.00 | 0.00 | 0.00 | 0.02 | 0.00 | 0.09 | 0.09 | 0.80 | 0.30 | 0.92 | 0.00 | 0.00 | 0.02 | 0.00 |
| Female_06 | 9.62% | 0.17 | 0.09 | 0.00 | 0.00 | 0.00 | 0.11 | 0.20 | 0.20 | 0.02 | 0.54 | 0.02 | 0.00 | 0.07 | 0.00 |
| Female_07 | 5.44% | 0.00 | 0.00 | 0.00 | 0.02 | 0.01 | 0.06 | 0.00 | 0.00 | 0.16 | 0.49 | 0.00 | 0.00 | 0.02 | 0.13 |
| Female_08 | 9.72% | 0.07 | 0.00 | 0.00 | 0.18 | 0.04 | 0.03 | 0.00 | 0.00 | 0.09 | 0.97 | 0.02 | 0.00 | 0.04 | 0.00 |
| Female_09 | 8.90% | 0.00 | 0.03 | 0.00 | 0.02 | 0.00 | 0.00 | 0.02 | 0.20 | 0.09 | 0.82 | 0.18 | 0.00 | 0.11 | 0.00 |
| Female_10 | 8.90% | 0.00 | 0.03 | 0.00 | 0.02 | 0.00 | 0.00 | 0.02 | 0.20 | 0.09 | 0.82 | 0.18 | 0.00 | 0.11 | 0.00 |
| Male_01 | 8.51% | 0.00 | 0.00 | 0.00 | 0.09 | 0.00 | 0.00 | 0.04 | 0.60 | 0.09 | 0.90 | 0.00 | 0.00 | 0.02 | 0.00 |
| Male_02 | 6.43% | 0.00 | 0.00 | 0.00 | 0.00 | 0.00 | 0.00 | 0.00 | 0.00 | 0.00 | 0.97 | 0.02 | 0.00 | 0.00 | 0.00 |
| Male_03 | 6.75% | 0.00 | 0.00 | 0.00 | 0.05 | 0.00 | 0.06 | 0.00 | 0.00 | 0.00 | 0.90 | 0.00 | 0.01 | 0.02 | 0.00 |
| Male_04 | 6.92% | 0.00 | 0.00 | 0.00 | 0.02 | 0.01 | 0.03 | 0.08 | 0.00 | 0.02 | 0.74 | 0.00 | 0.00 | 0.00 | 0.07 |
| Male_05 | 13.10% | 0.17 | 0.06 | 0.10 | 0.07 | 0.01 | 0.06 | 0.11 | 0.00 | 0.07 | 0.95 | 0.00 | 0.04 | 0.27 | 0.00 |
| Male_06 | 6.43% | 0.00 | 0.00 | 0.00 | 0.00 | 0.00 | 0.00 | 0.00 | 0.00 | 0.00 | 1.00 | 0.00 | 0.00 | 0.00 | 0.00 |
| Male_07 | 8.11% | 0.03 | 0.00 | 0.00 | 0.05 | 0.00 | 0.00 | 0.07 | 0.00 | 0.00 | 0.97 | 0.02 | 0.00 | 0.02 | 0.00 |
| Male_08 | 6.10% | 0.17 | 0.00 | 0.00 | 0.23 | 0.00 | 0.03 | 0.00 | 0.40 | 0.05 | 0.36 | 0.00 | 0.00 | 0.07 | 0.00 |
| Male_09 | 8.24% | 0.13 | 0.00 | 0.00 | 0.20 | 0.00 | 0.03 | 0.00 | 0.40 | 0.00 | 0.74 | 0.07 | 0.00 | 0.04 | 0.00 |
| Male_10 | 9.23% | 0.07 | 0.00 | 0.10 | 0.11 | 0.01 | 0.00 | 0.06 | 0.00 | 0.09 | 0.85 | 0.00 | 0.00 | 0.07 | 0.07 |
| Male_11 | 20.10% | 0.37 | 0.43 | 0.60 | 0.23 | 0.19 | 0.03 | 0.03 | 1.00 | 0.09 | 0.85 | 0.13 | 0.00 | 0.24 | 0.07 |
| Male_13 | 5.27% | 0.00 | 0.00 | 0.00 | 0.05 | 0.00 | 0.03 | 0.00 | 0.40 | 0.02 | 0.54 | 0.02 | 0.00 | 0.04 | 0.13 |
| Male_14 | 6.75% | 0.10 | 0.00 | 0.00 | 0.00 | 0.00 | 0.00 | 0.01 | 0.00 | 0.00 | 0.95 | 0.00 | 0.00 | 0.00 | 0.00 |
| Male_15 | 10.87% | 0.10 | 0.00 | 0.00 | 0.16 | 0.01 | 0.00 | 0.06 | 0.00 | 0.14 | 0.82 | 0.09 | 0.00 | 0.07 | 0.27 |
| Male_16 | 7.74% | 0.00 | 0.06 | 0.00 | 0.00 | 0.01 | 0.09 | 0.04 | 0.20 | 0.00 | 0.92 | 0.00 | 0.00 | 0.00 | 0.00 |
| Male_17 | 4.98% | 0.03 | 0.00 | 0.00 | 0.02 | 0.00 | 0.17 | 0.02 | 0.40 | 0.00 | 0.28 | 0.00 | 0.00 | 0.16 | 0.00 |
| Average | **8.41%** | 0.06 | 0.03 | 0.03 | 0.07 | 0.01 | 0.05 | 0.04 | 0.21 | 0.06 | 0.77 | 0.04 | 0.00 | 0.07 | 0.03 |
| Females | **8.31%** | \multicolumn{7}{c}{6.04%} | | | | | | 18.10% | | |
| Males | **8.47%** | \multicolumn{7}{c}{Consonants} | | | | | | Vowels | | |
| | **8.54%** | Excluding moustache & beard | | | | | | | | | | | | | |
| | **8.16%** | Moustache & beard | | | | | | | | | | | | | |

**Table 6.14.** Word recognition rate using HMM for visemes sequences, compared to visual words approach. SD1 from Table 6.6, SD2 and Language model from Table 6.7, and the last column from Table 6.8.

| Subject | Speaker-dependent | | | | Speaker-independent | |
|---|---|---|---|---|---|---|
| | Visemes/ HMM | SD1 | SD2 | Language model | Visemes/ HMM | Visual words |
| Female_01 | 33% | 69% | 55% | 73% | 15% | 27% |
| Female_02 | 62% | 97% | 68% | 81% | 18% | 49% |
| Female_03 | 49% | 87% | 65% | 73% | 11% | 35% |
| Female_04 | 32% | 81% | 41% | 57% | 13% | 39% |
| Female_05 | 45% | 83% | 17% | 40% | 7% | 26% |
| Female_06* | 20% | 75% | - | - | 10% | 43% |
| Female_07 | 37% | 82% | 29% | 55% | 5% | 31% |
| Female_08 | 35% | 85% | 52% | 68% | 18% | 29% |
| Female_09 | 37% | 81% | 50% | 57% | 13% | 41% |
| Female_10 | 43% | 88% | 62% | 70% | 12% | 42% |
| Male_01 | 31% | 79% | 44% | 64% | 19% | 43% |
| Male_02 | 20% | 61% | 41% | 51% | 5% | 15% |
| Male_03 | 14% | 44% | 31% | 45% | 7% | 23% |
| Male_04 | 27% | 63% | 13% | 41% | 8% | 16% |
| Male_05 | 37% | 85% | 36% | 57% | 17% | 36% |
| Male_06 | 23% | 65% | 23% | 50% | 2% | 19% |
| Male_07 | 44% | 84% | 64% | 71% | 9% | 36% |
| Male_08 | 19% | 75% | 12% | 37% | 7% | 27% |
| Male_09 | 27% | 92% | 69% | 81% | 6% | 41% |
| Male_10 | 37% | 88% | 46% | 68% | 10% | 31% |
| Male_11 | 52% | 84% | 70% | 75% | 22% | 53% |
| Male_13 | 17% | 69% | 53% | 77% | 5% | 26% |
| Male_14 | 19% | 53% | 39% | 49% | 12% | 15% |
| Male_15 | 28% | 69% | 49% | 58% | 14% | 33% |
| Male_16 | 39% | 83% | 57% | 72% | 16% | 39% |
| Male_17 | 17% | 62% | 43% | 55% | 7% | 28% |
| **Average** | **32%** | **76.38%** | **45%** | **61%** | **11%** | **33%** |
| Females | 39% | 83% | 49% | 64% | 12% | 36% |
| Males | 28% | 72% | 43% | 59% | 10% | 30% |
| Excluding moustache & beard | 35% | 77% | 45% | 61% | 12% | 33% |
| Moustache & beard | 27% | 71% | 44% | 59% | 9% | 30% |

Table 6.13, above, contains the results of the speaker-independent experiment which shows that the visemes problems get worse. This is natural and compatible with the earlier conclusion that the visual speech recognition problem is speaker-dependent. The very low word recognition rates in this experiment are again explained by the visemes problems, but this time on a larger scale; not only the same viseme of the same person gives different signatures, but also some examples of different visemes, from different persons, match the tested viseme better than the actual viseme examples in the training

---

* Female_06 was not evaluated using SD2 and the language model, because this subject was not recorded in Session 2.

set. This situation mixes the visemes together, forming a chaotic pool of features, which makes the learning process very difficult, and gives bad results while evaluating a specific viseme. For example, the viseme recognition rate for Male 17 is (4.98%), which is even less than a random guess (a random guess for a specific viseme out of 14 visemes is (7.1%) if the numbers of examples of each viseme are equal).

The output visemes sequences of each experiment were passed to other HMMs that were designed for each word. The word associated with the HMM with the highest probability (after inputting the observed visemes sequences) is the recognized word. Table 6.14 depicts the word recognition rate using the visemic approach, designed by means of HMM.

It can be noticed from the previous table (6.14) that the word recognition rate of speaker-dependent experiments (32% visemes/HMM), is almost 3 times that of speaker-independent experiments (11% visemes/HMM). Referring again to the visual words approach, the word recognition rate for speaker-dependent was (76.38% SD1, 45% SD2 and 61% when using some kind of language model), and it was only 33% for the speaker-independent experiment. These results emphasize that the visual speech recognition problem is a **speaker-dependent problem**. Our low word recognition rate (32%) using the visemic approach and HMM, *is backed up by Alizadeh et al.'s (2008) VSR method, who used the same software (Murphy (2009)) and they reported almost the same results, 67.65% word error rate, i.e. 32.35% word recognition rate*.

It can be noticed also that the word recognition rates using the visemic approach are very low, which can be justified by the problems of adopting the visemic approach, which were discussed in chapter 2. The main reason is apparently the very low rate in recognizing the viseme itself. A quick look at the visemes recognition table reveals that the average viseme recognition rate is 21%, so the output visemes sequences are true by 21%, and corrupted by 79%, which definitely leads to very low results in the next stage (the word recognition stage).

The results in the previous table (6.14) show that the visual words approach outperforms the visemes approach for the visual speech recognition problem, in both types of experiments (speaker-dependent and independent). The accuracy of the proposed approach is more than twice the accuracy of the visemic approach (76% compared to 32%), for the speaker-dependent experiment, and the same for the speaker-independent (33% compared to 11%).

This comparison is up to standard, and far removed from bias, because all the experiments have been conducted using the same database, the same features, and the same test protocol, for the same type of experiment. Moreover, the visemic approach experiments have been conducted in the same manner that most researchers have done in the literature. Even the HMM software was taken as it is, without any major modifications. This software is well tested and has been used by many researchers for the same purposes (VSR) such as Alizadeh et al. (2008), and for other purposes by other researchers.

## 6.8 Similarity based grouping of visual words

A major drawback of training a "visual word" system is that it is not feasible to train all the English language words. One may ask whether we can group/cluster the English words into a manageable number of groups/clusters using similarity of their visual signatures. If this can be answered positively, then it is enough to train one or a few words from each cluster in order to classify any other untrained word of the same group, which has the same signature. And this is the reason why we used similar words such as *night* and *light* in the in-house database.

There are 2 word groups in the in-house database, where the words in each group have approximately a similar visual signature – they are LAL1 and LAL2. Motivated by the observations made in section 6.4 about the low recognition rates of words in these groups, we designed an experiment to measure such similarity, and to see if some of the words from each group can be used to predict an untrained word (test word). For all the subjects in the database, LAL1 and LAL2 were taken to construct an evaluation file, so that this file contained 10 different words repeated 5 times for each subject.

Using leave-one-word-out cross validation, the WKNN classifier attempted to predict the class of the tested word (we have 2 classes here, LAL1 and LAL2), after removing all the examples (from all the speakers that related to the tested word) from the training set, so the 5 repeated examples (related to the test example) for each subject were removed from the training set, leaving the remaining 4 words from the related group and the 5 words from the other group, all repeated 5 times by each subject (see Figure 6.6).

As can be noted from Table 6.15, the similarity between the words produces a good amount of information, which enabled the algorithm (WKNN) to recognize the group of tested words. The best result was when k=3, although there was no significant difference between the results regarding the number of nearest neighbours (k), hence all the averages in the range 93.77% to 94.07%.

Also it can be noticed that Vwords of LAL1 are more similar to each other than are Vwords of LAL2, hence the average recognition rate was 96.15% and 92.15% for LAL1 and LAL2 respectively.

|  | Subject 1 | Subject 2 | Subject 3 | ……. | Subject n |  |
|---|---|---|---|---|---|---|
| **LAL1** | Fold | Fold | Fold | ……. | Fold | **Test** |
|  | Sold | Sold | Sold | ……. | Sold |  |
|  | Hold | Hold | Hold | ……. | Hold |  |
|  | Bold | Bold | Bold | ……. | Bold |  |
|  | Cold | Cold | Cold | ……. | Cold | **Train** |
| **LAL2** | Knife | Knife | Knife | ……. | Knife |  |
|  | Night | Night | Night | ……. | Night |  |
|  | Kite | Kite | Kite | ……. | Kite |  |
|  | Light | Light | Light | ……. | Light |  |
|  | Fight | Fight | Fight | ……. | Fight |  |

**Figure 6.6**. The word similarity grouping experiment setting.

**Table 6.15.** The word similarity grouping experiment using WKNN with ED and interpolation.

| VW | K=1 | K=2 | K=3 | K=4 | K=5 |
|---|---|---|---|---|---|
| Fold | 95.38% | 95.38% | **96.15%** | **96.15%** | **96.15%** |
| Sold | 95.35% | 95.35% | 93.80% | 95.35% | **96.12%** |
| Hold | **96.92%** | **96.92%** | 95.38% | 94.62% | 93.08% |
| Bold | 98.46% | 98.46% | 98.46% | 99.23% | **100%** |
| Cold | 94.62% | 94.62% | **96.15%** | 94.62% | 95.38% |
| Average | *96.15%* | *96.15%* | *95.99%* | *95.99%* | *96.15%* |
| Knife | 82.31% | 82.31% | 83.08% | **83.85%** | 83.08% |
| Night | 97.67% | 97.67% | 98.45% | **99.22%** | **99.22%** |
| Kite | **93.85%** | **93.85%** | **93.85%** | 92.31% | 92.31% |
| Light | **100%** | **100%** | **100%** | **100%** | **100%** |
| Fight | 84.62% | 84.62% | **85.38%** | 82.31% | 83.08% |
| Average | *91.69%* | *91.69%* | *92.15%* | *91.54%* | *91.54%* |
| Overall average | 93.92% | 93.92% | **94.07%** | 93.77% | 93.84% |

These encouraging results show that it might be possible to group all the English language words into specific groups, based on their visual similarity, then train only some of these words to represent each group instead of training all English language words. These results show the high similarity between such words in both groups LAL1 and LAL2, which justifies the low recognition rates of the previous experiments. However, much work needs to be done to draw a credible conclusion in this respect.

After recognizing the group of the word, we need to find other features that hide the similarity and highlight the differences between these Vwords within the same group. This can be done using a multi level recognition scheme, backed up with a proper language model, which will be left for future work.

## 6.9 Summary

In this chapter several experiments to evaluate the Vwords approach to the automatic lip-reading problem were discussed. These experiments were:

1. Speaker-dependent experiments: the average recognition rate for all subjects is 76.38%, ranging from (44-97%). Several reasons were found that affected the various results, such as the appearance of facial hair, and the individual's ability to produce a clear visual signal, as some subjects produce weak signals (termed as VSP).

2. To remove the effect of video specific characteristics another SD experiment has been designed, SD2 trains the system on subjects from session 2 and test it using subjects from session1. The average word recognition rate was 45% increased to 61% after using a simple language model.

3. Speaker-independent experiments: the average word recognition rate for all subjects is 32.51%-DTW increased to 52.82% after using a simple language model. Ranging from (13.33-57.33%), the performance of SI experiments was less than that of SD. Individual differences in the mouth appearance, and in the way of talking, produce different visual and audio signals for the same spoken word, which emphasizes that the visual speech recognition problem is a speaker-dependent problem. Our experiments show that the speaker-dependent word recognition rate is much higher than that of the speaker-independent, this claim is backed up by several researchers such as Jun and Hua (2009).

4. The visual words approach was compared with the visemic approach, which was modelled by HMM. The experiment results show that the visual words approach outperforms the visemic approach to the visual speech recognition problem, in both types of experiments (speaker-dependent and independent). The accuracy of the proposed approach is more than twice the accuracy of the visemic approach (76% compared to 32%), for the speaker-dependent experiment, and for the speaker-independent (30% compared to 11%).

5. The look-alike experiments were conducted to prove that it is not necessary to train all the English language words to get an automatic lip-reading system. The experiments' excellent average (94%) suggests that it is enough to train one or some of the similar words, rather than training all the similar words.

All these experiments emphasize the superiority of the visual words approach as a solution for the visual speech recognition problem. However, many challenges still remain in this area of research, particularly the large unseen part of speech, as the word recognition rate is greatly affected by its phoneme components, because producing these sounds involves the movement (dynamic) of some parts of the speech production system, and some of these parts can be seen (lips and mouth) and others cannot. This lack of information can be compensated using a proper language model, as using a simple language model increase the results of SD2 by 16%, and SI by 20%.

This major challenge (the lack of information in the visual domain), along with some others, opens the door for more research in the future.

# Chapter 7

# Visual Word Recognition - Applications

Techniques developed over the last two chapters concerning visual word recognition schemes and future improved versions can be applied in many areas. Here we focus on three possible areas where VW system can be applied effectively:

1. Speaker identification: training the computer on some speakers' Vwords, considering one of those speakers, and giving only his/her visual word signal; is it possible to identify him/her?

2. Speaker verification: this application can be called "*Visual Passwords*", training a secure system on some examples of a particular Vword of a particular speaker, giving only his/her visual word signal (the same in the training set in the system); is it possible to log him/her in?

3. Lip-reading security surveillance system: train a surveillance system on some Vwords of interest, like some words related to security issues such as {"Bomb", "Kill, "Run", "Gun", "Fire"}, and some other non-related Vwords; is it possible for the system to alarm when it is given a visual word signal, if the given word is related to the words of interest?

In the rest of this chapter we conduct experiments to describe the VW-based solution and present the results of the experiments conducted in each case to test performance.

## 7.1 Speaker identification

The results from the previous chapter (6) show that VSR is a speaker-dependent problem, since the speaker-independent word recognition rate is about 32% on average, i.e. the word error rate is about 68% on average, which means that the same word spoken by two different speakers is more likely (68%) to produce two different signatures. This difference varies between people and depends mainly on the individual differences between them, such as the way they speak (speech style and behaviour), and the different appearance of the lips and mouth region.

These individual differences form something like "speechprint", which can be beneficial in other applications such as speaker recognition, which is the process of recognizing who is speaking on the basis of his/her (unique to some extent) individual information embedded in his/her speech signal. Speaker recognition can be either speaker identification or verification. Speaker identification identifies a specific speaker (given a specific utterance) from a set of priori known speakers, while speaker verification admits or denies the identity claim of a speaker. A speechprint allows the system to identify the speaker based on his/her speech, after training the system on the way he/she speaks.

Having been based on an holistic approach, where most of the speaker's information and the word information are preserved in each word signature, the proposed VW system can be used to back up speaker recognition systems. To measure the feasibility of this claim, several experiments have been conducted. Using the available 26 different subjects in the in-house database, the system should be trained on some examples of each subject to identify any of them. The set features of all the subjects are constructed using only one word (five examples of each), and the class of this word is the subject (the person him/herself). The leave-one-out cross validation is then used to identify the class of each word, which represents the speaker that the algorithm is trying to identify.

These experiments used the words ("Bomb" and "Two") that are special in the sense that they are the best-recognized words (see Figure 6.2). We might need to find some specific word or words, which are more suitable for identifying each speaker in future work, but in the meanwhile we will stick with these two words in our experiments.

The speaker identification experiment is similar to the speaker-dependent experiment. After collecting word samples ("Bomb" or "Two") from the database, and replacing each word's class by its speaker ID, leave-one-word-out cross validation is then used to identify the speaker of each word (see Table 7.1).

Again, female speakers in different experiments proved to be better than males for lip-reading speaker identification, where accuracy was 94% and 90% for females and males respectively, for the same reasons discussed earlier in chapter 6. Also, the appearance of facial hair proves again to be problematic for VSR, hence the lowest 3 results in the first column (40%, 60% and 60%) belong to males with facial hair (Male-17, Male-3 and Male-7 respectively).

**Table 7.1**. Speaker identification experiment results using WKNN, and either DTW or ED as distance measures with linear interpolation, training either "Bomb" or "Two".

|  | "Bomb" | | "Two" | |
|---|---|---|---|---|
| Subject | DTW | ED | DTW | ED |
| Female_01 | 100% | 100% | 100% | 100% |
| Female_02 | 100% | 70% | 100% | 80% |
| Female_03 | 100% | 100% | 80% | 100% |
| Female_04 | 60% | 50% | 60% | 60% |
| Female_05 | 100% | 90% | 100% | 80% |
| Female_06 | 80% | 90% | 80% | 100% |
| Female_07 | 100% | 80% | 100% | 100% |
| Female_08 | 100% | 100% | 100% | 100% |
| Female_09 | 100% | 70% | 100% | 100% |
| Female_10 | 100% | 100% | 100% | 100% |
| Male_01 | 100% | 60% | 100% | 100% |
| Male_02 | 100% | 100% | 100% | 100% |
| Male_03 | 60% | 40% | 100% | 100% |
| Male_04 | 80% | 90% | 100% | 100% |
| Male_05 | 100% | 90% | 100% | 80% |
| Male_06 | 100% | 100% | 100% | 100% |
| Male_07 | 60% | 30% | 100% | 100% |
| Male_08 | 100% | 100% | 100% | 100% |
| Male_09 | 100% | 100% | 100% | 100% |
| Male_10 | 100% | 80% | 100% | 100% |
| Male_11 | 100% | 100% | 60% | 40% |
| Male_13 | 100% | 90% | 100% | 100% |
| Male_14 | 100% | 100% | 100% | 100% |
| Male_15 | 100% | 80% | 80% | 80% |
| Male_16 | 100% | 100% | 100% | 100% |
| Male_17 | 40% | 0% | 80% | 80% |
| **Average** | **92%** | **81%** | **94%** | **92%** |

The excellent identification rate in these experiments (92% and 94% using "Bomb" and "Two" respectively) mainly comes from the variety of people's appearance and way of speech. By different ways of speech, in this study, is meant not the output audio signal (normal speech), but the different visual appearance (Vwords signals) in producing speech; each person has his/her own mouth movements, circulations, opening, closing and appearance, to produce the same speech that can be produced by another person, but with different VW signals (speechprint).

These individual differences, visually, in the way people speak, introduces the possibility of using the proposed visual words as a paradigm for a visual passwords approach to increase security and prevent "fraud".

## 7.2 Visual passwords (speaker verification)

One application for the proposed visual words approach is to use it for the visual passwords problem, or call it pass-visual-words. For security reasons over the Internet, to access a database or to log into a secure system, string passwords are normally used, consisting of numbers, alphabetic letters and special characters. These are easy to access fraudulently, by trying some or all-possible combinations using many different techniques.

Visual passwords, which can be generated using the proposed Vwords, can be used as a backup for the normal string passwords, or as an alternative to face verification, as the latter can be faked by putting a fake image in front of a camera. For more security, visual passwords can be obtained for a system by asking the user to say a specific word several times. This word can be chosen by the user or provided by the system. Even if someone hacks into the string password, he/she will not be able to produce a visual word in the same manner that the real user did because of individual differences in mouth appearance, and the different way people speak and speak visually. Thus, the security of that system should be increased.

The visual passwords problem can be thought of as a speaker verification problem, unlike the speaker identification problem in which the computer a priori knows all the speakers and uses some training examples from their visual words to identify them. In visual passwords (or speaker verification) the examples of only one speaker (user) is known previously, and the computer has to authenticate the speaker of a visual word approximately equal to one of those in the training set (system database). A specific threshold needs to be defined to approximate the test words, i.e. if the distance between the test word and the visual password is less than a threshold, then access to the system is granted, otherwise access to the system is not granted.

There are 2 scenarios to be considered here; the first being if the impostor knows the password and wants to produce it visually to access the system. The other is when the impostor does not know what the visual word is, so he/she tries to say any word to access the system.

To test the visual passwords in this study, five subjects were randomly selected, assuming that the client is one of them and the others are 4 impostors. The word "zero" spoken by subject 1 was assumed to be the visual password. To define the suitable threshold, the visual password ("zero") spoken by subject 1 from session 2, repeated 5 times, was used as a training set, while the test set was the "zeros" spoken by the five subjects from session 1.

After defining the threshold, the system was evaluated using a test set of the same subject, but this time from session 2 instead of session 1, and the training set was the "zeros" spoken by the client (subject 1), but this time from session 1 instead of session 2 to avoid bias, thus training the threshold in one session, and testing the method in the other session using the best threshold.

If the distance (D) of the nearest neighbour (NN) is less than the threshold (T), then the claimer is granted access to the system, otherwise, the claimer is given several tries to log in, and if he/she fails after n tries, he/she is considered an impostor, and the system will then block him/her (see Figure 7.1).

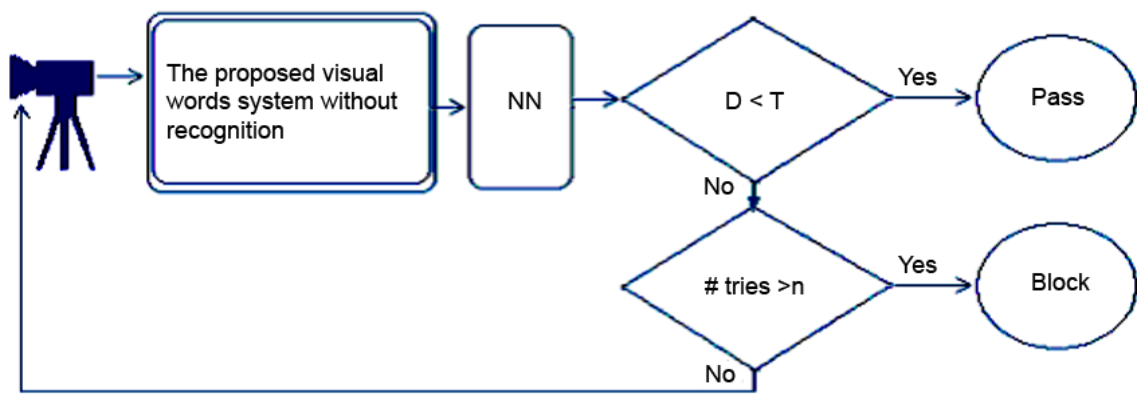

**Figure 7.1**. The proposed visual passwords system, if the distance to the nearest neighbour was less than a specific threshold then pass, otherwise try again n times then block.

To find the best threshold to minimize errors in the system, the visual password experiment was repeated several times, starting with threshold = 1 then increasing it by 0.1, until reaching 5. There are 2 types of errors in such experiments, the false rejection error (FRE) and the false acceptance error (FAE).

The FRE occurs when the system rejects the client considering him/her as an impostor, and the FAE occurs when the system accepts an impostor as a client. Both the false rejection ratio (FRR) and the false acceptance ratio (FAR) were recorded each time the threshold was increased (see Figure 7.2). The best threshold is the one where the total of FAR and FRR is minimized.

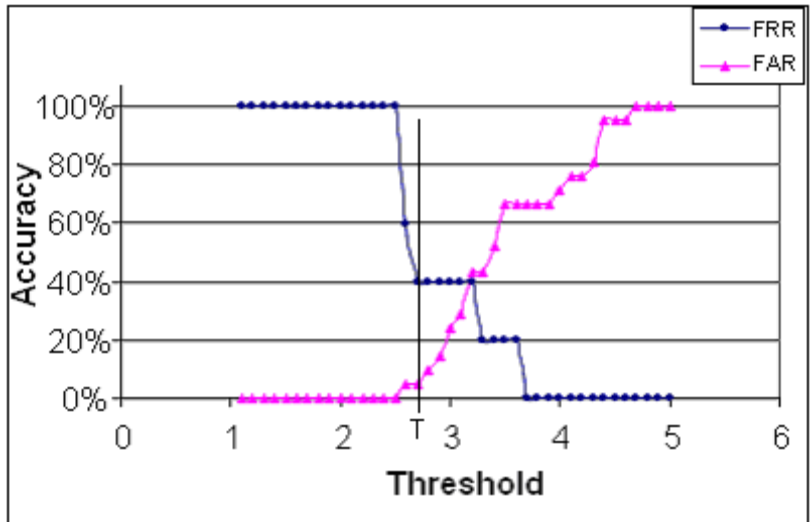

**Figure 7.2**. Threshold choosing for the visual password "zero".

Figure 7.3 depicts the system performance in accepting and rejecting the word "zero", assuming that the impostor knows the password. The FRR curve in Figure 7.2 and the error curve in Figure 7.3 both look like a step function, because there are only 5 examples of the visual password "zero" in the training set. The number of rejected words can be either 0, 1, 2, 3, 4 or 5, giving FRR values as 0, 20%, 40%, 60%, 80% or 100% respectively. Obviously, increasing the number of visual words in the training set for each client improves the performance of the system.

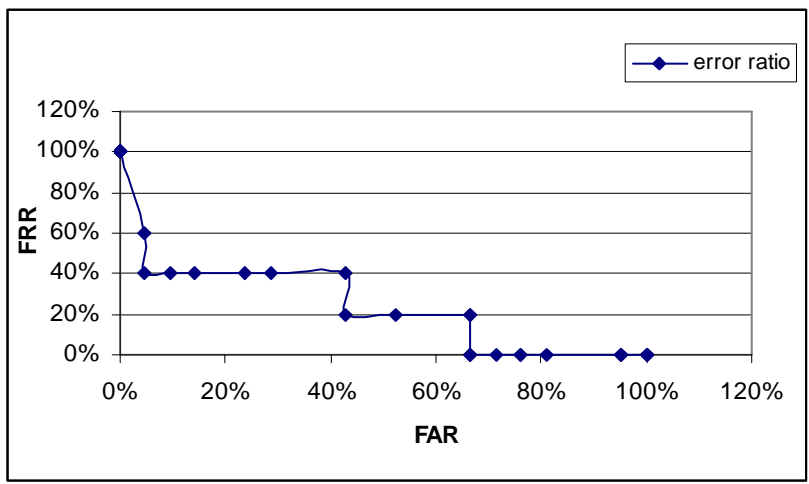

**Figure 7.3**. FRR in terms of FAR for visual passwords system - "zero" is the visual password..

The threshold that minimizes the errors in the previous experiment is (2.7). Table 7.2 shows the results of the visual passwords system, using the threshold (2.7). The experiment was repeated twice, in the first scenario, assuming that the impostors know the password, so the test set contains only the word "zero" spoken by 5 subjects (1 client who is subject 1 and 4 impostors).

In the second scenario, the experiment assumes that the impostors do not know the password, so the test set contains 5 different words (non-zero Vwords: *"Three"*, *"Hold"*, *"Kite"*, *"Situation"*, *"Run"*), which were spoken by 4 subjects (the impostors trying to log into the system), in addition to 5 examples of the word "zero", which were spoken by subject 1 (the client) from session 2.

**Table 7.2**. Visual passwords results, using "zero" as the visual password.

| Experiment scenario | Threshold | Pass | No Pass | FRR | FAR | Error average |
|---|---|---|---|---|---|---|
| Scenario 1 | 2.7 | 60% | 70% | 40% | 30% | 35% |
| Scenario 2 | 2.7 | 60% | **97.98%** | 40% | **2.02%** | 21.01% |
| | Average | 60% | 84% | 40% | 16% | 28% |

So the word "zero" of the real client (subject 1) has 10 examples, 5 from session 1, and 5 from session 2. When finding the threshold, the 5 examples from one session were used as a training set, and the other 5 from the other session were used in the test set. In the evaluation process, the 5 examples used in the test set became a training set, and the 5 examples used as a training set joined the test set with the other words spoken by the impostors.

It can be noticed from Table 7.2 the improvement in the performance of the visual passwords, particularly the reduction of the false acceptance error FAR in scenario 2. Hence the number of unknown different visual passwords was trying to gain access to the system, while the FRR remained the same because the number of the word "zero" spoken by the client has not been changed.

The threshold obtained in the previous experiments was chosen on the basis of minimizing both FRR and FAR. However, some systems have different needs, and these needs reflect the amount and type of errors. For example, take the visual passwords system; it is important not to accept words other than the stored visual password, so FAR should be minimized as much as possible, regardless of the FRR. In this case, such systems use a weighted error rate instead, where the FRR has a different weight from that of the FAR, and the system uses the threshold that minimizes the weighted error rate, which can be defined as:

$$WeightedErrorRate = \frac{\omega FAR + FRR}{\omega + 1}$$

where $\omega$ is the weight of FAR. If $\omega$ was 2 for instance, then the threshold used in the previous experiment would be less than 2.7, allowing less words to pass, but including some real examples, because some of these examples will be rejected by reducing the threshold. Nevertheless, the rejection is justified by these systems, considering security issues, as security for them comes first.

The performance of the proposed visual passwords system can be improved by considering more than one visual word in the training set. This will increase the length of the visual password, and provide a stronger signal, which reduces the probability of being hacked. Using the same protocol as for the visual passwords experiments and the same five subjects: for each speaker (client) all the possible combinations of the words "four" and "five" from session 1 have been concatenated to create a new signal, each word being repeated 5 times, so the number of concatenated words is 25 new signals (double Vwords), which is the size of the training set for each speaker, when evaluating one speaker as a client and the other four speakers as impostors. Bootstrapping is used in this experiment to produce a larger number of samples by concatenating 2 words with all their possible combinations.

According to Fox (2002) "*Bootstrapping is a general approach to statistical inference based on building a sampling distribution for a statistic by re-sampling from the data at hand*". Hence the data at hand is relatively small; it is useful to use Bootstrapping, also the way of saying the word "A" might be different from the way of saying it next time, so mixing all these different ways of saying the word "A" with all these different ways of saying the word "B", for instance, enriches the training set by covering more possibilities, in addition to a better validation when using an affluent test set.

The test set contains all the possible combinations of 2 different words (the words "six" and "seven") spoken by the rest of the speakers in addition to all the possible combinations of the words "four" and "five" spoken by the client in session 2.

To find the best threshold for each speaker, data from session 1 was tested with training data from session 2. As in the previous visual password experiments, the best threshold is the one that minimizes both FRR and FAR (see Figure 7.4).

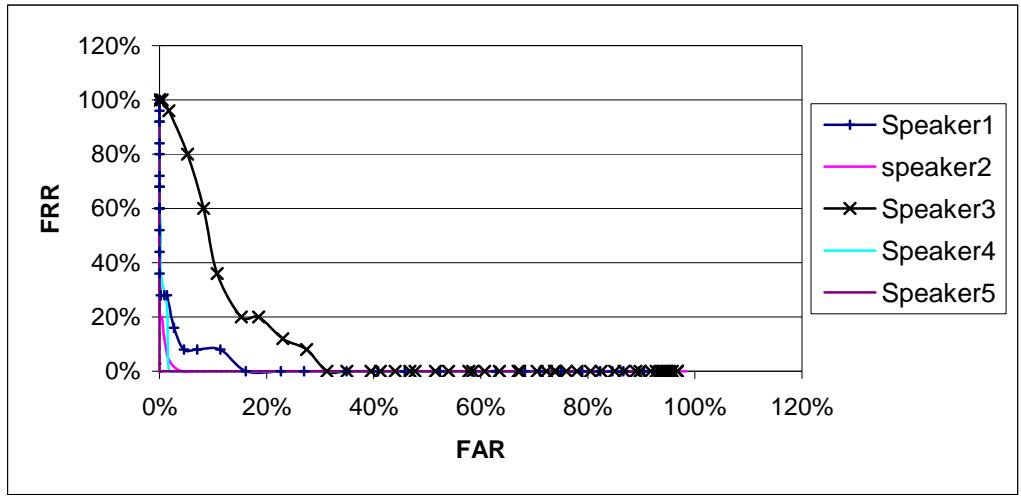

**Figure 7.4**. Visual passwords system error rates using combinations of "four" and "five" Vwords as visual passwords. FRR as a function of FAR.

After defining the best threshold, data from session 2 were used for testing, and the training set data came from session 1. This experiment was repeated 5 times for each speaker, each time the speaker being evaluated being considered as a client, by training his new Vwords (gained from the concatenation of "four" and "five"), and the other speakers being considered as impostors, trying to get into the system by providing their Vwords (gained from the concatenation of "six" and "seven"). The results are shown in Table 7.3.

**Table 7.3**. Visual passwords system results using combination of "four" and "five" Vwords as a visual password.

| Subject | Threshold | Pass | No Pass | FRR | FAR | Error average |
|---|---|---|---|---|---|---|
| Speaker 1 | 6.2 | 100% | 58.75% | 0.0% | 41.3% | 20.63% |
| Speaker 2 | 4.8 | 100% | 87.00% | 0.0% | 13.0% | 6.50% |
| Speaker 3 | 4.4 | 100% | 94.78% | 0.0% | 5.2% | 2.61% |
| Speaker 4 | 4.3 | 80% | 100.00% | 20.0% | 0.0% | 10.00% |
| Speaker 5 | 4.1 | 100% | 99.50% | 0.0% | 0.5% | 0.25% |
| | Average | 96% | 88% | 4.0% | 12.0% | **8.00%** |

It can be noticed from Table 7.3 that the performance of the system has improved by using more than one word in the training set. The average error is only 8%, which is much less (28%) than that in Table 7.2 (where only one Vword was used as a visual password). Despite speakers each having their own unique ways of speaking (visual and audio), they still have something in common – allowing some people to defraud/trick the system to some extent (see Tables 7.2 and 7.3). So, the more the client has a matchless, individual way of speaking visually and is consistent, the less chance there is of impostors defrauding his visual password (such as Speaker 5 and in contrast with Speaker 1).

The performance of the visual passwords system can be further improved significantly by combining other biometric factors, such as face authentication or even combining or fusing a PIN number, or the string password with the visual word signature. The fusion can be done by concatenating the feature vectors, score-based fusion, or as a decision-based fusing, i.e. each system works alone, and before the computer takes the final decision it considers the output of each system alone, and uses a specific voting or weighting technique to get to the final decision. There can also be a multi-level approach, if the user gets the string password right, then the system moves him to another check, using the visual passwords for instance, and so on. All these improvements will be investigated in future work.

## 7.3 Lip-reading security surveillance system

The last possible application of Vwords to discuss in this thesis is the security surveillance system. This system can monitor sensitive places such as airports for instance, to see if people say some specific words that may lead to security issues. In the new database, 5 words *("Bomb", "Kill, "Run", "Gun", "Fire")* were chosen as examples of such words. The first type of experiment to evaluate such a system was conducted by training the security words with some non-security words, and testing all the words using leave-one-out cross validation. For each person in the database, a file of all the security words (5 different words) and 5 non-security different words was constructed, so that each file contained 10 different words repeated 5 times, and spoken by a specific person.

Cross-validation technique was used to evaluate the recognition of the security words; by leaving one file out and training the other files, the evaluation was based on classifying the test word whether it was a security word or not. This is considered as a speaker-independent experiment, because all the examples of the tested subject are outside the training set. Thus it can be used in surveillance cameras to alarm the system if someone says such words from a distance. The experiment was repeated 3 times, each time with a different word group alongside the security group:

1- Security group (Sec) and look-alike $1^{st}$ group (LAL1)

2- Security group (Sec) and look-alike $1^{st}$ group (LAL2)

3- Security group (Sec) and long words group (LG).

**Table 7.4**. Word recognition rate by training both security and non-security words, using WKNN, ED and linear interpolation.

| Subject | LAL1 | LAL2 | LG | Filename | LAL1 | LAL2 | LG |
|---|---|---|---|---|---|---|---|
| Female_01 | 82% | 74% | 78% | Male_04 | 84% | 58% | 78% |
| Female_02 | 82% | 46% | 82% | Male_05 | 88% | 68% | 78% |
| Female_03 | 80% | 84% | 96% | Male_06 | 84% | 62% | 70% |
| Female_04 | 82% | 56% | 90% | Male_07 | 48% | 69% | 68% |
| Female_05 | 68% | 72% | 98% | Male_08 | 78% | 66% | 64% |
| Female_06 | 82% | 62% | 88% | Male_09 | 86% | 90% | 96% |
| Female_07 | 96% | 42% | 74% | Male_10 | 68% | 46% | 80% |
| Female_08 | 70% | 76% | 68% | Male_11 | 94% | 86% | 88% |
| Female_09 | 92% | 80% | 82% | Male_13 | 66% | 76% | 86% |
| Female_10 | 82% | 74% | 84% | Male_14 | 52% | 52% | 76% |
| Male_01 | 70% | 74% | 90% | Male_15 | 82% | 66% | 74% |
| Male_02 | 78% | 62% | 74% | Male_16 | 76% | 96% | 76% |
| Male_03 | 70% | 58% | 72% | Male_17 | 59% | 72% | 72% |
|  |  |  |  | Average | 77% | 68% | 80% |

As can be noted from Table 7.4, the highest recognition rates are when comparing the security words with the long words (average of 80%). Obviously, the difference in the signal lengths between LG and Sec causes Vwords from both groups to be very dissimilar. The lowest recognition rate for the security words results was when comparing Sec with LAL2, hence these two groups share some similarity, for example the words "fire" and "fight".

Security words in this database have nothing in common, i.e. they have no syntax, visual features, many phonemes, nor many visemes in common. They are just different words having a specific meaning, related to some issue, in this case security.

The previous experiments cannot be applied as a lip-reading surveillance system, since it is difficult to train the system on all possible words as security and non-security words. An alternative experiment to test a lip-reading surveillance system can be as follows:

1- Train the system on the words of interest (in this case the words related to security issues).

2- Take a word at random from the test set (it can be either a security word or a non-security word).

3- Use a specific recogniser (say KNN) to get the nearest word in the training set or all the nearest k words.

4- If the distance of the predicted word (or average distances if K>1) is less than a specific threshold, then recognize the test word as a security word; otherwise it is a non-security word.

The problem is: how to decide the threshold?

To avoid bias, the data used to define the threshold should not be used to evaluate the system. In this experiment, the database was divided into 2 sets, the odd-numbered videos and the even-numbered videos.

Odd-numbered videos={Female_01, Female_03,..., Male_01, Male_03,…, Male_17}
Even-numbered videos={Female_02, Female_04,…, Male_02, Male_04,…, Male_16}

The idea is to use one set to find a threshold, and the other set to evaluate the surveillance system. To avoid bias, each set should be tested using the other set's security words as a training set. Therefore, the odd-numbered videos set was used to define the threshold, by testing all the words (security and non-security words) against a training set consisting of all the security words from the even-numbered videos set, using leave-one-out cross validation.

The best threshold is the one that minimizes the errors in the system. In this case, there are two types of errors: false rejection (FR) and false acceptance (FA). The FR errors occur when the system tests a security word and recognizes it as a non-security word, and the FA errors occur when the system recognizes a non-security word as a security word. The next figure (7.5) shows the best threshold defined by minimizing both false rejection and false acceptance rates (FRR and FAR) in the system using the previous experiment.

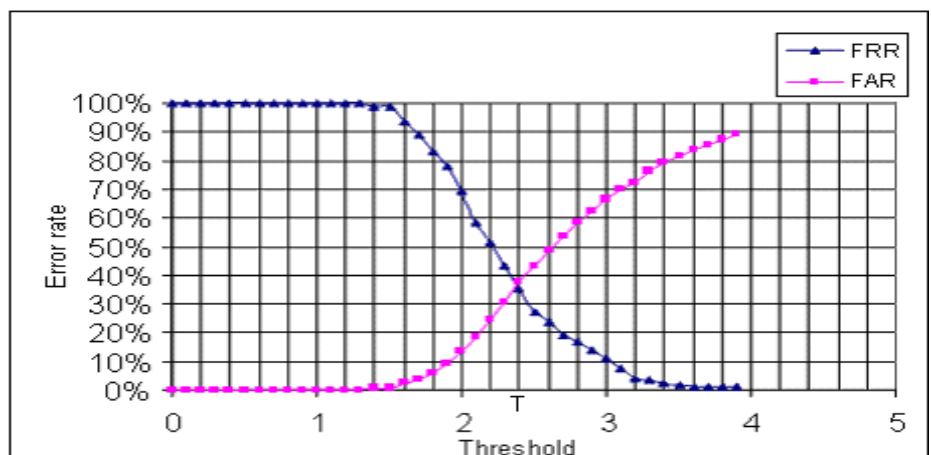

**Figure 7.5**. Average of FRR and FAR of all the test subjects used to define the threshold.

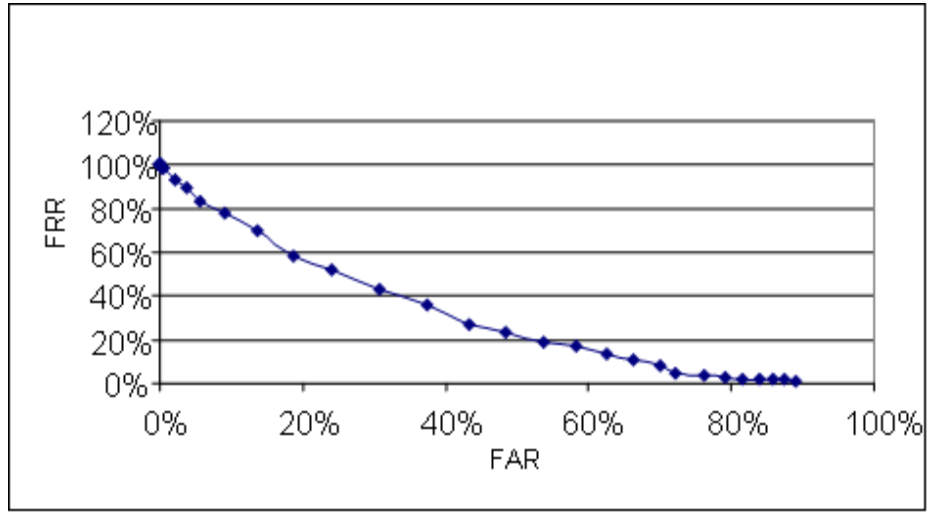

**Figure 7.6**. Lip-reading security surveillance system error rates. FRR as a function of FAR.

It can be seen from Figure 7.5 that the best threshold is 2.4. Using this threshold, the even-numbered videos set is tested with the training set of the security words of the odd-numbered videos. If the nearest distance is less than the threshold (2.4) then the word is predicted as a security word, and the system will be alarmed; otherwise it will be considered as a non-security word. The next table (7.5) illustrates the results of this experiment.

**Table 7.5**. Lip-reading security surveillance system experiments by training security words from odd videos, using WKNN, ED, linear interpolation and threshold = 2.4.

| Subject | Security | Non-security | FRR | FAR | Error Average |
|---|---|---|---|---|---|
| Female_02 | 4% | 83.20% | 96% | 17% | 56% |
| Female_04 | 56% | 61.60% | 44% | 38% | 41% |
| Female_06 | 48% | 70.16% | 52% | 30% | 41% |
| Female_08 | 100% | 17.60% | 0% | 82% | 41% |
| Female_10 | 100% | 50.40% | 0% | 50% | 25% |
| Male_02 | 92% | 48.80% | 8% | 51% | 30% |
| Male_04 | 96% | 40.80% | 4% | 59% | 32% |
| Male_06 | 80% | 39.20% | 20% | 61% | 40% |
| Male_08 | 100% | 16% | 0% | 84% | 42% |
| Male_10 | 28% | 62.40% | 72% | 38% | 55% |
| Male_14 | 68% | 73.60% | 32% | 26% | 29% |
| Male_16 | 60% | 28.80% | 40% | 71% | 56% |
| | | | | average | 41% |

The average of both error types is 41%, which means that the system performance has a 59% security alarm/no-alarm rate. However, these results can be changed if the data or the threshold have been changed. Reducing the threshold in this system increases the FRR, allowing more words to pass the system as non-security words, leaving the sensitivity of the system low to security words. On the other hand, increasing the threshold reduces the FRR, making the system more sensitive to security words, but at the same time it increases the FAR, allowing more non-security words to alarm the system as security words. The weighted error rate can be used to adjust such systems as needed.

The results of the visual security words experiment were not as good as expected, because the experiment used training data selected from subjects (the odd videos) different from the subjects of the test data (the even videos). This experiment thus inherited the problems of the speaker-independent approach, where each speaker has his/her own way of speaking and his/her own mouth appearance. Consequently, each speaker defines his/her own threshold, i.e. for each speaker the error is minimized when using a local threshold. However, this is difficult to maintain in the real life application.

If the previous experiment had been repeated using a local threshold (the best threshold that minimizes the errors of a specific speaker), the performance of the system would be improved, but these results would be **biased** results, because in such a surveillance system, we are looking for a threshold that is common to all the speakers. Alternatively, if the extracted features of the visual words were less image-based features and more geometric-based features, the results might then be improved in such a system.

The global threshold which was used in the previous experiment was obtained by minimizing the averages of all the FRR and the averages of all the FAR for all the test speakers. The proposed visual security words surveillance system might look like the one in Figure 7.7.

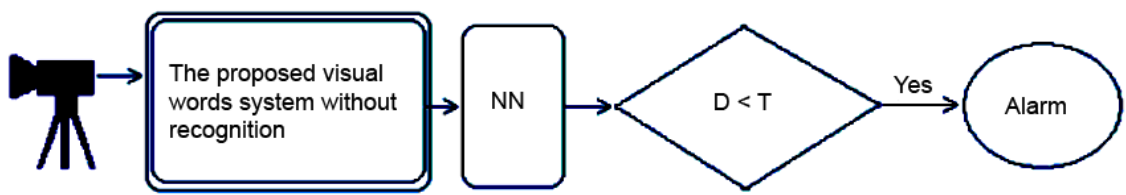

**Figure 7.7**. The proposed lip-reading security surveillance system, where D is the nearest distance of the tested word, with the trained security words, T is the threshold, and NN is the nearest neighbour algorithm.

## 7.4 Summary

This chapter discussed applying the proposed VW approach to some possible real life problems such as speaker identification, speaker verification and lip-reading security surveillance problems. Several experiments were conducted to evaluate the VW approach as a solution for these problems. These experiments were:

1. Speaker identification experiments, using the VW approach, these experiments revealed 92% and 94% speaker identification rate using the words "bomb" and "two" respectively to represent each speaker.

2. Speaker verification (authentication), The visual passwords system was proposed using the VW approach. The average error rate for the conducted experiments was 28% using only one word as a visual password, reduced to 8% by concatenating 2 visual words.

3. A lip-reading security surveillance system was proposed using the Vwords approach; the accuracy of the system (by training both security and non-security Vwords) was 77%, 68% and 80% by training LAL1, LAL2 and LG respectively. The average error rate for the system by training only the security words was about 41%.

The results of these experiments emphasize the superiority of the visual words approach as a solution for the visual speech recognition problem and its applications, and open the door for more applications and research to come.

# Chapter 8

# Conclusion and Perspectives

This thesis investigated the challenging task of automatic lip reading, also referred to as visual speech recognition (VSR). A multi-stage VSR system is proposed using the visual words (Vwords) approach. Each stage of the proposed approach addresses a particular difficulty faced by an automatic VSR system. The main stages of the proposed approach include face localization, lip localization and lip reading. A summary of each stage of the proposed method is given here. The thesis will conclude with a description of future work.

## 8.1 Face localization

The proposed face localization method, which was introduced in chapter 3, is a hybrid of the knowledge-based approach, template matching approach and feature invariant approach (skin colour). This method uses a wavelet transform to decrease the time needed for various scanning steps compared to scanning the spatial domain. The colour information was not affected much by the lighting conditions, because it was used only to choose the best face location from the best 5 candidates' locations using fuzzy logic.

It was demonstrated in chapter 3 that the proposed scheme has a high accuracy rate (93.40%), and significantly outperforms the Rowley's et al. (1998) method (85.93%). At least two PhD candidates in the University of Buckingham used this method to localize faces in the PDA database for their research projects.

Although the proposed method was designed to work only under certain assumptions, it is possible to modify it to work under any other different set of assumptions. Moreover, the method is efficient with modest time-complexity ($O(n^2)$), easy to implement even on constrained devices such as mobile phones and PDAs, and robust against the light conditions.

## 8.2 Lip localization

This study proposes two new methods for lip localization. One is called the "layer fusion", which includes votes from different well-known approaches indicating lip and non-lip pixels, but it is not very accurate, and it does take a long time. The other method is called the "nearest colour" lips localization method (see chapter 4).

The later method depends on colour information, because of the colour of the lips being different from the colour of the face. This method uses one of the colour-based lip-detection methods from the literature, which partially segments the lips. Yet to get the whole lip area, the "nearest colour" uses other information such as r, g, b, warped hue, Cb, Cr, the X co-ordinate and the Y co-ordinate of the segmented lip pixels. When comparing the unclassified pixels in ROI with this information, the nearest pixels are classified as lip pixels.

The results show that the "nearest colour" algorithm accuracy (91.15%) is much better than that of state-of-the-art lip detection methods, such as ASM (64.22%), "Hybrid Edge" (68.65%) and some other lip detection colour-based approaches.

The "nearest colour" algorithm can be applied in real time and online applications, and it is also robust against the appearance of the beard and the moustache (see Table 4.2). Therefore it was used to detect lips after the face localization stage in the proposed VSR system.

## 8.3 Lip reading

The proposed "*visual words*" (Vwords) scheme uses an holistic approach to tackle the VSR problem, where the system recognizes the whole word rather than just parts of it (visemes). In this approach, a word is represented by a signature that consists of several signals or feature vectors (or feature matrix). Each signal is constructed by temporal measurements of its associated feature. The mouth height feature, for instance, is measured over the time period of a spoken word.

Several experiments were discussed in chapter 6 to evaluate the Vwords approach to the automatic lip-reading problem. These experiments were:

1. Speaker-dependent experiments; we conducted four types of this experiment:

    - Leave-one-example-out cross validation using the same video samples in both test and training sets (see Table 6.5).

- In the in-house database we have two groups of words, where words in each group have nearly the same signature (the look alike words). In this type of experiment we consider them as one word for each group, as any word can be used to recognize the others (see Table 6.6, 3).

- Another experiment, which is called SD2, trains a subject's video from session 2 and tests his/her video from session 1 (see Table 6.7).

- The last one is the same as SD2 but using a language model simulation (see Table 6.7).

Several reasons were found that affected the various results such as the appearance of facial hair, and the individual's aptitude to produce a clear visual signal. Some subjects produce weak signals (termed as visual speechless persons (VSP)), and the word recognition rate is greatly affected by its phonemes' components, because producing these sounds involves the movement (dynamic) of some parts of the speech production system.

2. Speaker-independent experiments: the performance of SI experiments was less than that of SD. Individual differences in the mouth appearance, and in the way they talk, produces different visual signals for the same spoken word, which emphasizes that the visual speech recognition problem is a speaker-dependent problem.

3. Vwords can be applied in several applications effectively; for example, speaker identification experiments revealed 92% and 94% speaker identification rates using the words "bomb" and "two" respectively.

4. A visual passwords application (speaker verification) was proposed using the Vwords approach. The average error rates for the experiments conducted were 28% using only one word as a visual password, and 8% by concatenating 2 visual words.

5. A lip-reading surveillance system was proposed using the Vwords approach; the accuracy of the system (by training both security and non-security Vwords) was 77%, 68% and 80% by training LAL1, LAL2 and LG respectively. The average error rate for the system by training only the security words was about 41%.

6. The look-alike experiments were conducted to prove that it is not necessary to train all the English language words to obtain an automatic lip-reading system. The experiments' excellent average (94%) shows that it is enough to train one or some of the similar words, rather than training all the words.

7. The visual words approach was compared with the visemic approach, which was applied using HMM. The experiment results show that the visual words approach significantly outperforms the visemic approach to the visual speech recognition problem, in both types of experiments (speaker-dependent and independent).

By analysing all these experiments and investigating the VSR problem, we could draw several conclusions:

1. Using the in-house database, the results of adapting the new visual words approach are much better than those for the traditional visemic approach, the low results of the visemic approach experiments being due to the problems associated with the visemes, particularly the ones related to detecting the visemes.

2. The lip-reading problem is a speaker-dependent problem. Some speakers can be easily read, while for some others it becomes quite a difficult problem. This claim can be backed up by Jun and Hua (2009), who reported a large difference in the word recognition rate between the speaker-dependent and speaker-independent, 65.9% and 23% respectively.

3. Some people do not fully or sufficiently produce the visual signals while speaking. This study termed them as visual speechless people (VSP).

4. Using the Vwords approach, it is possible to recognize some targeted words (security words) from other normal words using some kind of security surveillance system.

5. Speakers were easily identified using the Vwords approach, as they have a different (unique) way of visual speech.

6. Using the Visual passwords scheme, a system's security can be increased significantly, particularly when using more than one Vword as a visual password.

7. The VWR performance was greatly affected by the appearance of facial hair because facial hair appearance varies from one person to another, and even for the same person from time to time. This produces different features for the same spoken word.

8. The female lip-reading results are better than those for males, due to their highly distinguished mouth areas, and the absence of beard and moustache in the female case.

9. The proposed Weighted KNN is a better classifier than KNN, because the first takes into consideration the distances of all the k neighbours, and uses these distances as weights for each neighbour.

10. Using DTW as a distance function is slightly better than Euclidean distance with linear interpolation, However, Euclidean distance ($O(n)$) is much faster than DTW ($O(n^2)$).

11. The best k in the KNN is different from one speaker to another. Nevertheless, the majority of good results were recorded when k = 1.

Despite the many problems associated with the viseme model, it has a great advantage over the visual words model, which is generalizability, i.e. it is enough to train a specific number of visemes to be able to recognize different words. While the major problem that faces the proposed visual words approach, is that for a complete English language lip-reading system we need to train the whole English language words in the dictionary!

To overcome this problem we propose the following solutions:

1. English and other languages have a similar Vwords signature. It is adequate to train one of these words to recognize its group (*between groups recognition*), and then use other techniques such as syntax and semantic rules, or even some other features, to classify the spoken word within its group (*within group* *recognition*). This solution was put to the test and revealed promising results for the "*between group recognition*" (see chapter 6 section 8 and Table 6.14), leaving "*the within group recognition*" for future work.

2. Visual words can be thought of as an application-specific scheme, for instance, to be used in visual passwords, or in security surveillance systems, or even for deaf people to recognize the most common words in their daily life (see the security experiments, chapter 6 section 7.3). These domains have a limited number of words, which can ease the training process for the Vwords approach.

3. Using phonetic alphabets, where each character or phoneme has its own complete word to differentiate it from other characters or phonemes. For example, "A" is pronounced as Alpha, "B" as Bravo, …, "Y" as Yankee, and "Z" as Zulu. And for the numbers "0" might be pronounced as Zero, "1" as one, … and "9" as Nine. e.g. for "29 Nortons Place" speaker should say: Two, Nine, November, Oscar, Romeo, Tango, Oscar, November, Sierra, Papa, Lima, Alpha, Charlie, Echo). This process take longer time but solve the previous problem and should increase accuracy if such words where chosen carefully to guarantee variety, i.e. each word has its own different visual signature. Several experiments will be conducted in the future work to evaluate such an approach.

Another problem faces VSR in general is the lack of information that comes from the visual side of speech; about 50% or less of English language can be seen. This problem can be dealt with by adding extra information such as a language model (see Table 6.7). The overall results of the SD and SI experiments were increased by 16% and 20% respectively using just one rule, which was used for simulation reasons only. Furthermore, the phonetic alphabets approach can be used to compensate for the lack of information.

## 8.4 Possible future directions

Despite the relatively good results of the proposed solutions in this study, they are still not perfect, and there is room for improvement in future for all the investigated areas, which include face and lip localization and lip reading.

**8.4.1 Future work on face localization**

The accuracy of an automatic lip-reading system depends mainly on the accuracy of the face detection/localization method. None of the existing face detection/localization methods, nor the proposed one, are perfect solutions.

We are intending to work further on this area, attempting to deal with the following:

1. To upgrade the proposed face localization method to be a face detection method, and to check for more than one face in an image, with different poses and directions.

2. ANDing or ORing the proposed edge/feature detection filter with other edge detection filters such as the Sobel edge detection, which may result in better detection of the facial features.

**8.4.2 Future work on lip localization**

The accuracy of an automatic lip-reading system depends mainly on the accuracy of the lip detection/localization method. To the best of the researcher's knowledge, none of the existing lip detection/localization methods, nor the proposed ones, are perfect solutions in all circumstances. A 91.15% lip localization accuracy is not good enough for a robust lip-reading system. Therefore, the author is intending to work further on this area, trying to increase the accuracy and decrease the processing time of the proposed methods.

**8.4.3 Future work on lip reading**

Although the visual side of English language speech does not provide much information to recognize speech, either by human or machine, and the proposed Vwords outperform the existing approach (visemes), we are still not very satisfied about some of the results.

To increase the performance of the Vwords scheme and its applications, we are intending to work further on these projects to deal with the following:

1. To eliminate the video-specific features, the author will try to find better features to substitute for the image-based features, and to use a pre-processing technique to fix the light problems.

2. Using some kind of language model increased the SD2 accuracy from 45% to 61%. Studying and applying a real English language model with the proposed visual words is vital for increasing the performance, and is one of the author's priorities in the future.

3. Adding visual information to audio information results in better speech recognition (Neti, et al. 2000, Hazen, et al. 2004, Gurban and Thiran, 2005, Arsic and Thiran, 2006, Lucey and Sridharan, 2008). We are intending to add audio information to the proposed Vwords to improve the performance of the speech recognition.

4. Other languages, such as the Arabic language, will be investigated by applying the visual words scheme.

5. The use of other good reputable recognisers such as support vector machines (SVM), will be investigated.

6. The word end points need to be found using only visual information, particularly when there is no audio signal, or in a noisy environment. Audio information has been used in this study for experimental purposes. The visual pattern of the silence viseme can be used to detect such ends.

7. The Vwords approach will be investigated using continuous speech, and a large vocabulary, in a real-life (uncontrolled) environment.

8. The code of the algorithms used in this study needs to be optimised to work effectively in real-time applications and mobile devices (PDAs).

9. Other applications for Vwords will be investigated, such as the talking faces.

10. The performance of the visual password system will be further improved by:

    - Combining other biometric factors, such as face authentication.

    - Combining or fusing a PIN number, or the string password with the visual password's signature.

    - Using a multi-level approach; if the user gets the string password right, then the system moves him to another check, using the visual passwords for instance.

Addressing all or at least some of these problems in the future will enhance the performance of the Vwords approach and lay a strong foundation for further research and the investigation of other innovative applications.

# Publications

Part of the work contained in this thesis is based on the following articles:

Hassanat, A. and Jassim, S. (2008) "A special purpose knowledge-based face localization method", *Proceedings of the SPIE*, Volume 6982, pp. 69820M-69820M-9.

Hassanat, A. and Jassim, S. (2009) "Color-based Lip Localization Method", to appear in *Proceedings of the SPIE 2010.*

Hassanat, A. and Jassim, S. (2009) "Visual words for lip-reading", to appear in *Proceedings of the SPIE 2010.*

# Appendix A
# SAMPA and IPA list of phonemes.*

Table A1. SAMPA simplified list of consonants

| SAMPA | IPA | Description | Examples |
|---|---|---|---|
| **p** | p | voiceless bilabial stop | English *p*en |
| **b** | b | voiced bilabial stop | English *b*ut |
| **t** | t | voiceless alveolar or dental stop | English *t*wo, Spanish *t*oma, Italian fa*t*a |
| **d** | d | voiced alveolar or dental stop | English *d*o, Italian ca*d*e, Spanish an*d*ar |
| **ts** | ʦ | voiceless alveolar affricate | Italian a*zz*urro, pi*zz*a, German *Z*eit |
| **dz** | ʣ | voiced alveolar affricate | Italian *z*io, gra*z*ie |
| **tS** | ʧ | voiceless postalveolar affricate | English *ch*air, pi*c*ture, Spanish mu*ch*o, Italian *c*ena, German Deu*tsch*e |
| **dZ** | ʤ | voiced postalveolar affricate | English *g*in, *j*oy, Italian *gi*orno |
| **c** | c | voiceless palatal stop | Greek [ce] 'and', Hungarian *ty*k 'hen', like British *tu*ne |
| **J** | ɟ | voiced palatal stop | Hungarian e*gy* 'one', like British *du*ne |
| **k** | k | voiceless velar stop | English *c*at, *k*ill, *qu*een |
| **g** | g | voiced velar stop | English *g*o, *g*et |
| **q** | q | voiceless uvular stop | Arabic *q*of |
| **p\\** | φ | voiceless bilabial fricative | Japanese *f*u |
| **B** | β | voiced bilabial fricative | Spanish ca*b*o, cal*v*o (*) |
| **f** | f | voiceless labiodental fricative | English *f*ool, enou*gh*, Spanish and Italian *f*also |
| **v** | v | voiced labiodental fricative | English *v*oice, German *W*elt, Italian *v*edere |
| **T** | θ | voiceless dental fricative | English *th*ing, Castilian Spanish *c*aza |
| **D** | δ | voiced dental fricative | English *th*is, Spanish ca*d*a (*) |
| **s** | s | voiceless alveolar fricative | English *s*ee, pa*ss*, *c*ity, Spanish *s*í, German Gro*ss*, Italian *s*uono |
| **z** | z | voiced alveolar fricative | English *z*oo, ro*s*es, German *S*ee, Spanish rie*s*go, Italian ca*s*a |
| **S** | ʃ | voiceless postalveolar fricative | English *sh*e, *s*ure, emo*t*ion, French *ch*emin, Italian *sc*endo, German *Sp*rache |
| **Z** | ʒ | voiced postalveolar fricative | French *j*our, English plea*s*ure, Argentinian Spanish *ll*uvia, Ecuadorian Spanish a*rr*iba |
| **C** | ç | voiceless palatal fricative | Standard German I*ch*, Greek [Ceri] 'hand', some English pronunciations of *h*uman |
| **j\\ (jj)** | ʝ | voiced palatal fricative | Spanish *y*ate, a*y*uda |
| **x** | x | voiceless velar fricative | Scots lo*ch*, Castilian Spanish a*j*o, German Bu*ch* |
| **G** | γ | voiced velar fricative | Spanish al*g*o, a*gu*a (*) |

---

* www.phon.ucl.ac.uk/home/sampa/index.html and http://www.fact-index.com/s/sa/sampa_chart.html

Table A1. SAMPA simplified list of consonants ~ Continue

| SAMPA | IPA | Description | Examples |
|---|---|---|---|
| X\\ | ħ | voiceless pharyngeal fricative | Arabic *h.* |
| ?\\ | ʕ | voiced pharyngeal fricative | Arabic *'ayn* |
| h | h | voiceless glottal fricative | English *ham*, German *Hand*, Colombian Spanish *jamón* |
| h\\ | ɦ | voiced glottal fricative | Hungarian *lehet*, Some English pronunciations of *aha* |
| m | m | bilabial nasal | English *man*, Spanish *hambre*, Italian *fame* |
| F | ɱ | labiodental nasal | Spanish *infierno*, *enfermo*, Hungarian *kámfor*, *honvágy* |
| n | n | alveolar nasal | English, Spanish and Italian *no* |
| n^ | ɲ | palatal nasal | US English *canyon*, Spanish *año*, French *oignion*, Italian *gnocchi*, Hungarian *anyu* |
| N | ŋ | velar nasal | English *singer*, *ring*, Spanish *blanco*, *manguera*, Italian *bianco*, *pongo*, German *lange* Tagalog *ngay*, *ngong* |
| l | l | alveolar lateral | English *left*, Spanish *largo*, Italian *lungo* |
| L | ʎ | palatal lateral | Italian *aglio*, *famiglia*, Catalan *colla*, Castilian Spanish *cuello* |
| 5 | ɫ | velarized dental lateral | English *milk* (dark l), Catalan *alga* |
| 4 (r) | ɾ | alveolar flap | US English *better*, Spanish *pero*, Italian *essere* |
| r (rr) | r | alveolar trill | Spanish *perro*, *rey*, Italian *arrivare*, *terra* |
| r\\` | ɻ | retroflexed alveolar approximant | English *run*, *very* |
| R | ʀ | uvular trill | French *rue*, standard German *Reich*, *Farb* |
| w | w | rounded back semivowel | English *we*, Frech *oui*, Spanish *hueso*, Italian *acqua*, *suono* |
| H | ɥ | rounded front semivowel | French *huit* |
| j | j | unrounded front semivowel | English *yes*, Frech *yeux*, German *ja*, Italian *occhio*, *piove*, Spanish *pierna* |

Table A2. SAMPA simplified list of vowels

| SAMPA | IPA | Description | Examples |
|---|---|---|---|
| i | i | front closed unrounded vowel | English *see*, Spanish *sí*, French *vite*, German *mieten*, Italian *visto* |
| I | I | front closed unrounded vowel, but somewhat more centralised and relaxed | English *city*, German *mit* |
| e | e | front half closed unrounded vowel | US English *bear*, Spanish *él*, French *année*, German *mehr*, Italian *rete*, Catalan *més* |
| E | ɛ | front half open unrounded vowel | English *bed*, French *même*, German *Herr*, *Männer*, Italian *ferro*, Catalan *mes*, Spanish *perro* |
| { | æ | front open unrounded vowel | English *cat* |
| y | y | front closed rounded vowel | French *du*, German *Tür* |
| 2 | ø | front half closed rounded vowel | French *deux* (hence '2'), German *Höhle* |
| 9 | œ | front half open rounded vowel | French *neuf* (hence '9'), German *Hölle* |
| 1 | ɨ | central closed unrounded vowel | Russian [m1s] 'mouse' |
| @ | ə | central neutral unrounded vowel | English *about*, *winner*, German *bitte* |
| @\ | ɘ | close-mid central unrounded vowel | English n*u*t [nɘt] |
| 6 | ɐ | central neutral unrounded vowel | German *besser* |
| 3 | ɜ | front half open unrounded vowel, but somewhat more centralised and relaxed | English *bird* |
| 3\ | ɞ | open-mid central rounded vowel | Irish t*omha*il |
| a | a | central open vowel | Spanish *da*, *barra*, French *bateau*, *lac*, German *Haar*, Italian *pazzo* |
| } | ʉ | central closed rounded vowel | Scottish English *pool*, Swedish *sju* |
| 8 | ɵ | central neutral rounded vowel | Swedish *kust* |
| & | ɶ | front open rounded vowel | American English *that* |
| M | ɯ | back closed unrounded vowel | Japanese *fuji*, Vietnamese ư Korean ㅡ |
| 7 | ɣ | back half closed unrounded vowel | Vietnamese ơ Korean 어 |
| V | ʌ | back half open unrounded vowel | RP and US English *run*, *enough* |

Table A2. SAMPA simplified list of vowels ~ Continue.

| | | | |
|---|---|---|---|
| **A** | ɑ | back open unrounded vowel | English *arm*, US English *law*, standard French *âme* |
| **u** | u | back closed rounded vowel | English *soon*, Spanish *tú*, French *goût*, German *Hut*, *Mutter*, Italian *azzurro*, *tutto* |
| **U** | ʊ | back closed rounded vowel somewhat more centralised and relaxed | English *put*, *Buddhist* |
| **o** | o | back half closed rounded vowel | US English *sore*, Scottish English *boat*, Spanish *yo*, French *beau*, German *Sohle*, Italian *dove*, Catalan *ona* |
| **O** | ɔ | back half open rounded vowel | English *law*, *caught*, Italian *uomo*, Catalan *dona*, |
| **Q** | ɒ | back open rounded vowel | British English *not*, *cough*, German *Toll* |